%% file: main.tex
\pdfoutput=1 
\PassOptionsToPackage{svgnames,dvipsnames}{xcolor} 
\documentclass[manuscript,nonacm,oneside]{acmart}
\setcopyright{none}

\usepackage{forest}
\useforestlibrary{edges}   
\usepackage{tikz}
\usetikzlibrary{arrows.meta,positioning,fit,backgrounds,shapes.geometric,calc}
\usepackage{pifont}
\usepackage{longtable}
\usepackage{array}      
\usepackage{placeins}   
\usepackage{booktabs}
\usepackage{multirow}

\input{styles/taxonomy_style}

\definecolor{ForestGreen}{rgb}{0.13,0.55,0.13}
\definecolor{BrickRed}{rgb}{0.70,0.13,0.13}
\newcommand{\yy}{\textcolor{ForestGreen}{\ding{52}}}       
\newcommand{\nn}{\textcolor{BrickRed}{\ding{56}}}          
\newcommand{\dm}{\textcolor{gray!65}{\textbf{-{}-}}}        
\newcommand{\pp}{\textcolor{orange!90!black}{\textbf{\textasciitilde}}} 
\newcommand{\dopen}{\textcolor{magenta!80!black}{\textsf{open}}}    
\newcommand{\dclosed}{\textcolor{teal!80!black}{\textsf{closed}}}   
\newcommand{\dbridge}{\textcolor{violet!80!black}{\textsf{bridge}}} 

\begin{document}

\title{Do World Models Make Better Robots?}
\subtitle{A Survey of Evaluation Benchmarks for Predictive Embodied Intelligence}

\author{Gaytri Jena}
\authornote{The authors contributed to this work independently of their roles and employment.}
\affiliation{\institution{UC Berkeley}\city{Berkeley}\state{CA}\country{USA}}
\author{Kapil Wanaskar}
\affiliation{\institution{San Jose State University}\country{USA}}
\author{Vinija Jain}
\affiliation{\institution{Meta}\country{USA}}
\author{Aman Chadha}
\affiliation{\institution{Apple}\country{USA}}
\author{Vasu Sharma}
\affiliation{\institution{PocketFM}\country{USA}}
\author{Amitava Das}
\affiliation{\institution{Pragya Lab, BITS Pilani Goa}\country{India}}
\renewcommand{\shortauthors}{Jena et al.}

\input{sections/0_abstract}

\ccsdesc[500]{Computing methodologies~Robotic planning}
\ccsdesc[300]{Computing methodologies~Computer vision}
\ccsdesc[300]{Computing methodologies~Machine learning}

\keywords{robot learning, world models, evaluation, benchmarks, vision-language-action
  models, embodied AI, predictive models, survey}

\maketitle

\input{sections/1_intro}
\input{sections/2_methodology}

\input{sections/3_taxonomy}
\input{sections/4_gap}
\input{sections/5_future}
\input{sections/limitations}
\input{sections/conclusion}

\clearpage
\bibliographystyle{ACM-Reference-Format}
\bibliography{refs}

\clearpage
\appendix
\input{sections/appendix}

\end{document}

%% file: styles/taxonomy_style.tex
\definecolor{hidden-draw}{RGB}{40,40,40}
\definecolor{tx-root}{HTML}{D1D5DB}

\definecolor{cat-code}{HTML}{93C5FD}\definecolor{leaf-code}{HTML}{DBEAFE}   
\definecolor{cat-vla}{HTML}{5EEAD4}\definecolor{leaf-vla}{HTML}{CCFBF1}     
\definecolor{cat-rew}{HTML}{FDBA74}\definecolor{leaf-rew}{HTML}{FEEBC8}     
\definecolor{cat-skill}{HTML}{C4B5FD}\definecolor{leaf-skill}{HTML}{EDE9FE} 
\definecolor{cat-trans}{HTML}{F9A8D4}\definecolor{leaf-trans}{HTML}{FCE7F3} 
\definecolor{cat-bench}{HTML}{86EFAC}\definecolor{leaf-bench}{HTML}{DCFCE7} 
\definecolor{barneutral}{HTML}{64748B} 

\definecolor{secA}{HTML}{93C5FD}\definecolor{secAl}{HTML}{DBEAFE} 
\definecolor{secB}{HTML}{5EEAD4}\definecolor{secBl}{HTML}{CCFBF1} 
\definecolor{secC}{HTML}{86EFAC}\definecolor{secCl}{HTML}{DCFCE7} 
\definecolor{secD}{HTML}{FCD34D}\definecolor{secDl}{HTML}{FEF3C7} 
\definecolor{secE}{HTML}{FCA5A5}\definecolor{secEl}{HTML}{FEE2E2} 
\definecolor{secF}{HTML}{C4B5FD}\definecolor{secFl}{HTML}{EDE9FE} 
\definecolor{secG}{HTML}{F9A8D4}\definecolor{secGl}{HTML}{FCE7F3} 

\forestset{
  default preamble={
    forked edges,
    for tree={
      grow=east,
      reversed=true,
      anchor=base west,
      parent anchor=east,
      child anchor=west,
      base=center,
      font=\large,
      rectangle,
      draw=hidden-draw,
      rounded corners,
      align=center,
      text centered,
      minimum width=5em,
      edge+={darkgray, line width=1pt},
      s sep=3pt,
      inner xsep=2pt,
      inner ysep=3pt,
      line width=0.8pt,
      ver/.style={rotate=90, child anchor=north, parent anchor=south, anchor=center},
      if level=1{text width=19em, font=\normalsize}{},
      if level=2{text width=33em, font=\normalsize}{},
    },
  },
}

%% file: sections/0_abstract.tex
\begin{abstract}
Robot learning now advances along two tracks that rarely meet. On one side, direct
Vision-Language-Action (VLA) policies map observations to actions and are scored by closed-loop task
success. On the other, predictive and generative world models forecast future observations and are
scored by open-loop prediction or generation quality. A natural question sits between them: does
world modelling earn a measurable, closed-loop \emph{advantage} over a direct policy, and for which
robotic capabilities? We argue that the field cannot yet answer this question, and that the reason
is a gap in how it is measured, not in the models themselves. World-model benchmarks score
prediction without ever executing it, while task-success suites host a single policy and never build
a world-model versus VLA contrast.

This survey maps the evaluation landscape around that gap. We catalogue 160 web-verified benchmarks
spanning 2017 to 2026 and organise them by evaluation mode, robotic capability, and model family
into four lanes: policy suites, embodied agents, world-model evaluation, and prediction-to-action
bridges. Across the corpus, 138 of 160 benchmarks are model-agnostic and only 11 (7\%) build an
explicit VLA-versus-world-model contrast; counterfactual capability is almost entirely unmeasured,
and only four benchmarks turn prediction into executed action. We contribute an operational
taxonomy, a coverage comparison against the eight closest surveys (ours is the only one to cross
capability with model family), an evaluation loop that isolates the advantage of prediction, and an
actionable protocol of four advantage-aware metrics anchored on named testbeds. The organising claim
is not that world models help or do not help, but that answering the question requires benchmarks
built to ask it.
\end{abstract}

%% file: sections/1_intro.tex
\section{Introduction}
\label{sec:intro}

A robot can be taught to act in two broadly different ways. The first learns a \emph{direct policy}:
a function that maps an observation, and often a language instruction, straight to an action. Recent
Vision-Language-Action (VLA) models are the prominent instance of this track, and they are judged by
running them in an environment and counting task success, on suites such as
LIBERO~\cite{libero}, CALVIN~\cite{calvin}, RLBench~\cite{rlbench}, and
ManiSkill2~\cite{maniskill2}. The second track learns a \emph{world model}: a predictor of future
observations that a controller can plan against. This track spans latent-dynamics predictors
(DreamerV3, TD-MPC2), action-conditioned video predictors (iVideoGPT, GR-2), and large generative
video models (Sora, Cosmos, Genie), and it is judged not by acting but by the quality of what it
predicts or generates, on suites such as VBench~\cite{vbench}, EvalCrafter~\cite{evalcrafter}, and
EWMBench~\cite{ewmbench}.

Between these two tracks sits a question that neither is built to answer: does world modelling earn a
measurable, closed-loop \emph{advantage} over a direct policy, and for which robotic capabilities?
Figure~\ref{fig:corpus_collage} poses that question with real benchmarks. On the left, closed-loop
policy and embodied suites score success by executing a policy in the environment. On the right,
open-loop world-model and video-generation suites score prediction or generation with no action
taken. The axis down the centre, whether prediction buys a closed-loop advantage, is the one the
field has not instrumented.

\input{figures/fig_corpus_collage}

The gap is methodological rather than empirical. A world-model benchmark scores a rollout for
realism or physical plausibility and then stops; it never closes the loop, so it cannot report
whether the prediction would have helped a robot act. A task-success suite does close the loop, but
it hosts a single policy and is deliberately model-agnostic, so it never contrasts a world-model
policy against a direct VLA policy under one protocol. A benchmark that could answer the question
must run \emph{both} branches in one closed loop and report the per-capability gain
(Section~\ref{sec:gap}, Figure~\ref{fig:eval_loop}). Only a handful of recent
``bridge'' benchmarks attempt this, and none yet reports a capability-sliced VLA head-to-head.

This survey is therefore about \emph{evaluation}, not about models. It maps the benchmark landscape
that surrounds the question and asks which benchmarks could, in principle, establish an
advantage of prediction. We catalogue 160 web-verified benchmarks spanning 2017 to 2026 and organise
them by evaluation mode, robotic capability, and model family. The headline finding is stark: 138 of
the 160 benchmarks are model-agnostic and only 11 (7\%) build an explicit VLA-versus-world-model
contrast, counterfactual capability is almost entirely unmeasured, and only four benchmarks turn
prediction into executed action.

\paragraph{Contributions.} This survey offers the following.
\begin{itemize}
  \item An \textbf{operational taxonomy} of 160 robot-evaluation benchmarks across four evaluation
        lanes and twelve capability areas (Figure~\ref{fig:taxonomy_main}), with per-benchmark detail
        in Tables~\ref{tab:compare_main} and~\ref{tab:landscape} and placement rules in
        Table~\ref{tab:defs}.
  \item A \textbf{profile of the corpus} (Figures~\ref{fig:corpus_overview},
        \ref{fig:corpus_dist},~\ref{fig:lane_trend}) and two galleries of the benchmarks' own subject
        and results figures (Figures~\ref{fig:subject_gallery},~\ref{fig:results_gallery}), each panel
        provenance-tracked.
  \item A \textbf{coverage comparison} against the eight closest benchmark and evaluation surveys
        (Table~\ref{tab:survey_compare}): ours is the only one that crosses robotic capability with
        model family under a single protocol.
  \item A \textbf{contrast-gap analysis} (Section~\ref{sec:gap}): an evaluation loop that isolates
        the advantage of prediction (Figure~\ref{fig:eval_loop}), a disambiguation of the overloaded
        term ``world model'' (Table~\ref{tab:wm_senses}), and a lane-by-lane account of why the
        contrast is missing (Table~\ref{tab:branch_limits}).
  \item An \textbf{actionable protocol} for advantage-aware evaluation: four measurable quantities
        with named testbeds (Table~\ref{tab:future_metrics}, Figure~\ref{fig:future_protocol}).
\end{itemize}

\noindent The organising claim is deliberately modest. We do not argue that world models help, or
that they do not. We argue that the question is currently unanswerable with the benchmarks the field
has built, and we specify what a benchmark that could answer it would have to do.

\subsection{Related surveys and the empty cell}
\label{sec:related}

Table~\ref{tab:survey_compare} places the eight closest surveys and position papers against five axes
that a survey of predictive embodied evaluation could organise around: ($\alpha$) evaluation mode,
open- versus closed-loop, as the \emph{primary} axis; ($\beta$) robotic capability, such as
hidden-state, counterfactual, long-horizon, and social; ($\gamma$) a capability-by-model-family
cross-tabulation, direct VLA against latent, action-conditioned, and generative world models under
one protocol; ($\delta$) a behavioural advantage-of-prediction metric, which we \emph{assess} in the
literature rather than adopt; and ($\varepsilon$) a comprehensive benchmark catalogue. Existing
surveys populate $\varepsilon$ well and touch $\alpha$ and $\beta$ in passing, but the capability-by-family
cross-tab $\gamma$ is absent everywhere, and the advantage metric $\delta$ is reached only partially,
and only by a position paper~\cite{pos_yangyu}, never by a survey. Figure~\ref{fig:surveys} places
these works on a timeline: two predate the recent world-model wave, and the rest cluster in the last
two years, none of them crossing capability with family.

\input{tables/tab_survey_compare}
\input{figures/fig_survey_timeline}

%% file: figures/fig_corpus_collage.tex
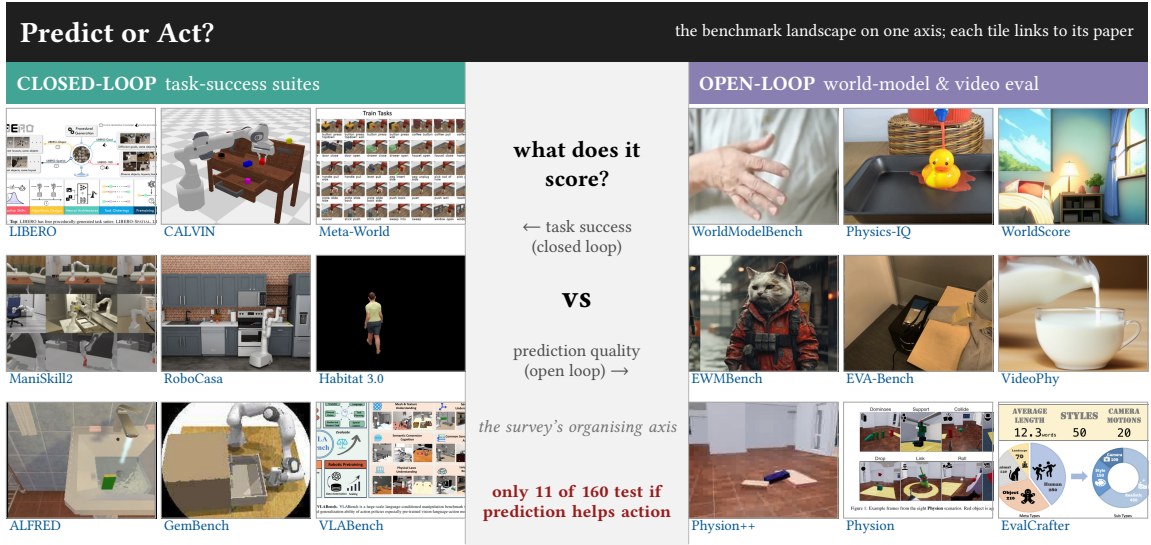
\begin{figure*}[tp]
\centering
\resizebox{\textwidth}{!}{%
\begin{tikzpicture}[x=1cm,y=1cm]
\fill[black!88] (0,0) rectangle (13.80,-0.72);
\node[anchor=west,text=white,font=\large\bfseries,inner sep=0] at (0.16,-0.36) {Predict or Act?};
\node[anchor=east,text=white!70,font=\scriptsize,inner sep=0] at (13.64,-0.36) {the benchmark landscape on one axis; each tile links to its paper};
\fill[cat-vla!70!black] (0,-0.72) rectangle (5.55,-1.22);
\node[anchor=west,text=white,font=\footnotesize,inner sep=0] at (0.12,-0.97) {{\bfseries CLOSED-LOOP}\, task-success suites};
\fill[cat-skill!70!black] (8.25,-0.72) rectangle (13.80,-1.22);
\node[anchor=west,text=white,font=\footnotesize,inner sep=0] at (8.37,-0.97) {{\bfseries OPEN-LOOP}\, world-model \& video eval};
\fill[black!5] (5.55,-0.72) rectangle (8.25,-6.55);
\draw[black!25,line width=0.4pt] (5.55,-0.72)--(5.55,-6.55);
\draw[black!25,line width=0.4pt] (8.25,-0.72)--(8.25,-6.55);
\node[text=black,font=\small\bfseries,align=center,inner sep=0] at (6.90,-1.95) {what does it\\ score?};
\node[text=black!70,font=\scriptsize,align=center,inner sep=0] at (6.90,-2.85) {$\leftarrow$ task success\\ (closed loop)};
\node[text=black,font=\Large\bfseries,inner sep=0] at (6.90,-3.60) {vs};
\node[text=black!70,font=\scriptsize,align=center,inner sep=0] at (6.90,-4.35) {prediction quality\\ (open loop) $\rightarrow$};
\node[text=black!55,font=\scriptsize\itshape,align=center,text width=2.4cm,inner sep=0] at (6.90,-5.15) {the survey's organising axis};
\node[text=BrickRed!85!black,font=\scriptsize\bfseries,align=center,text width=2.5cm,inner sep=0] at (6.90,-6.05) {only 11 of 160 test if prediction helps action};
\node[anchor=north west,inner sep=0] at (0.000,-1.280) {\href{https://arxiv.org/abs/2306.03310}{\includegraphics[width=1.803cm,height=1.402cm]{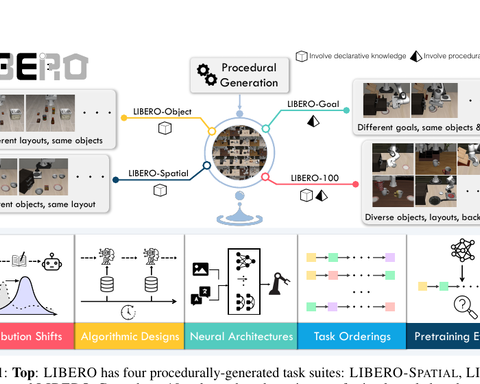}}};
\draw[black!35,line width=0.35pt] (0.000,-1.280) rectangle (1.803,-2.682);
\node[anchor=north west,text=ACMDarkBlue,font=\tiny,inner sep=0] at (0.030,-2.702) {\href{https://arxiv.org/abs/2306.03310}{LIBERO}};
\node[anchor=north west,inner sep=0] at (1.873,-1.280) {\href{https://arxiv.org/abs/2112.03227}{\includegraphics[width=1.803cm,height=1.402cm]{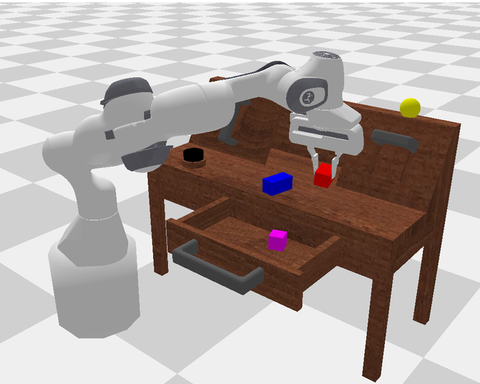}}};
\draw[black!35,line width=0.35pt] (1.873,-1.280) rectangle (3.676,-2.682);
\node[anchor=north west,text=ACMDarkBlue,font=\tiny,inner sep=0] at (1.903,-2.702) {\href{https://arxiv.org/abs/2112.03227}{CALVIN}};
\node[anchor=north west,inner sep=0] at (3.747,-1.280) {\href{https://arxiv.org/abs/1910.10897}{\includegraphics[width=1.803cm,height=1.402cm]{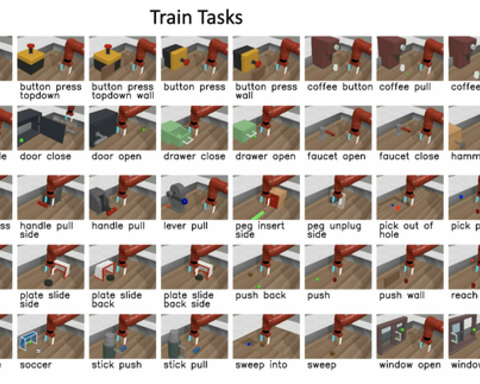}}};
\draw[black!35,line width=0.35pt] (3.747,-1.280) rectangle (5.550,-2.682);
\node[anchor=north west,text=ACMDarkBlue,font=\tiny,inner sep=0] at (3.777,-2.702) {\href{https://arxiv.org/abs/1910.10897}{Meta-World}};
\node[anchor=north west,inner sep=0] at (0.000,-3.057) {\href{https://arxiv.org/abs/2302.04659}{\includegraphics[width=1.803cm,height=1.402cm]{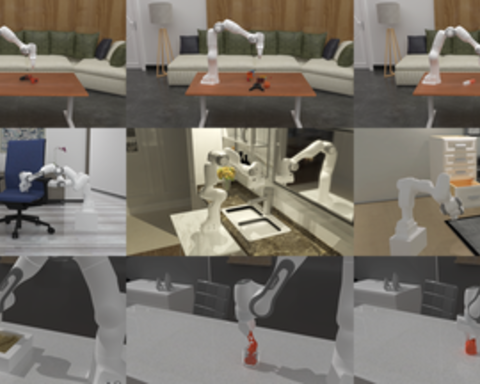}}};
\draw[black!35,line width=0.35pt] (0.000,-3.057) rectangle (1.803,-4.459);
\node[anchor=north west,text=ACMDarkBlue,font=\tiny,inner sep=0] at (0.030,-4.479) {\href{https://arxiv.org/abs/2302.04659}{ManiSkill2}};
\node[anchor=north west,inner sep=0] at (1.873,-3.057) {\href{https://arxiv.org/abs/2406.02523}{\includegraphics[width=1.803cm,height=1.402cm]{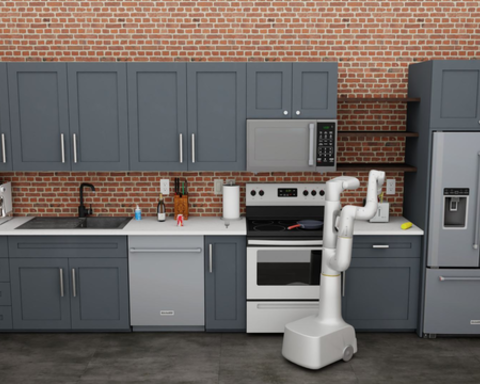}}};
\draw[black!35,line width=0.35pt] (1.873,-3.057) rectangle (3.676,-4.459);
\node[anchor=north west,text=ACMDarkBlue,font=\tiny,inner sep=0] at (1.903,-4.479) {\href{https://arxiv.org/abs/2406.02523}{RoboCasa}};
\node[anchor=north west,inner sep=0] at (3.747,-3.057) {\href{https://arxiv.org/abs/2310.13724}{\includegraphics[width=1.803cm,height=1.402cm]{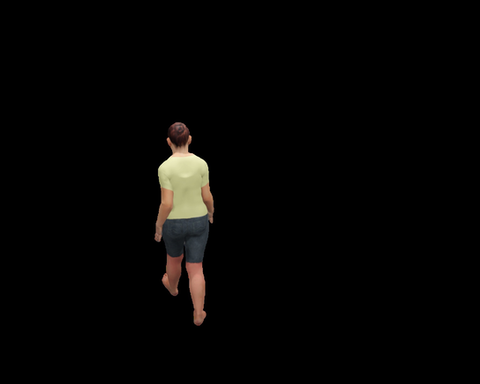}}};
\draw[black!35,line width=0.35pt] (3.747,-3.057) rectangle (5.550,-4.459);
\node[anchor=north west,text=ACMDarkBlue,font=\tiny,inner sep=0] at (3.777,-4.479) {\href{https://arxiv.org/abs/2310.13724}{Habitat 3.0}};
\node[anchor=north west,inner sep=0] at (0.000,-4.833) {\href{https://arxiv.org/abs/1912.01734}{\includegraphics[width=1.803cm,height=1.402cm]{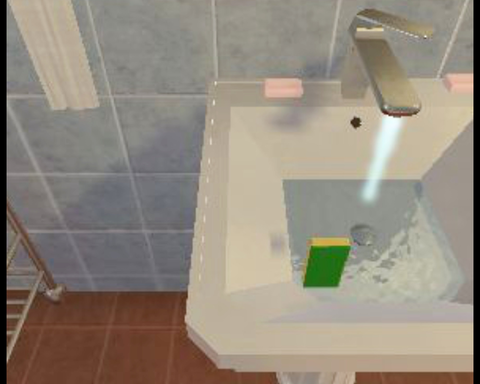}}};
\draw[black!35,line width=0.35pt] (0.000,-4.833) rectangle (1.803,-6.235);
\node[anchor=north west,text=ACMDarkBlue,font=\tiny,inner sep=0] at (0.030,-6.255) {\href{https://arxiv.org/abs/1912.01734}{ALFRED}};
\node[anchor=north west,inner sep=0] at (1.873,-4.833) {\href{https://arxiv.org/abs/2410.01345}{\includegraphics[width=1.803cm,height=1.402cm]{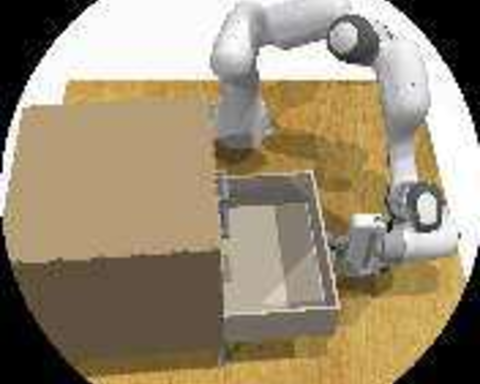}}};
\draw[black!35,line width=0.35pt] (1.873,-4.833) rectangle (3.676,-6.235);
\node[anchor=north west,text=ACMDarkBlue,font=\tiny,inner sep=0] at (1.903,-6.255) {\href{https://arxiv.org/abs/2410.01345}{GemBench}};
\node[anchor=north west,inner sep=0] at (3.747,-4.833) {\href{https://arxiv.org/abs/2412.18194}{\includegraphics[width=1.803cm,height=1.402cm]{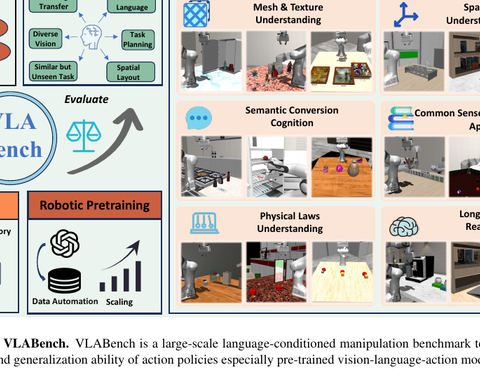}}};
\draw[black!35,line width=0.35pt] (3.747,-4.833) rectangle (5.550,-6.235);
\node[anchor=north west,text=ACMDarkBlue,font=\tiny,inner sep=0] at (3.777,-6.255) {\href{https://arxiv.org/abs/2412.18194}{VLABench}};
\node[anchor=north west,inner sep=0] at (8.250,-1.280) {\href{https://arxiv.org/abs/2502.20694}{\includegraphics[width=1.803cm,height=1.402cm]{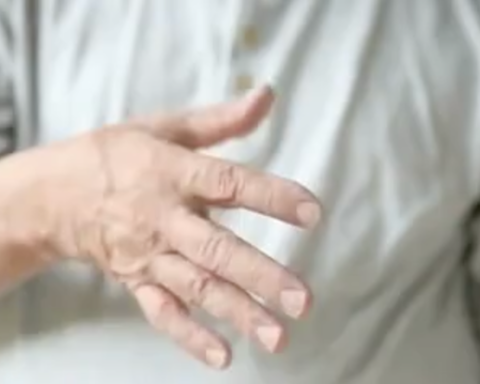}}};
\draw[black!35,line width=0.35pt] (8.250,-1.280) rectangle (10.053,-2.682);
\node[anchor=north west,text=ACMDarkBlue,font=\tiny,inner sep=0] at (8.280,-2.702) {\href{https://arxiv.org/abs/2502.20694}{WorldModelBench}};
\node[anchor=north west,inner sep=0] at (10.123,-1.280) {\href{https://arxiv.org/abs/2501.09038}{\includegraphics[width=1.803cm,height=1.402cm]{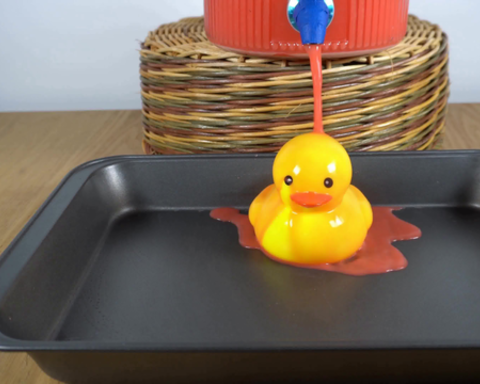}}};
\draw[black!35,line width=0.35pt] (10.123,-1.280) rectangle (11.926,-2.682);
\node[anchor=north west,text=ACMDarkBlue,font=\tiny,inner sep=0] at (10.153,-2.702) {\href{https://arxiv.org/abs/2501.09038}{Physics-IQ}};
\node[anchor=north west,inner sep=0] at (11.997,-1.280) {\href{https://arxiv.org/abs/2504.00983}{\includegraphics[width=1.803cm,height=1.402cm]{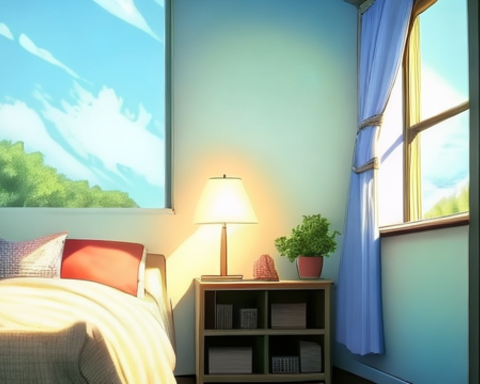}}};
\draw[black!35,line width=0.35pt] (11.997,-1.280) rectangle (13.800,-2.682);
\node[anchor=north west,text=ACMDarkBlue,font=\tiny,inner sep=0] at (12.027,-2.702) {\href{https://arxiv.org/abs/2504.00983}{WorldScore}};
\node[anchor=north west,inner sep=0] at (8.250,-3.057) {\href{https://arxiv.org/abs/2505.09694}{\includegraphics[width=1.803cm,height=1.402cm]{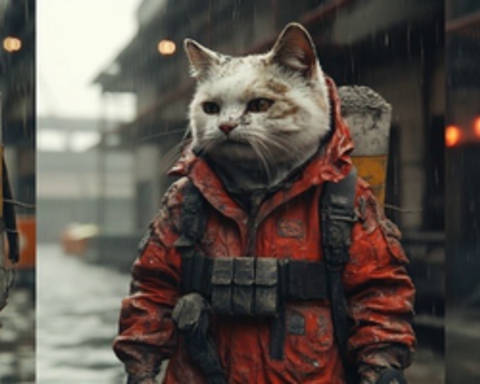}}};
\draw[black!35,line width=0.35pt] (8.250,-3.057) rectangle (10.053,-4.459);
\node[anchor=north west,text=ACMDarkBlue,font=\tiny,inner sep=0] at (8.280,-4.479) {\href{https://arxiv.org/abs/2505.09694}{EWMBench}};
\node[anchor=north west,inner sep=0] at (10.123,-3.057) {\href{https://arxiv.org/abs/2410.15461}{\includegraphics[width=1.803cm,height=1.402cm]{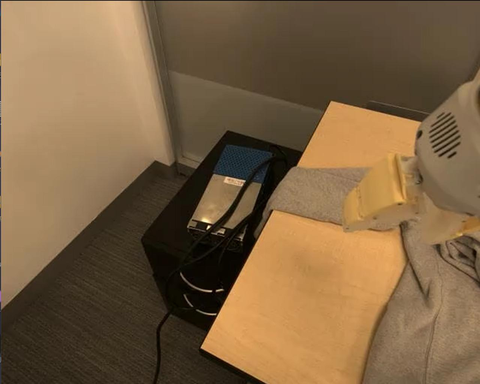}}};
\draw[black!35,line width=0.35pt] (10.123,-3.057) rectangle (11.926,-4.459);
\node[anchor=north west,text=ACMDarkBlue,font=\tiny,inner sep=0] at (10.153,-4.479) {\href{https://arxiv.org/abs/2410.15461}{EVA-Bench}};
\node[anchor=north west,inner sep=0] at (11.997,-3.057) {\href{https://arxiv.org/abs/2406.03520}{\includegraphics[width=1.803cm,height=1.402cm]{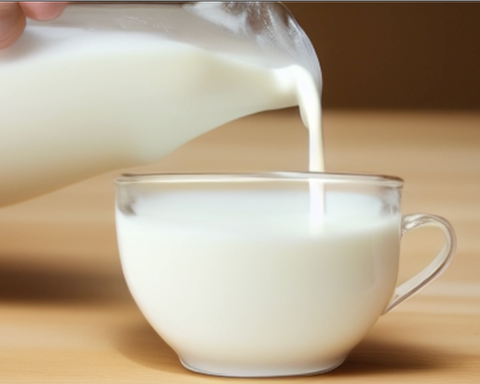}}};
\draw[black!35,line width=0.35pt] (11.997,-3.057) rectangle (13.800,-4.459);
\node[anchor=north west,text=ACMDarkBlue,font=\tiny,inner sep=0] at (12.027,-4.479) {\href{https://arxiv.org/abs/2406.03520}{VideoPhy}};
\node[anchor=north west,inner sep=0] at (8.250,-4.833) {\href{https://arxiv.org/abs/2306.15668}{\includegraphics[width=1.803cm,height=1.402cm]{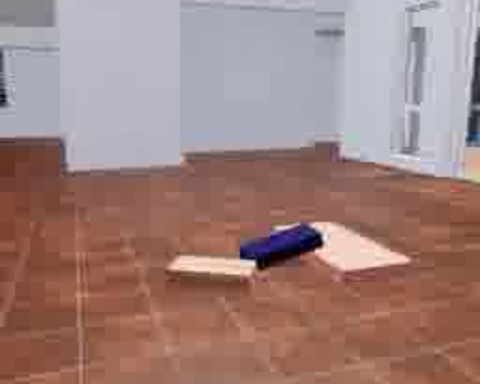}}};
\draw[black!35,line width=0.35pt] (8.250,-4.833) rectangle (10.053,-6.235);
\node[anchor=north west,text=ACMDarkBlue,font=\tiny,inner sep=0] at (8.280,-6.255) {\href{https://arxiv.org/abs/2306.15668}{Physion++}};
\node[anchor=north west,inner sep=0] at (10.123,-4.833) {\href{https://arxiv.org/abs/2106.08261}{\includegraphics[width=1.803cm,height=1.402cm]{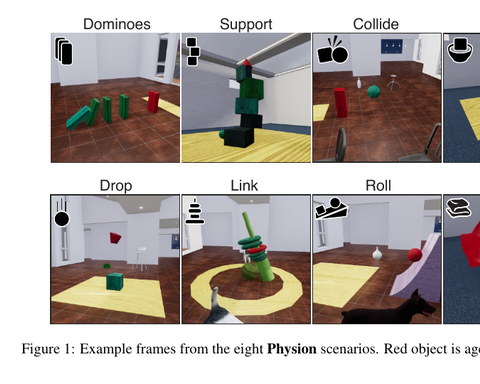}}};
\draw[black!35,line width=0.35pt] (10.123,-4.833) rectangle (11.926,-6.235);
\node[anchor=north west,text=ACMDarkBlue,font=\tiny,inner sep=0] at (10.153,-6.255) {\href{https://arxiv.org/abs/2106.08261}{Physion}};
\node[anchor=north west,inner sep=0] at (11.997,-4.833) {\href{https://arxiv.org/abs/2310.11440}{\includegraphics[width=1.803cm,height=1.402cm]{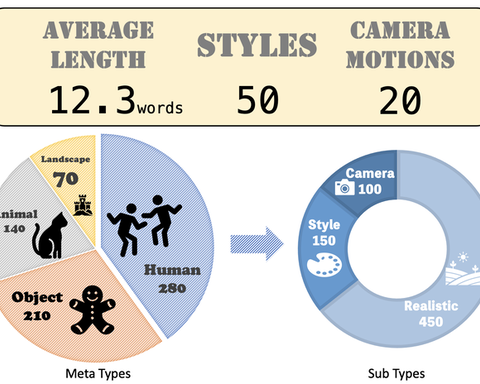}}};
\draw[black!35,line width=0.35pt] (11.997,-4.833) rectangle (13.800,-6.235);
\node[anchor=north west,text=ACMDarkBlue,font=\tiny,inner sep=0] at (12.027,-6.255) {\href{https://arxiv.org/abs/2310.11440}{EvalCrafter}};
\end{tikzpicture}}
\caption{\textbf{Predict or Act?} The survey's organising question, posed with real benchmarks. \emph{Closed-loop} (left): policy and embodied suites score task success by running a policy in the environment. \emph{Open-loop} (right): world-model and video-generation benchmarks score prediction or generation quality with no action. The centre axis is the question neither pole answers, whether prediction earns a closed-loop advantage, built explicitly by only 11 of 160 benchmarks (Table~\ref{tab:landscape}). Every tile is a representative figure from the benchmark's paper and links to it; per-tile provenance is in Table~\ref{tab:collage_provenance}. Images \textcopyright\ their respective authors, reproduced for scholarly review.}
\Description{A two-pole collage of benchmark figures. A black title bar reads ``Predict or Act?''. The left band, CLOSED-LOOP, shows nine policy and embodied benchmark figures; the right band, OPEN-LOOP, shows nine world-model and video-generation benchmark figures; a centre spine contrasts task success (closed loop) with prediction quality (open loop) and notes that only 11 of 160 build the contrast.}
\label{fig:corpus_collage}
\end{figure*}

%% file: tables/tab_survey_compare.tex
\begin{table}[t]
\centering
\footnotesize
\setlength{\tabcolsep}{4.5pt}
\renewcommand{\arraystretch}{1.25}
\resizebox{\textwidth}{!}{%
\begin{tabular}{@{}p{0.26\textwidth}c c *{5}{c}@{}}
\toprule
\textbf{Survey} & \textbf{Venue} & \textbf{Yr}
 & \shortstack{$\alpha$\\{\scriptsize mode}} & \shortstack{$\beta$\\{\scriptsize cap.}}
 & \shortstack{$\gamma$\\{\scriptsize family}} & \shortstack{$\delta$\\{\scriptsize metric}}
 & \shortstack{$\varepsilon$\\{\scriptsize catalog}}\\
\midrule
Evaluation of Embodied AI~\cite{survey_evalembodied} & Authorea & 2026 & \pp & \pp & \nn & \nn & \yy\\
VLA Datasets \& Bench.~\cite{survey_vladata} & arXiv & 2026 & \nn & \nn & \nn & \nn & \yy\\
Benchmark Construction~\cite{survey_benchconstruct} & arXiv & 2026 & \nn & \pp & \nn & \nn & \yy\\
RL Reproducibility~\cite{survey_rlrepro} & CoRL & 2020 & \nn & \nn & \nn & \nn & \nn\\
World Models for Robots~\cite{survey_wmrobot} & arXiv & 2026 & \pp & \nn & \pp & \pp & \pp\\
Embodied World Models~\cite{survey_embodiedwm} & preprint & 2026 & \pp & \nn & \pp & \nn & \pp\\
WM Evaluation \emph{(pos.)}~\cite{pos_yangyu} & arXiv & 2026 & \yy & \pp & \nn & \pp & \nn\\
Embodied AI: Simulators~\cite{survey_embodiedsim} & TETCI & 2022 & \pp & \pp & \nn & \nn & \yy\\
\midrule
\textbf{Ours (this survey)} & \dm & 2026 & \yy & \yy & \yy & \yy & \yy\\
\bottomrule
\end{tabular}}
\caption{\textbf{Topical coverage of the closest benchmark/evaluation surveys versus ours.}
\yy\ covered as an organising axis, \pp\ partial or mentioned in passing, \nn\ absent.
Axes: \textbf{$\alpha$} evaluation mode (open- vs closed-loop as the \emph{primary} axis);
\textbf{$\beta$} capability cuts (hidden-state, counterfactual, long-horizon, social);
\textbf{$\gamma$} capability~$\times$~model-family cross-tab (direct VLA vs latent /
action-conditioned / generative world models under one protocol); \textbf{$\delta$} behavioural
advantage-of-prediction metric (\emph{assessed, not adopted}; e.g.\ ACTION-ATLAS's World Advantage
Score); \textbf{$\varepsilon$} comprehensive benchmark catalogue. This survey is the only work that
covers \textbf{$\gamma$}, and \textbf{$\delta$} is otherwise reached only \emph{partially}, and only
by a \emph{position} paper~\cite{pos_yangyu}, never by a survey. Venues: CoRL, IEEE TETCI; the
remaining rows are arXiv or unrefereed preprints (Authorea, Preprints.org).}
\label{tab:survey_compare}
\end{table}

%% file: figures/fig_survey_timeline.tex
\begin{figure}[t]
\centering
\resizebox{\columnwidth}{!}{%
\begin{tikzpicture}[
 abv/.style={font=\scriptsize, anchor=south, align=center},
 blw/.style={font=\scriptsize, anchor=north, align=center},
 stem/.style={gray!55, line width=0.5pt},
]
 \draw[-{Stealth[length=3.2mm]}, line width=1.3pt, gray!65] (-0.4,0) -- (13.4,0);
 \node[font=\scriptsize\itshape, gray, anchor=west] at (13.0,-0.28) {time};
 \foreach \x/\yr in {0.3/2019, 2.3/2021, 8.5/2026}{\draw[gray!35] (\x,-0.12)--(\x,0.12);
   \node[font=\footnotesize\bfseries,gray!80] at (\x,-1.15) {\yr};}
 \draw[gray!30,dashed] (3.3,-0.9)--(3.3,0.9); 
 \filldraw[fill=secD,draw=hidden-draw] (-0.35,1.5) circle (3pt);
 \node[font=\scriptsize,gray,anchor=west] at (-0.12,1.5) {= refereed venue (CoRL, IEEE TETCI); others are preprints};
 \filldraw[fill=secD,draw=hidden-draw] (0.3,0) circle (3.6pt);
 \draw[stem] (0.3,0.12)--(0.3,0.5); \node[abv] at (0.3,0.53) {\textbf{RL Reprod.}~\cite{survey_rlrepro}};
 \filldraw[fill=secD,draw=hidden-draw] (2.3,0) circle (3.6pt);
 \draw[stem] (2.3,-0.12)--(2.3,-0.5); \node[blw] at (2.3,-0.53) {\textbf{Embodied AI: Sim.}~\cite{survey_embodiedsim}};
 \filldraw[fill=secB,draw=hidden-draw] (5.0,0) circle (3.6pt);
 \draw[stem] (5.0,0.12)--(5.0,0.5); \node[abv] at (5.0,0.53) {\textbf{Eval.\ of Embodied AI}~\cite{survey_evalembodied}};
 \filldraw[fill=secB,draw=hidden-draw] (7.0,0) circle (3.6pt);
 \draw[stem] (7.0,-0.12)--(7.0,-0.5); \node[blw] at (7.0,-0.53) {\textbf{VLA Datasets}~\cite{survey_vladata}};
 \filldraw[fill=secB,draw=hidden-draw] (8.6,0) circle (3.6pt);
 \draw[stem] (8.6,0.12)--(8.6,0.5); \node[abv] at (8.6,0.53) {\textbf{WM for Robots}~\cite{survey_wmrobot}};
 \filldraw[fill=secB,draw=hidden-draw] (10.0,0) circle (3.6pt);
 \draw[stem] (10.0,-0.12)--(10.0,-0.5); \node[blw] at (10.0,-0.53) {\textbf{Embodied WM}~\cite{survey_embodiedwm}};
 \filldraw[fill=secB,draw=hidden-draw] (11.4,0) circle (3.6pt);
 \draw[stem] (11.4,0.12)--(11.4,0.5); \node[abv] at (11.4,0.53) {\textbf{Bench.\ Constr.}~\cite{survey_benchconstruct}};
 \filldraw[fill=secE!55,draw=hidden-draw] (12.5,0) circle (3.6pt);
 \draw[stem] (12.5,-0.12)--(12.5,-0.5); \node[blw] at (12.5,-0.53) {\textbf{WM Eval.} \emph{(pos.)}~\cite{pos_yangyu}};
 \node[star,star points=5,star point ratio=2.4,fill=secE,draw=hidden-draw,minimum size=16pt,inner sep=0] at (13.1,0) {};
 \draw[secE!70!black,line width=0.6pt] (13.1,0.2)--(13.1,0.5);
 \node[abv,text=secE!45!black] at (13.1,0.6) {\textbf{Ours}};
\end{tikzpicture}}
\caption{The closest benchmark/evaluation surveys as a timeline. Two predate the recent wave, an
IEEE~TETCI survey of embodied-AI simulators and tasks~\cite{survey_embodiedsim} and a CoRL survey of
real-robot RL reproducibility~\cite{survey_rlrepro} (\textcolor{secD!55!black}{amber} = refereed
venue); the rest are 2026 preprints on embodied evaluation, VLA data, and world models. None
organises the robot-evaluation landscape by evaluation-mode $\times$ capability $\times$ model-family;
the sole work reaching the advantage-of-prediction axis is a \emph{position} paper~\cite{pos_yangyu}.
\textbf{Ours} ($\star$) fills that cell; per-axis coverage is in Table~\ref{tab:survey_compare}.}
\Description{A horizontal timeline of eight related surveys. Two amber refereed markers sit at 2019
(RL reproducibility, CoRL) and 2021 (embodied-AI simulators, IEEE TETCI); after a time break, a dense
cluster of 2026 preprint markers covers evaluation of embodied AI, VLA datasets, world models for
robots, embodied world models, benchmark construction, and a world-model-evaluation position paper. A
red star marks the present survey at the end.}
\label{fig:surveys}
\end{figure}
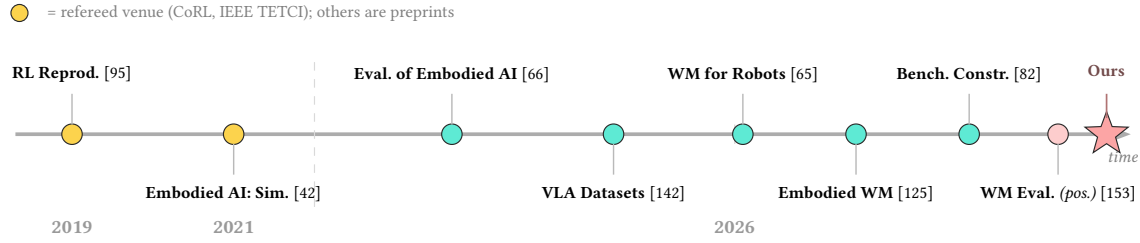

%% file: sections/2_methodology.tex
\section{Survey methodology and corpus}
\label{sec:method}

\subsection{Scope, search, and verification}

The unit of this survey is the \emph{benchmark}: an individually citeable evaluation artefact, a
suite, a dataset paired with a scoring protocol, or an interactive environment with a task
distribution, that measures robot, embodied-agent, or world-model behaviour. We deliberately exclude
three neighbouring categories that a benchmark survey is often confused with: individual models and
policies (the systems a benchmark scores), raw datasets with no scoring protocol, and pure simulators
with no task distribution. Where a work contributes both a model and a benchmark, only its benchmark
enters the corpus.

Candidates were gathered by parallel web search over arXiv, publisher venues, and project pages,
using query strings built from the cross-product of the evaluation lanes (policy, embodied,
world-model, bridge) with capability terms (manipulation, navigation, long-horizon, social,
counterfactual, physical reasoning, generation quality) and with the terms
\emph{benchmark}, \emph{evaluation}, and \emph{suite}. Every candidate was then web-verified against
its arXiv or venue page for exact title, first author, year, and identifier; any candidate that could
not be verified was dropped rather than recorded from memory, and entries were de-duplicated by both
citation key and arXiv identifier. Figure~\ref{fig:prisma} records the resulting funnel in the
PRISMA~2020 style~\cite{prisma2020}; stages that were not logged as counts are marked as such rather
than estimated. The procedure yields 160 web-verified benchmarks drawing on 164 references.

\input{figures/fig_prisma}

\subsection{Placement, definitions, and the core-versus-appendix rule}

Each benchmark is placed on four axes: its evaluation \emph{lane} (Section~\ref{sec:taxonomy}), the
robotic \emph{capability} it primarily stresses, its evaluation \emph{mode} (open-loop prediction or
generation quality, closed-loop task success, or a prediction-to-action bridge), and whether it
builds a native VLA-versus-world-model \emph{contrast} (explicit, partial, or model-agnostic).
Table~\ref{tab:defs} states the operational definition used for each placement, so that boundary
cases are resolved by a written rule rather than by taste; a benchmark counts as building a contrast,
for example, only if a direct-policy and a world-model policy are run under one protocol and compared.

Because the full corpus is large, we split it. A benchmark enters the core taxonomy
(Figure~\ref{fig:taxonomy_main}) when it is cleanly axis-placeable, distinct from its neighbours, and
fully characterisable from its paper; the 86 representative benchmarks that meet all three sit in the
taxonomy, and the complete set of 160 is catalogued in the landscape table
(Table~\ref{tab:landscape}, Appendix~\ref{app:landscape}). Every count in the survey reconciles with
that table.

\input{tables/tab_definitions}

\subsection{The corpus at a glance}

Figures~\ref{fig:corpus_overview} and~\ref{fig:corpus_dist} profile the corpus, with all counts
re-tallied directly from the catalogue rather than taken from any single paper. The 160 benchmarks
divide by lane into 37 policy suites, 85 embodied-agent benchmarks, 34 world-model-evaluation
benchmarks, and 4 prediction-to-action bridges, and by capability area into twelve clusters led by
long-horizon planning (26), manipulation and dexterity (22), and generation quality (19). By
evaluation mode, 113 score closed-loop task success, 43 score open-loop prediction or generation, and
only 4 bridge the two. Publication years show the field's shape clearly: a steady rise from 2 in 2017
to a peak of 36 in 2024, with the world-model-evaluation lane arriving almost entirely in the last
three years (Figure~\ref{fig:lane_trend}). The single most consequential distribution is the contrast
axis: 138 of 160 benchmarks are model-agnostic, 11 build a partial contrast, and 11 (7\%) build one
explicitly. That imbalance is the empirical core of the survey and is taken up in
Section~\ref{sec:gap}.

\input{figures/fig_corpus_overview}
\input{figures/fig_corpus_dist}
\input{figures/fig_lane_trend}

%% file: figures/fig_prisma.tex
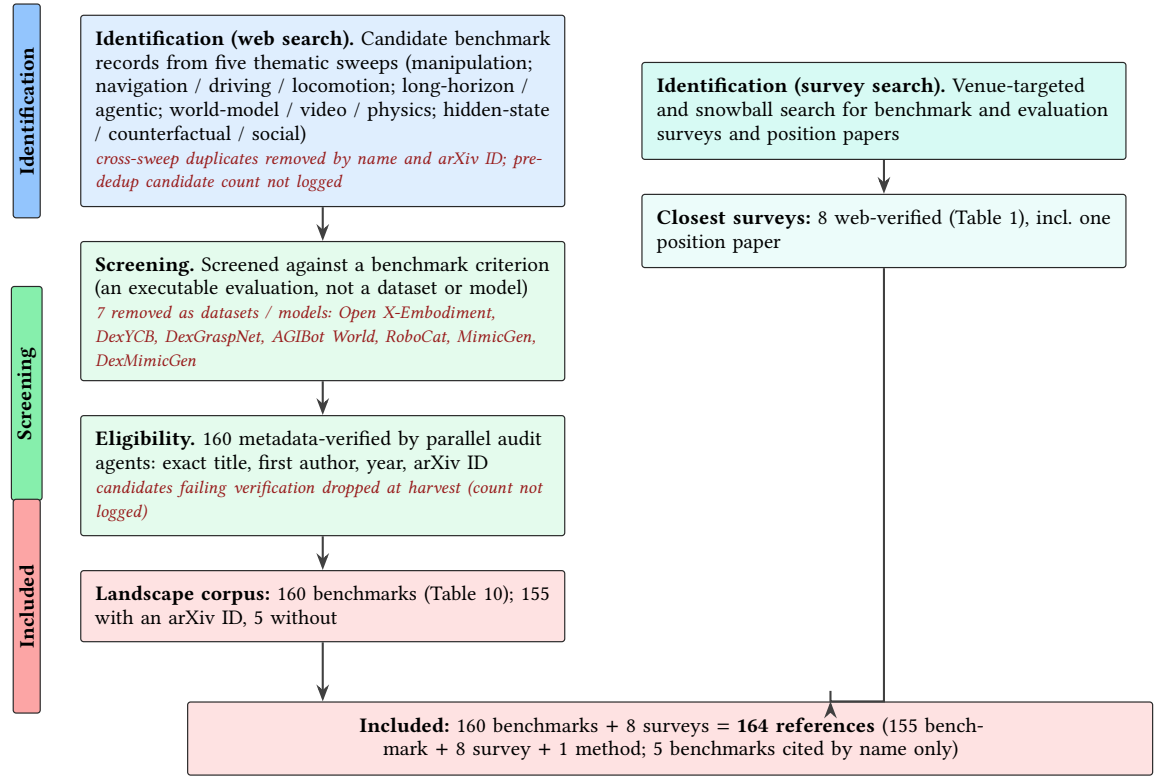
\begin{figure}[t]
\centering
\resizebox{\textwidth}{!}{%
\begin{tikzpicture}[
  font=\footnotesize,
  box/.style={draw=hidden-draw, rounded corners=1pt, align=left, inner sep=5pt,
              text width=17em, fill=white},
  ex/.style={font=\scriptsize\itshape, text=BrickRed!85!black},
  bar/.style={draw=hidden-draw, rounded corners=1pt, inner sep=2pt, rotate=90,
              anchor=center, align=center, font=\footnotesize\bfseries, minimum width=8em},
  ar/.style={-{Stealth[length=2mm]}, darkgray, line width=0.7pt},
  node distance=4mm and 9mm,
]
\node[box, fill=secA!30] (L1)
  {\textbf{Identification (web search).} Candidate benchmark records from five thematic sweeps
   (manipulation; navigation / driving / locomotion; long-horizon / agentic;
   world-model / video / physics; hidden-state / counterfactual / social)\\
   {\scriptsize\itshape\color{BrickRed!85!black} cross-sweep duplicates removed by name and arXiv ID; pre-dedup candidate count not logged}};
\node[box, fill=secC!22, below=of L1] (L2)
  {\textbf{Screening.} Screened against a benchmark criterion (an executable evaluation, not a dataset or model)\\
   {\scriptsize\itshape\color{BrickRed!85!black} 7 removed as datasets / models: Open X-Embodiment, DexYCB, DexGraspNet, AGIBot World, RoboCat, MimicGen, DexMimicGen}};
\node[box, fill=secC!22, below=of L2] (L3)
  {\textbf{Eligibility.} 160 metadata-verified by parallel audit agents: exact title, first author, year, arXiv ID\\
   {\scriptsize\itshape\color{BrickRed!85!black} candidates failing verification dropped at harvest (count not logged)}};
\node[box, fill=secEl, below=of L3] (L4)
  {\textbf{Landscape corpus:} 160 benchmarks (Table~\ref{tab:landscape}); 155 with an arXiv ID, 5 without};
\node[box, fill=secB!25, right=9mm of L1] (R1)
  {\textbf{Identification (survey search).} Venue-targeted and snowball search for benchmark and
   evaluation surveys and position papers};
\node[box, fill=secB!12, below=of R1] (R2)
  {\textbf{Closest surveys:} 8 web-verified (Table~\ref{tab:survey_compare}), incl.\ one position paper};
\node[box, fill=secEl, below=7mm of L4, xshift=13em, text width=35em, align=center] (INC)
  {\textbf{Included:} 160 benchmarks $+$ 8 surveys $=$ \textbf{164 references}
   (155 benchmark $+$ 8 survey $+$ 1 method; 5 benchmarks cited by name only)};
\draw[ar] (L1) -- (L2); \draw[ar] (L2) -- (L3); \draw[ar] (L3) -- (L4);
\draw[ar] (R1) -- (R2);
\draw[ar] (L4.south) -- (L4.south |- INC.north);
\draw[ar] (R2.south) |- ([xshift=6em]INC.north) -- ([xshift=6em]INC.north |- INC.north);
\node[bar, fill=secA] at ($(L1.west)+(-2em,0)$) {Identification};
\node[bar, fill=secC] at ($(L2.west)!0.5!(L3.west)+(-2em,0)$) {Screening};
\node[bar, fill=secE] at ($(L4.west)+(-2em,0)$) {Included};
\end{tikzpicture}}
\caption{\textbf{Corpus construction} in the PRISMA 2020 style~\citep{prisma2020}. The left column is
the web-search stream behind the landscape corpus (Table~\ref{tab:landscape}): five thematic sweeps,
de-duplicated by name and arXiv ID, screened to drop datasets and models, and metadata-verified by
parallel audit agents. The right column is the venue-targeted and snowball search behind the closest
surveys (Table~\ref{tab:survey_compare}). Both streams feed the included set. Candidates whose
metadata could not be verified were discarded at harvest; their count was not logged, which we state
rather than estimate.}
\Description{A PRISMA 2020 style flow diagram with two identification columns. The left column
(structured web search) shows candidate benchmark records from five thematic sweeps, de-duplicated by
name and arXiv ID, screened to remove seven datasets and models, and 160 metadata-verified records
forming the landscape corpus (155 with an arXiv ID, 5 without). The right column (survey search)
yields 8 closest surveys including one position paper. Both columns feed an included box of 160
benchmarks plus 8 surveys and one methodology citation, totalling 164 references.}
\label{fig:prisma}
\end{figure}

%% file: tables/tab_definitions.tex
\begin{table*}[t]
\centering
\footnotesize
\setlength{\tabcolsep}{5pt}
\renewcommand{\arraystretch}{1.3}
\resizebox{\textwidth}{!}{%
\begin{tabular}{@{}p{0.19\textwidth} p{0.40\textwidth} p{0.30\textwidth}@{}}
\toprule
\textbf{Construct} & \textbf{Counts (a benchmark has it if and only if\ldots)} & \textbf{Does not count} \\
\midrule
\textbf{Open-loop eval (\dopen)} &
it scores prediction or generation \emph{quality} on fixed inputs without executing the
prediction in a feedback loop: video fidelity (FVD), physics or rollout consistency, or
frame-level accuracy &
any protocol where an action is executed and the environment responds; offline scoring of a
policy's success \\
\textbf{Closed-loop eval (\dclosed)} &
success is measured by running a policy \emph{in} the environment with state feedback and
scoring task outcome: success rate, sub-goal completion, or reward under interaction &
scoring a generated video or a static prediction; a metric computed without stepping the
environment \\
\textbf{Bridge (\dbridge)} &
a prediction (imagined rollout, video, or latent plan) is \emph{converted into executed actions}
and scored on the resulting task success &
a benchmark that stops at prediction quality; one that never turns a prediction into a control
signal \\
\textbf{Native VLA-vs-WM contrast ($\gamma$)} &
the benchmark \emph{itself} runs both a direct Vision-Language-Action policy and a world-model
policy under one protocol and reports the head-to-head &
a model-agnostic suite that merely \emph{hosts} either family; a suite evaluated with only one
family in the original paper \\
\textbf{Advantage-of-prediction ($\delta$)} &
a metric quantifies the closed-loop \emph{gain} of a predictive policy over a matched direct-VLA
baseline on the same tasks (assessed, not adopted) &
absolute success with no matched baseline; open-loop quality scores; a ranking that never
isolates the predictive component \\
\bottomrule
\end{tabular}}
\caption{Operational definitions used to place each benchmark on the evaluation axes
(\S\ref{sec:method}). They resolve the ambiguous boundary cases: hosting a policy is not the same
as \emph{building} a VLA-vs-world-model contrast ($\gamma$), and an absolute success number is not
an advantage-of-prediction measurement ($\delta$) without a matched baseline. \dopen\ open-loop,
\dclosed\ closed-loop, \dbridge\ bridge.}
\label{tab:defs}
\end{table*}

%% file: figures/fig_corpus_overview.tex
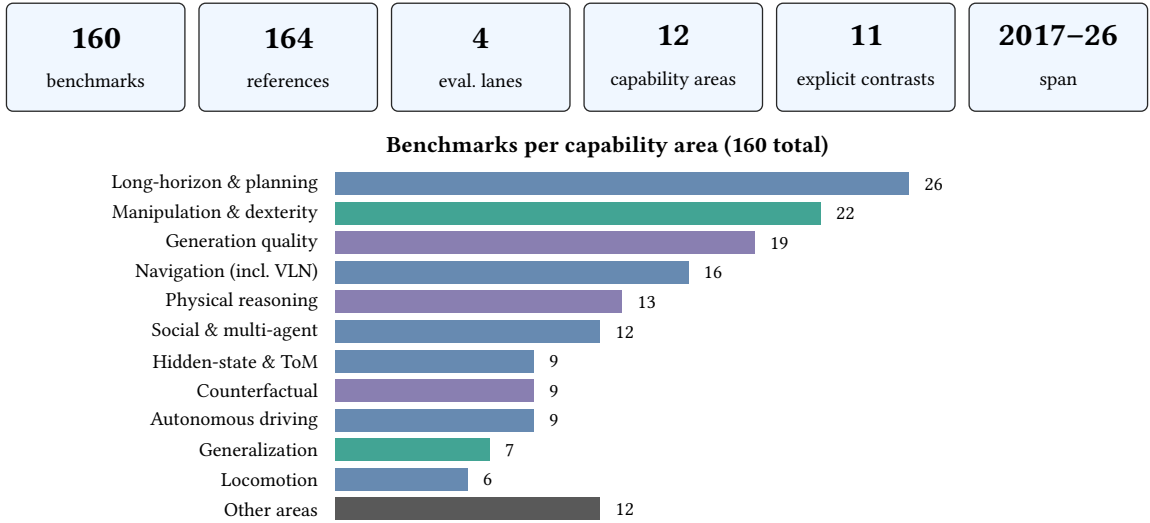
\begin{figure}[tp]
\centering
\resizebox{\textwidth}{!}{%
\begin{tikzpicture}
\foreach [count=\i] \num/\lab in {160/benchmarks, 164/references, 4/{eval.\ lanes},
                                  12/{capability areas}, 11/{explicit contrasts}, {2017--26}/span}{
  \node[draw=hidden-draw, rounded corners=2pt, fill=secA!14, minimum width=1.9cm,
        minimum height=1.15cm, align=center] at ({\i*2.05-0.2},4.9)
        {{\large\bfseries\num}\\[1pt]{\scriptsize\lab}};
}
\foreach [count=\i] \l/\v/\c in {{Long-horizon \& planning}/26/cat-code,
   {Manipulation \& dexterity}/22/cat-vla, {Generation quality}/19/cat-skill,
   {Navigation (incl.\ VLN)}/16/cat-code, {Physical reasoning}/13/cat-skill,
   {Social \& multi-agent}/12/cat-code, {Hidden-state \& ToM}/9/cat-code,
   {Counterfactual}/9/cat-skill, {Autonomous driving}/9/cat-code,
   {Generalization}/7/cat-vla, {Locomotion}/6/cat-code, {Other areas}/12/gray}{
  \fill[\c!70!black] (4.4,{3.75-\i*0.315}) rectangle ({4.4+\v*0.235},{3.75-\i*0.315+0.24});
  \node[left=2pt,font=\scriptsize] at (4.4,{3.75-\i*0.315+0.12}) {\l};
  \node[right=1.5pt,font=\scriptsize] at ({4.4+\v*0.235},{3.75-\i*0.315+0.12}) {\v};
}
\node[font=\footnotesize\bfseries] at (7.3,3.95) {Benchmarks per capability area (160 total)};
\end{tikzpicture}}
\caption{\textbf{The catalogued corpus at a glance.} The survey maps 160 web-verified benchmarks
across 4 evaluation lanes, spanning 2017--2026 and totalling 164 references; only 11 build an
explicit Vision-Language-Action vs world-model contrast. The bars give benchmarks per capability
area, with kindred clusters merged (dexterity into manipulation, vision-language navigation into
navigation, theory-of-mind into hidden-state); the twelve areas sum to the 160 works of
Table~\ref{tab:landscape}. Bars are coloured by the dominant lane of each area (teal policy, blue
embodied, purple world-model).}
\Description{An overview figure. Six stat callouts read: 160 benchmarks, 164 references, 4
evaluation lanes, 12 capability areas, 11 explicit contrasts, span 2017 to 2026. Below, a horizontal bar chart gives
benchmarks per capability area: long-horizon and planning 26, manipulation and dexterity 22,
generation quality 19, navigation 16, physical reasoning 13, social and multi-agent 12, hidden-state
and theory-of-mind 9, counterfactual 9, autonomous driving 9, generalization 7, locomotion 6, other
areas 12.}
\label{fig:corpus_overview}
\end{figure}

%% file: figures/fig_corpus_dist.tex
\begin{figure*}[tp]
\centering
\definecolor{pal1}{HTML}{C0392B}\definecolor{pal2}{HTML}{E67E22}\definecolor{pal3}{HTML}{B7950B}%
\definecolor{pal4}{HTML}{1E8449}\definecolor{pal5}{HTML}{7D3C98}\definecolor{pal6}{HTML}{C2185B}%
\definecolor{pal7}{HTML}{566573}\definecolor{pal8}{HTML}{6E2C00}\definecolor{pal9}{HTML}{7B241C}%
\definecolor{pal10}{HTML}{4A235A}%
\resizebox{\textwidth}{!}{%
\begin{tabular}{@{}c@{\hspace{7mm}}c@{}}
\begin{tikzpicture}[baseline]
  \draw[->,gray] (0,0)--(0,3.0); \draw[->,gray] (0,0)--(6.4,0);
  \foreach [count=\i] \l/\v in {17/2,18/5,19/13,20/17,21/23,22/11,23/31,24/36,25/20,26/2}{
    \fill[pal\i] ({\i*0.58-0.19},0) rectangle ({\i*0.58+0.19},{\v*0.0722});
    \node[font=\scriptsize] at ({\i*0.58},-0.18) {\l};
    \node[font=\scriptsize,above=-1pt] at ({\i*0.58},{\v*0.0722}) {\v};
  }
  \node[font=\footnotesize\bfseries] at (3.2,3.35) {(a) Benchmarks per year (2017--2026)};
\end{tikzpicture}
&
\begin{tikzpicture}[baseline]
  \draw[->,gray] (0,0)--(0,3.0); \draw[->,gray] (0,0)--(5.0,0);
  \foreach [count=\i] \l/\v/\c in {policy/37/cat-vla, embodied/85/cat-code,
                                    {wm-eval}/34/cat-skill, bridge/4/cat-bench}{
    \fill[\c!75!black] ({\i*1.1-0.3},0) rectangle ({\i*1.1+0.3},{\v*0.0306});
    \node[font=\scriptsize] at ({\i*1.1},-0.18) {\l};
    \node[font=\scriptsize,above=-1pt] at ({\i*1.1},{\v*0.0306}) {\v};
  }
  \node[font=\footnotesize\bfseries] at (2.5,3.35) {(b) Evaluation lane};
\end{tikzpicture}
\\[7mm]
\begin{tikzpicture}[baseline]
  \foreach [count=\i] \l/\v/\c in {closed/113/pal4, open/43/pal6, bridge/4/pal5}{
    \fill[\c] (1.5,{1.75-\i*0.5}) rectangle ({1.5+\v*0.020},{1.75-\i*0.5+0.32});
    \node[left=2pt,font=\scriptsize] at (1.5,{1.75-\i*0.5+0.16}) {\l};
    \node[right=1.5pt,font=\scriptsize] at ({1.5+\v*0.020},{1.75-\i*0.5+0.16}) {\v};
  }
  \node[font=\footnotesize\bfseries] at (2.2,2.1) {(c) Evaluation mode};
\end{tikzpicture}
&
\begin{tikzpicture}[baseline]
  \foreach [count=\i] \l/\v/\c in {agnostic/138/pal7, partial/11/pal2, explicit/11/pal4}{
    \fill[\c] (1.6,{1.75-\i*0.5}) rectangle ({1.6+\v*0.016},{1.75-\i*0.5+0.32});
    \node[left=2pt,font=\scriptsize] at (1.6,{1.75-\i*0.5+0.16}) {\l};
    \node[right=1.5pt,font=\scriptsize] at ({1.6+\v*0.016},{1.75-\i*0.5+0.16}) {\v};
  }
  \node[font=\footnotesize\bfseries] at (2.4,2.1) {(d) VLA-vs-world-model contrast};
\end{tikzpicture}
\end{tabular}}
\caption{\textbf{Profile of the 160 catalogue benchmarks} (Table~\ref{tab:landscape}), re-counted
directly from the catalogue. (a)~publication year, showing the 2023--2024 surge; (b)~evaluation lane
(embodied navigation dominates, then policy suites); (c)~evaluation mode (closed-loop task success
dominates, open-loop prediction is second, only four bridges reach action); (d)~whether the benchmark
builds a native Vision-Language-Action vs world-model contrast, the survey's organising question:
\textbf{138 of 160 are model-agnostic}, and only \textbf{11 (7\%)} build the contrast explicitly.
Counts are the exact column tallies of Table~\ref{tab:landscape}.}
\Description{Four bar charts of the 160 benchmarks. (a) Benchmarks per year rise from 2 in 2017 to 31
in 2023 and 36 in 2024, then 20 in 2025. (b) Evaluation lane: policy 37, embodied 85, world-model
evaluation 34, bridge 4. (c) Evaluation mode: closed 113, open 43, bridge 4. (d) VLA-vs-world-model
contrast: model-agnostic 138, partial 11, explicit 11, with a note that only 11 (7 percent) build a
contrast.}
\label{fig:corpus_dist}
\end{figure*}
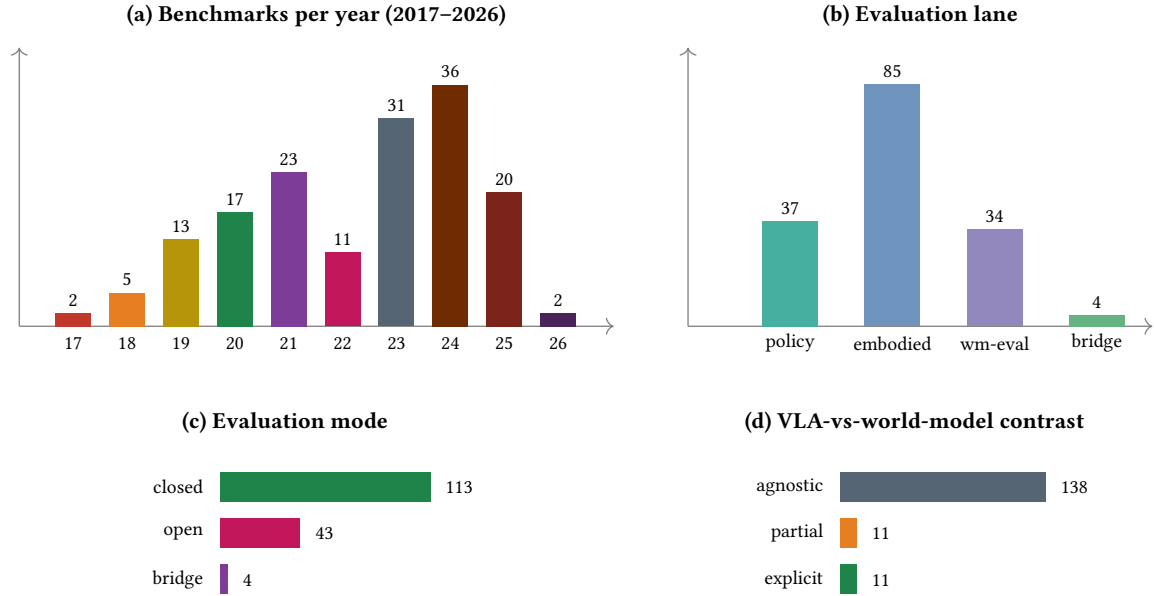

%% file: figures/fig_lane_trend.tex
\begin{figure}[t]
\centering
\definecolor{cclosed}{HTML}{1E8449}\definecolor{copen}{HTML}{C2185B}%
\resizebox{\textwidth}{!}{%
\begin{tikzpicture}
  \foreach \i in {5,10,15,20}{\node[font=\scriptsize,left=1pt,text=black!55] at (0.5,{\i*0.11}) {\i};
    \draw[black!12] (0.5,{\i*0.11})--(8.4,{\i*0.11});}
  \draw[->,black!45] (0.5,0)--(0.5,2.95); \draw[->,black!45] (0.5,0)--(8.6,0);
  \foreach [count=\g] \yr/\c/\o in {2019/10/3,2020/16/1,2021/19/4,2022/8/3,2023/24/7,2024/24/11,2025/6/13}{
    \pgfmathsetmacro\x{\g*1.12+0.15}
    \fill[cclosed] ({\x-0.34},0) rectangle ({\x-0.02},{\c*0.11});
    \fill[copen]   ({\x+0.02},0) rectangle ({\x+0.34},{\o*0.11});
    \node[font=\tiny,text=black!65,above=-1.5pt] at ({\x-0.18},{\c*0.11}) {\c};
    \node[font=\tiny,text=black!65,above=-1.5pt] at ({\x+0.18},{\o*0.11}) {\o};
    \node[font=\scriptsize] at (\x,-0.25) {\yr};}
  \fill[cclosed] (6.3,3.06) rectangle (6.53,3.24);
  \node[font=\scriptsize,right=1pt] at (6.53,3.15) {closed-loop};
  \fill[copen] (6.3,2.76) rectangle (6.53,2.94);
  \node[font=\scriptsize,right=1pt] at (6.53,2.85) {open-loop};
  \node[font=\footnotesize\bfseries] at (3.2,3.25) {Benchmarks by evaluation mode, per year};
\end{tikzpicture}}
\caption{\textbf{The world-model-evaluation wave is recent.} Per-year counts of closed-loop
task-success benchmarks (green) and open-loop world-model / video-generation benchmarks (magenta) in
the catalogue. Closed-loop suites dominate through 2024; in \textbf{2025 open-loop evaluation
overtakes them} (13 vs 6) as the world-model-generation literature arrives. The four
prediction-to-action bridges are newer still, all appearing in 2024 or later. Years are the
benchmarks' first-release years (2026 is partial and omitted); counts are the tallies of
Table~\ref{tab:landscape}.}
\Description{A grouped bar chart of benchmarks by evaluation mode per year, 2019 to 2025. Closed-loop
(green): 10, 16, 19, 8, 24, 24, 6. Open-loop (magenta): 3, 1, 4, 3, 7, 11, 13. Closed-loop dominates
until 2024, then open-loop overtakes it in 2025.}
\label{fig:lane_trend}
\end{figure}
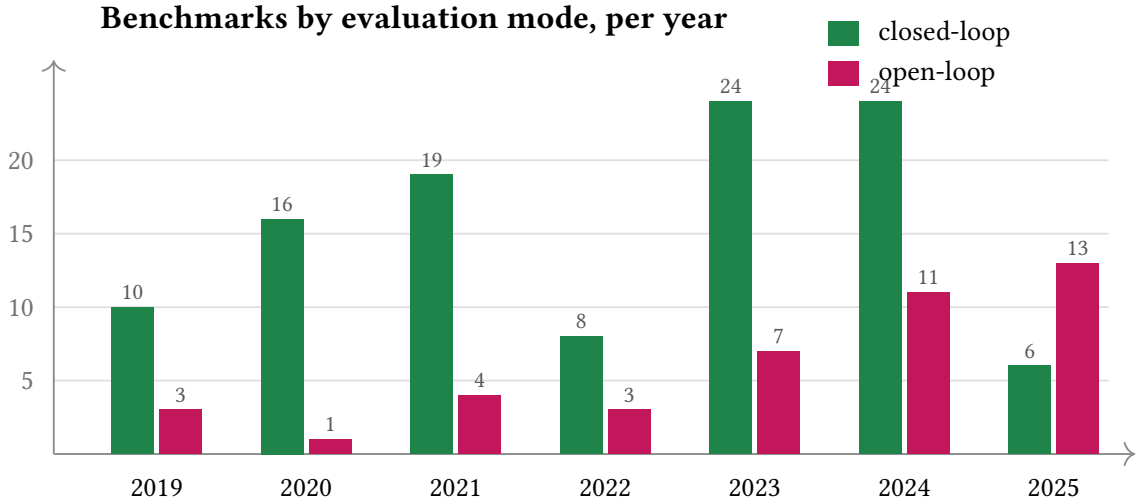

%% file: sections/3_taxonomy.tex
\section{A taxonomy of robot-evaluation benchmarks}
\label{sec:taxonomy}

Figure~\ref{fig:taxonomy_main} organises 86 representative benchmarks under a single root and four
evaluation lanes. The lanes are ordered by how close their scoring sits to executed behaviour: from
closed-loop policy suites (Section~\ref{sec:taxonomy} \S3.1) and embodied-agent benchmarks (\S3.2),
through open-loop world-model evaluation (\S3.3), to the prediction-to-action bridges (\S3.4) that
alone turn prediction into action. Within each lane, benchmarks are grouped by the capability they
stress. Table~\ref{tab:compare_main} gives a detailed comparison of a representative subset on
year, venue, evaluation mode, and contrast, and Table~\ref{tab:matrix} places one exemplar per lane
against a common set of capability dimensions; neither lane's exemplar builds the VLA-versus-world-model
contrast that Section~\ref{sec:gap} isolates.

\input{figures/fig_taxonomy_main}
\input{tables/tab_compare_main}
\input{tables/tab_capability_matrix}

The two galleries make the corpus concrete. Figure~\ref{fig:subject_gallery} collects the
benchmarks' own \emph{subject} figures, the task and scene images that show what each benchmark
depicts, grouped by lane rather than by embodiment. Figure~\ref{fig:results_gallery} collects the
\emph{results} figures, the leaderboards, radar plots, and distributions that show what each
benchmark reports. Every panel links to its source paper and is provenance-tracked
(Tables~\ref{tab:panel_provenance},~\ref{tab:results_provenance}); benchmarks with no locatable
figure are omitted, not fabricated.

\input{figures/fig_subject_gallery}

\subsection{Policy suites}
\label{sec:lane_policy}

The policy lane (37 benchmarks) scores a policy by executing it and counting task success. Its
sub-families run from single- and dual-arm \emph{manipulation} (\S3.1.1: ARNOLD~\cite{arnold},
PerAct~\cite{peract}, FMB~\cite{fmb}, ManiSkill2~\cite{maniskill2}), through
\emph{generalization} suites that vary objects, scenes, and sim-to-real gaps (\S3.1.2:
LIBERO~\cite{libero}, THE COLOSSEUM~\cite{thecolosseum}, SIMPLER~\cite{simpler},
VIMA-Bench~\cite{vimabench}), to \emph{dexterous} hands (\S3.1.4: Bi-DexHands~\cite{bidexhands},
DexArt~\cite{dexart}, UniDexGrasp~\cite{unidexgrasp}). These suites are the backbone of policy
evaluation, and they are deliberately model-agnostic: they host whatever policy a user brings and are
careful not to privilege one architecture. That neutrality is exactly why they cannot, on their own,
answer whether a world-model policy beats a direct one. Table~\ref{tab:compare_capability} audits,
capability by capability, which benchmarks come closest to building such a contrast, and finds the
lane almost entirely agnostic.

\input{tables/tab_compare_capability}

\subsection{Embodied agents}
\label{sec:lane_embodied}

The embodied-agent lane is the largest (85 benchmarks) and the most diverse. It spans
\emph{long-horizon} household and planning tasks (\S3.2.1: ALFRED~\cite{alfred},
BEHAVIOR-1K~\cite{behavior1k}, EmbodiedBench~\cite{embodiedbench}), \emph{social and multi-agent}
settings (\S3.2.2: Habitat~3.0~\cite{habitat30}, Overcooked-AI~\cite{overcookedai},
Melting~Pot~\cite{meltingpot}), \emph{navigation} (\S3.2.3: Habitat~\cite{habitat},
HM3D~\cite{hm3d}, GOAT-Bench~\cite{goatbench}), \emph{autonomous driving} (\S3.2.4:
CARLA~\cite{carla}, Bench2Drive~\cite{bench2drive}, Waymax~\cite{waymax}), \emph{counterfactual}
reasoning (\S3.2.5: CausalWorld~\cite{causalworld}, CoPhy~\cite{cophy}, ACRE~\cite{acre}), and
\emph{locomotion} (\S3.2.6: HumanoidBench~\cite{humanoidbench}, Barkour~\cite{barkour},
Isaac~Gym~\cite{isaacgym}). Two observations matter for the survey's question. First, like the policy
lane, these suites are overwhelmingly model-agnostic. Second, the counterfactual sub-family, which
is where predictive world modelling should most plausibly help, is the thinnest: it is populated by
causal-reasoning tasks but almost never run as a closed-loop VLA-versus-world-model contrast.

\subsection{World-model evaluation}
\label{sec:lane_wmeval}

The world-model-evaluation lane (34 benchmarks) scores prediction and generation directly, with no
action taken. Its sub-families are \emph{generation quality} (\S3.3.1: VBench~\cite{vbench},
EvalCrafter~\cite{evalcrafter}, EWMBench~\cite{ewmbench}, EVA-Bench~\cite{evabench}),
\emph{physical reasoning} (\S3.3.2: IntPhys~\cite{intphys}, PhysBench~\cite{physbench},
PhyGenBench~\cite{phygenbench}, Physics-IQ~\cite{physicsiq}), \emph{counterfactual} video reasoning
(\S3.3.3: CLEVRER~\cite{clevrer}, WorldPrediction~\cite{worldprediction}), and a driving world model
(\S3.3.4: Vista~\cite{vista}). Table~\ref{tab:compare_wmeval} is a deep-dive over all 34: it records
what each scores (generation, question-answering, or prediction), the signal it uses (automatic
metric, accuracy, violation-of-expectation, or human rating), and whether it touches physical
grounding. The lane's defining limitation is uniform: every benchmark scores open-loop, and none
executes what it predicts, so a high generation score certifies visual quality but says nothing about
task success.

\input{tables/tab_compare_wmeval}
\input{figures/fig_results_gallery}

\subsection{Prediction-to-action bridges}
\label{sec:lane_bridge}

The bridge lane is the frontier, and it is tiny: four benchmarks. RoboWM-Bench~\cite{robowmbench}
tests whether a world model's predictions are \emph{executable}; World-in-World~\cite{worldinworld}
and WorldArena~\cite{worldarena} place world models in a closed loop and score task success rather
than rollout realism; WorldSimBench~\cite{worldsimbench} scores generative models both as renderers
and as controllable simulators. These are the only benchmarks that carry prediction all the way into
executed action, and they are the shaded frontier of Figure~\ref{fig:taxonomy_main}. Even so, none of
the four yet reports a per-capability, head-to-head comparison between a direct VLA policy and a
world-model policy under one protocol. The bridges prove the loop can be closed; they do not yet
close it in a way that isolates the advantage of prediction. That is the gap Section~\ref{sec:gap}
formalises.

\FloatBarrier

%% file: figures/fig_taxonomy_main.tex
\begin{figure*}[p]
\centering
\resizebox{!}{0.9\textheight}{%
\begin{forest}
  for tree={align=center},
  where level=0{font=\Large, text width=10.5em}{},
  where level=1{font=\large, text width=13em}{},
  where level=2{font=\normalsize, text width=12.5em}{},
  where level=3{font=\footnotesize, text width=33em, align=left}{},
[\textbf{Robot-evaluation}\\\textbf{benchmarks for}\\\textbf{predictive}\\\textbf{embodied AI}, fill=tx-root, text width=10.5em
  [\textbf{\S3.1 Policy suites}, fill=cat-vla
    [\textbf{\S3.1.1 Manipulation}, fill=cat-vla
      [{\textbf{ARNOLD}: language-grounded continuous-state 3D manipulation~\citep{arnold}\\\textbf{BiGym}: mobile bimanual manipulation~\citep{bigym}\\\textbf{FMB}: real-world functional multi-step assembly manipulat\ldots~\citep{fmb}\\\textbf{Franka Kitchen}: multi-task kitchen manipulation~\citep{frankakitchen}\\\textbf{FurnitureBench}: real-world furniture assembly~\citep{furniturebench}\\\textbf{GRUtopia}: city-scale embodied navigation and manipulation~\citep{grutopia}\\\textbf{ManiSkill2}: generalizable skills~\citep{maniskill2}\\\textbf{PerAct}: language-conditioned 6-DoF single-arm manipulation~\citep{peract}}, fill=leaf-vla]
    ]
    [\textbf{\S3.1.2 Generalization}, fill=cat-vla
      [{\textbf{GemBench}: generalization levels~\citep{gembench}\\\textbf{LIBERO}: short-horizon + transfer~\citep{libero}\\\textbf{RoboArena}: real-world generalization~\citep{roboarena}\\\textbf{SIMPLER / SimplerEnv}: generalization (real vs sim)~\citep{simpler}\\\textbf{THE COLOSSEUM}: generalization / robustness~\citep{thecolosseum}\\\textbf{VIMA-Bench}: multimodal-prompt generalization~\citep{vimabench}}, fill=leaf-vla]
    ]
    [\textbf{\S3.1.3 Other}, fill=cat-vla
      [{\textbf{CortexBench}: pretrained visual representations for embodied cont\ldots~\citep{cortexbench}\\\textbf{GenManip}: LLM-scene tasks~\citep{genmanip}\\\textbf{HandoverSim}: human-to-robot object handover~\citep{handoversim}\\\textbf{Meta-World}: multi-task / meta-RL~\citep{metaworld}\\\textbf{RoboHive}: unified robot-learning environments~\citep{robohive}}, fill=leaf-vla]
    ]
    [\textbf{\S3.1.4 Dexterous}, fill=cat-vla
      [{\textbf{Bi-DexHands}: bimanual dexterous manipulation~\citep{bidexhands}\\\textbf{DexArt}: dexterous articulated-object manipulation~\citep{dexart}\\\textbf{UniDexGrasp}: universal dexterous grasping~\citep{unidexgrasp}\\\textbf{DeformableGym}: 3D deformable-object grasping}, fill=leaf-vla]
    ]
  ]
  [\textbf{\S3.2 Embodied agents}, fill=cat-code
    [\textbf{\S3.2.1 Long-horizon}, fill=cat-code
      [{\textbf{Alexa Arena}: task-planning~\citep{alexaarena}\\\textbf{ALFRED}: long-horizon, partial-obs~\citep{alfred}\\\textbf{BEHAVIOR-1K}: long-horizon household~\citep{behavior1k}\\\textbf{CoELA (C-WAH/TDW-MAT)}: long-horizon~\citep{coela}\\\textbf{DialFRED}: task-planning~\citep{dialfred}\\\textbf{EgoPlan-Bench}: task-planning~\citep{egoplanbench}\\\textbf{EmbodiedBench}: long-horizon~\citep{embodiedbench}}, fill=leaf-code]
    ]
    [\textbf{\S3.2.2 Social \& multi-agent}, fill=cat-code
      [{\textbf{CrowdNav}: social~\citep{crowdnav}\\\textbf{Habitat 3.0}: social / multi-agent~\citep{habitat30}\\\textbf{HuNavSim}: social~\citep{hunavsim}\\\textbf{JRDB}: social~\citep{jrdb}\\\textbf{JRDB-Act}: social~\citep{jrdbact}\\\textbf{Melting Pot}: social~\citep{meltingpot}\\\textbf{Overcooked-AI}: social~\citep{overcookedai}}, fill=leaf-code]
    ]
    [\textbf{\S3.2.3 Navigation}, fill=cat-code
      [{\textbf{Gibson Env}: navigation~\citep{gibsonenv}\\\textbf{GOAT-Bench}: multi-modal lifelong navigation~\citep{goatbench}\\\textbf{Habitat}: navigation~\citep{habitat}\\\textbf{Habitat 2.0}: rearrangement~\citep{habitat20}\\\textbf{HM3D}: navigation~\citep{hm3d}\\\textbf{iGibson}: interactive navigation~\citep{igibson}\\\textbf{Matterport3D}: navigation~\citep{matterport3d}}, fill=leaf-code]
    ]
    [\textbf{\S3.2.4 Autonomous driving}, fill=cat-code
      [{\textbf{Bench2Drive}: driving~\citep{bench2drive}\\\textbf{CARLA}: driving~\citep{carla}\\\textbf{DriveArena}: driving~\citep{drivearena}\\\textbf{MetaDrive}: driving~\citep{metadrive}\\\textbf{NAVSIM}: driving~\citep{navsim}\\\textbf{nuScenes}: driving~\citep{nuscenes}\\\textbf{Waymax}: driving~\citep{waymax}}, fill=leaf-code]
    ]
    [\textbf{\S3.2.5 Counterfactual}, fill=cat-code
      [{\textbf{ACRE}: counterfactual~\citep{acre}\\\textbf{CausalCity}: counterfactual~\citep{causalcity}\\\textbf{CausalVQA}: counterfactual~\citep{causalvqa}\\\textbf{CausalWorld}: counterfactual~\citep{causalworld}\\\textbf{CoPhy}: counterfactual~\citep{cophy}\\\textbf{CRAFT}: counterfactual~\citep{craft}\\\textbf{Filtered-CoPhy}: counterfactual~\citep{filteredcophy}}, fill=leaf-code]
    ]
    [\textbf{\S3.2.6 Locomotion}, fill=cat-code
      [{\textbf{Barkour}: legged locomotion~\citep{barkour}\\\textbf{Brax}: locomotion~\citep{brax}\\\textbf{DERL}: locomotion~\citep{derl}\\\textbf{HumanoidBench}: humanoid locomotion~\citep{humanoidbench}\\\textbf{Isaac Gym}: locomotion~\citep{isaacgym}\\\textbf{Legged Gym}: legged locomotion~\citep{leggedgym}}, fill=leaf-code]
    ]
  ]
  [\textbf{\S3.3 World-model eval}, fill=cat-skill
    [\textbf{\S3.3.1 Generation quality}, fill=cat-skill
      [{\textbf{CATER}: compositional-action temporal reasoning~\citep{cater}\\\textbf{DEVIL}: content-dynamics quality of T2V models~\citep{devil}\\\textbf{EVA-Bench}: embodied video anticipation~\citep{evabench}\\\textbf{EvalCrafter}: generation quality (17-metric suite)~\citep{evalcrafter}\\\textbf{EWMBench}: scene/motion/semantic quality~\citep{ewmbench}\\\textbf{FETV}: fine-grained text-to-video quality~\citep{fetv}\\\textbf{IPV-Bench}: impossible-video generation + understanding~\citep{ipvbench}\\\textbf{StoryBench}: continuous story visualization~\citep{storybench}}, fill=leaf-skill]
    ]
    [\textbf{\S3.3.2 Physical reasoning}, fill=cat-skill
      [{\textbf{ContPhy}: continuum (soft-body/fluid) physical reasoning~\citep{contphy}\\\textbf{GRASP}: grounding + intuitive physics in video MLLMs~\citep{grasp}\\\textbf{IntPhys}: intuitive physics (violation-of-expectation)~\citep{intphys}\\\textbf{IntPhys 2}: intuitive physics in complex scenes~\citep{intphys2}\\\textbf{PHYBench}: physical perception and reasoning~\citep{phybench}\\\textbf{PhyGenBench}: physical-commonsense correctness in video gen~\citep{phygenbench}\\\textbf{PhysBench}: physical-world understanding for VLMs~\citep{physbench}\\\textbf{Physics-IQ}: physical realism~\citep{physicsiq}}, fill=leaf-skill]
    ]
    [\textbf{\S3.3.3 Counterfactual}, fill=cat-skill
      [{\textbf{CLEVRER}: physical/causal video reasoning (counterfactual)~\citep{clevrer}\\\textbf{WorldPrediction}: counterfactual action recognition~\citep{worldprediction}}, fill=leaf-skill]
    ]
    [\textbf{\S3.3.4 Autonomous driving}, fill=cat-skill
      [{\textbf{Vista}: driving world model~\citep{vista}}, fill=leaf-skill]
    ]
  ]
  [\textbf{\S3.4 Bridges}, fill=secE
    [\textbf{\S3.4.1 Other}, fill=secE
      [{\textbf{RoboWM-Bench}: prediction executability~\citep{robowmbench}\\\textbf{World-in-World}: closed-loop task success of WMs~\citep{worldinworld}\\\textbf{WorldArena}: closed-loop WM arena~\citep{worldarena}}, fill=secEl]
    ]
  ]
]
\end{forest}}
\caption{\textbf{A taxonomy of robot-evaluation benchmarks for predictive embodied intelligence} (86 representative benchmarks of the 160 in Table~\ref{tab:landscape}). Each sub-family is one cell listing its benchmarks, each with a short statement of what it measures. Benchmarks are grouped by evaluation lane (\S3.1--\S3.4) and then by the capability they stress. The lanes run from closed-loop task-success suites to the \emph{bridge} frontier (shaded), the only lane that turns prediction into executed action, which is the survey's novelty axis. Full per-benchmark detail: Tables~\ref{tab:compare_main}, \ref{tab:landscape}.}
\Description{A tree diagram organising representative robot-evaluation benchmarks under the root ``robot-evaluation benchmarks for predictive embodied AI'' into four evaluation lanes (policy and manipulation suites; embodied navigation, planning and social; world-model and video-generation evaluation; and prediction-to-action bridges), each split into capability clusters whose cells list benchmarks with what each measures.}
\label{fig:taxonomy_main}
\end{figure*}
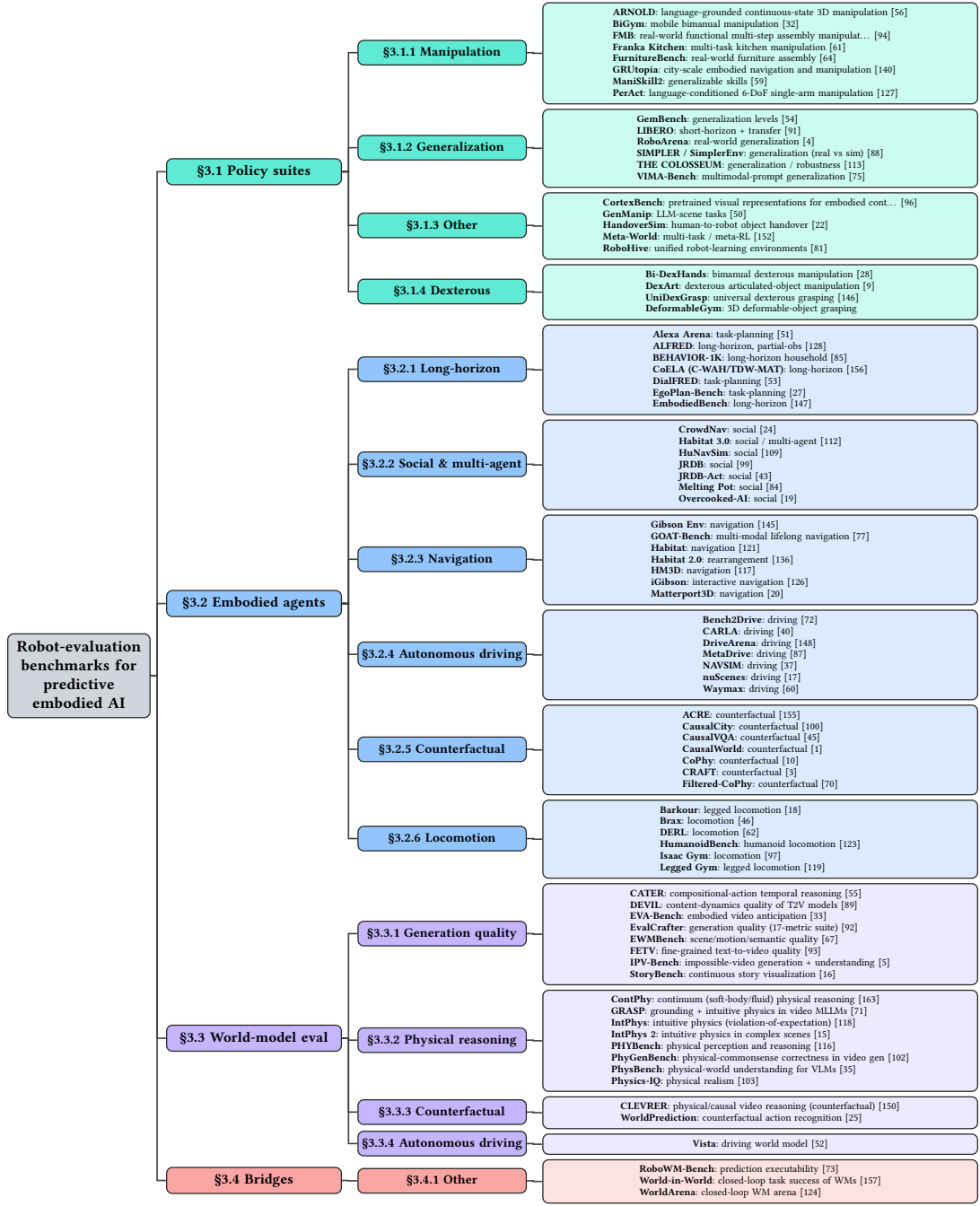

%% file: tables/tab_compare_main.tex
{\footnotesize
\setlength{\tabcolsep}{5pt}
\renewcommand{\arraystretch}{1.15}
\begin{longtable}{@{}p{0.22\textwidth} c l c >{\raggedright\arraybackslash}p{0.30\textwidth} c@{}}
\caption{\textbf{Detailed comparison of a representative 31 of the 160 benchmarks} in the landscape (Table~\ref{tab:landscape}), grouped by evaluation lane. \textbf{Eval}: \dopen\ open-loop prediction/generation quality, \dclosed\ closed-loop task success, \dbridge\ prediction-to-action bridge. \textbf{VLA vs WM}: \yy\ native direct-policy vs world-model contrast, \pp\ partial, \dm\ model-agnostic. Venues are web-verified; ``arXiv'' marks preprint-only works.}\label{tab:compare_main}\\
\toprule
\textbf{Benchmark} & \textbf{Year} & \textbf{Venue} & \textbf{Eval} & \textbf{Capability} & \textbf{VLA vs WM}\\
\midrule
\endfirsthead
\multicolumn{6}{@{}l}{\footnotesize\emph{Table~\ref{tab:compare_main} (continued)}}\\
\toprule
\textbf{Benchmark} & \textbf{Year} & \textbf{Venue} & \textbf{Eval} & \textbf{Capability} & \textbf{VLA vs WM}\\
\midrule
\endhead
\bottomrule
\endlastfoot
\multicolumn{6}{@{}l}{\rule{0pt}{2.6ex}\textbf{\textsf{\S3.1 Policy / manipulation suites}} \hfill \dclosed}\\[1pt]
\textbf{CALVIN}~\cite{calvin} & 2021 & RA-L & \dclosed & long-horizon language & \dm\\
\textbf{LIBERO}~\cite{libero} & 2023 & NeurIPS D\&B & \dclosed & short-horizon + transfer & \dm\\
\textbf{Meta-World}~\cite{metaworld} & 2020 & CoRL & \dclosed & multi-task / meta-RL & \dm\\
\textbf{ManiSkill2}~\cite{maniskill2} & 2023 & ICLR & \dclosed & generalizable skills & \dm\\
\textbf{SIMPLER}~\cite{simpler} & 2024 & CoRL & \dclosed & generalization (real vs sim) & \dm\\
\textbf{THE COLOSSEUM}~\cite{thecolosseum} & 2024 & RSS & \dclosed & generalization / robustness & \dm\\
\textbf{RoboCasa}~\cite{robocasa} & 2024 & RSS & \dclosed & long-horizon (data scaling) & \dm\\
\textbf{RoboArena}~\cite{roboarena} & 2025 & CoRL & \dclosed & real-world generalization & \dm\\
\textbf{GemBench}~\cite{gembench} & 2024 & ICRA & \dclosed & generalization levels & \dm\\
\textbf{VLABench}~\cite{vlabench} & 2025 & ICCV & \dclosed & long-horizon reasoning & \dm\\
\midrule
\multicolumn{6}{@{}l}{\rule{0pt}{2.6ex}\textbf{\textsf{\S3.2 Embodied navigation, planning \& social}} \hfill \dclosed}\\[1pt]
\textbf{Habitat 3.0}~\cite{habitat30} & 2023 & ICLR & \dclosed & social / multi-agent & \dm\\
\textbf{BEHAVIOR-1K}~\cite{behavior1k} & 2024 & CoRL & \dclosed & long-horizon household & \dm\\
\textbf{ALFRED}~\cite{alfred} & 2019 & CVPR & \dclosed & long-horizon, partial-obs & \dm\\
\textbf{TEACh}~\cite{teach} & 2021 & AAAI & \dclosed & hidden-state (dialog) & \dm\\
\textbf{RoboTHOR}~\cite{robothor} & 2020 & CVPR & \dclosed & navigation / sim-to-real & \dm\\
\textbf{SocNavBench}~\cite{socnavbench} & 2021 & ACM THRI & \dclosed & social navigation & \dm\\
\textbf{LIBERO-Mem}~\cite{liberomem} & 2025 & AAAI & \dclosed & hidden-state / occlusion & \yy\\
\textbf{VIMA-Bench}~\cite{vimabench} & 2023 & ICML & \dclosed & multimodal-prompt generalization & \dm\\
\midrule
\multicolumn{6}{@{}l}{\rule{0pt}{2.6ex}\textbf{\textsf{\S3.3 World-model \& video-generation evaluation}} \hfill \dopen}\\[1pt]
\textbf{WorldModelBench}~\cite{worldmodelbench} & 2025 & arXiv & \dopen & video-WM prediction quality & \dm\\
\textbf{Physics-IQ}~\cite{physicsiq} & 2025 & WACV & \dopen & physical realism & \dm\\
\textbf{VBench-2.0}~\cite{vbench20} & 2025 & arXiv & \dopen & generation faithfulness & \dm\\
\textbf{WorldScore}~\cite{worldscore} & 2025 & ICCV & \dopen & world-gen quality & \dm\\
\textbf{EWMBench}~\cite{ewmbench} & 2025 & arXiv & \dopen & scene/motion/semantic quality & \dm\\
\textbf{EVA-Bench}~\cite{evabench} & 2024 & ICML & \dopen & embodied video anticipation & \dm\\
\textbf{WorldPrediction}~\cite{worldprediction} & 2025 & arXiv & \dopen & counterfactual action recognition & \dm\\
\textbf{Physion}~\cite{physion} & 2021 & NeurIPS D\&B & \dopen & physical prediction from vision & \dm\\
\textbf{Physion++}~\cite{physionpp} & 2023 & NeurIPS D\&B & \dopen & physical prediction with online property inference & \dm\\
\midrule
\multicolumn{6}{@{}l}{\rule{0pt}{2.6ex}\textbf{\textsf{\S3.4 Bridges (prediction to action)}} \hfill \dbridge}\\[1pt]
\textbf{WorldSimBench}~\cite{worldsimbench} & 2024 & ICML & \dbridge & video-to-action consistency & \pp\\
\textbf{World-in-World}~\cite{worldinworld} & 2025 & ICLR & \dbridge & closed-loop task success of WMs & \pp\\
\textbf{RoboWM-Bench}~\cite{robowmbench} & 2026 & arXiv & \dbridge & prediction executability & \pp\\
\textbf{WorldArena}~\cite{worldarena} & 2026 & arXiv & \dbridge & closed-loop WM arena & \pp\\
\end{longtable}
}

%% file: tables/tab_capability_matrix.tex
\begin{table*}[t]
\centering
\footnotesize
\setlength{\tabcolsep}{5pt}
\renewcommand{\arraystretch}{1.25}
\resizebox{\textwidth}{!}{%
\begin{tabular}{@{}p{0.22\textwidth}*{4}{>{\centering\arraybackslash}p{0.15\textwidth}}@{}}
\toprule
\textbf{Dimension}
 & \textbf{LIBERO}~\cite{libero}
 & \textbf{Habitat 3.0}~\cite{habitat30}
 & \textbf{World\-Model\-Bench}~\cite{worldmodelbench}
 & \textbf{World\--in\--World}~\cite{worldinworld} \\
 & {\footnotesize\itshape policy suite} & {\footnotesize\itshape embodied nav}
 & {\footnotesize\itshape world-model eval} & {\footnotesize\itshape bridge} \\
\midrule
What it scores & task success & social-nav success & video-pred.\ quality & closed-loop WM success \\
Evaluation mode & \dclosed & \dclosed & \dopen & \dbridge \\
Executes actions in a loop? & \yy & \yy & \nn & \yy \\
Native VLA-vs-WM contrast? & \nn & \nn & \nn & \pp \\
Capability stressed & short-horizon + transfer & social / multi-agent & physical realism & plan executability \\
Per-capability slicing? & \pp & \pp & \nn & \nn \\
Advantage-of-prediction metric? & \nn & \nn & \nn & \pp \\
Real / sim / video & sim & sim & video & sim \\
Counterfactual probing? & \nn & \nn & \nn & \nn \\
Reusable leaderboard? & \yy & \yy & \yy & \pp \\
\bottomrule
\end{tabular}}
\caption{Capability matrix, one representative benchmark per evaluation lane. No lane's exemplar
builds a native VLA-vs-world-model contrast: policy and embodied suites are model-agnostic (they
\emph{host} any policy), world-model suites score generation quality with no action, and the
bridge exemplar reaches closed-loop success but reports only a partial contrast (``visual quality
is not task success''). Counterfactual probing is absent throughout. \yy~yes, \pp~partial,
\nn~no, \dm~n/a; \dopen\ open-loop, \dclosed\ closed-loop, \dbridge\ bridge.}
\label{tab:matrix}
\end{table*}

%% file: figures/fig_subject_gallery.tex
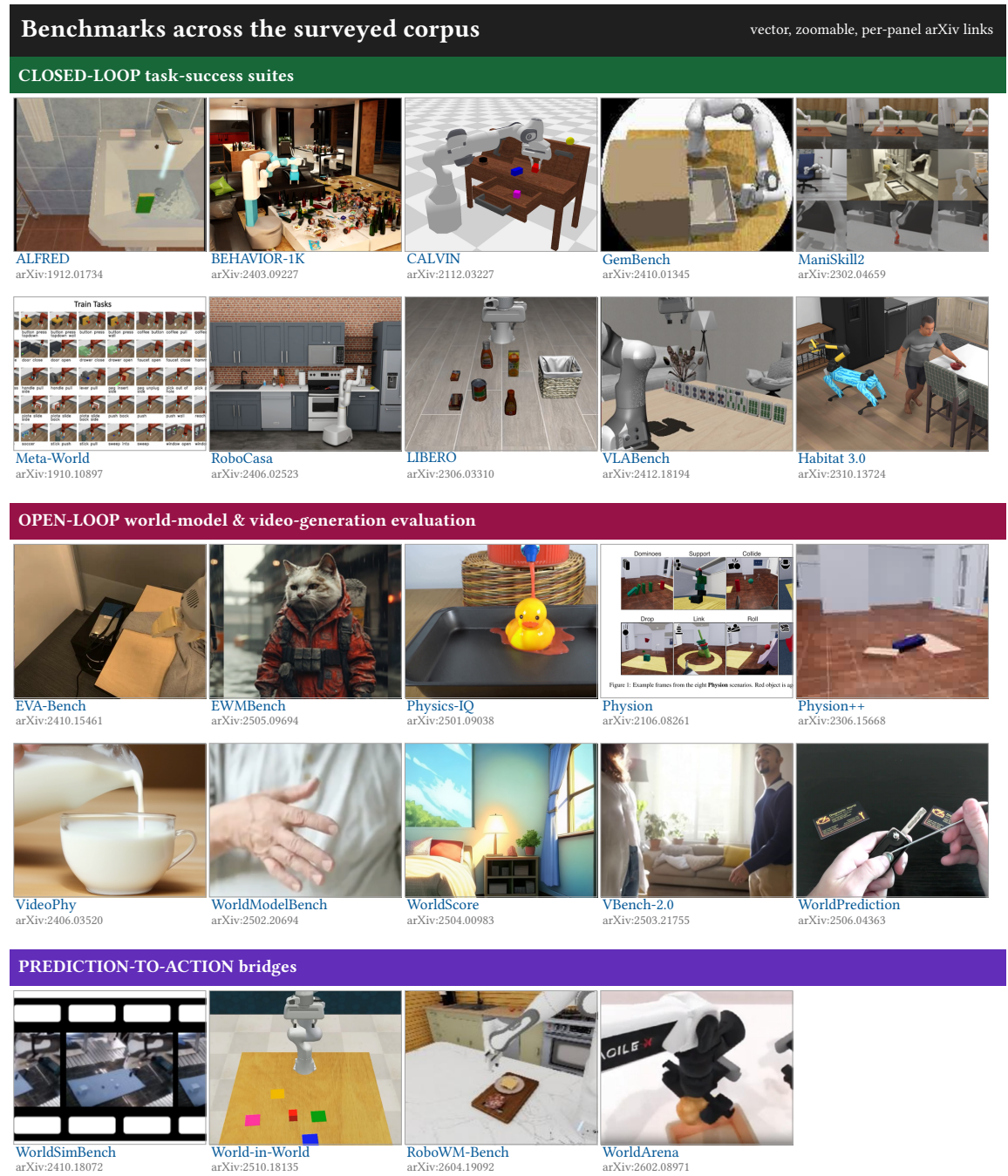
\begin{figure*}[p]
\centering
\definecolor{sgClosed}{HTML}{1E8449}\definecolor{sgWM}{HTML}{C2185B}\definecolor{sgBridge}{HTML}{7C3AED}%
\resizebox{\textwidth}{!}{%
\begin{tikzpicture}[x=1cm,y=1cm]
\fill[black!88] (0,0) rectangle (13.80,-0.72);
\node[anchor=west,text=white,font=\large\bfseries,inner sep=0] at (0.16,-0.36) {Benchmarks across the surveyed corpus};
\node[anchor=east,text=white!70,font=\scriptsize,inner sep=0] at (13.64,-0.36) {vector, zoomable, per-panel arXiv links};
\fill[sgClosed!78!black] (0,-0.720) rectangle (13.80,-1.220);
\node[anchor=west,text=white,font=\footnotesize\bfseries,inner sep=0] at (0.12,-0.970) {CLOSED-LOOP task-success suites};
\node[anchor=north west,inner sep=0] at (0.060,-1.290) {\href{https://arxiv.org/abs/1912.01734}{\includegraphics[width=2.660cm,height=2.128cm]{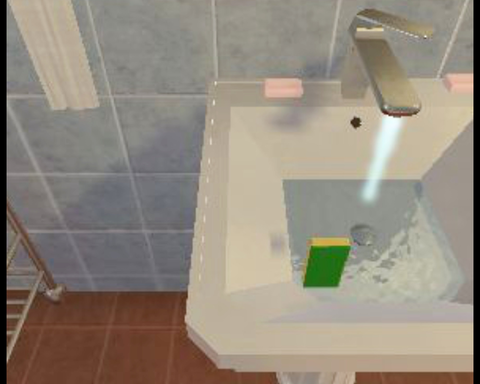}}};
\draw[black!35,line width=0.35pt] (0.060,-1.290) rectangle (2.720,-3.418);
\node[anchor=north west,text=ACMDarkBlue,font=\scriptsize,inner sep=0] at (0.080,-3.438) {\href{https://arxiv.org/abs/1912.01734}{ALFRED}};
\node[anchor=north west,text=black!55,font=\tiny,inner sep=0] at (0.080,-3.668) {\href{https://arxiv.org/abs/1912.01734}{arXiv:1912.01734}};
\node[anchor=north west,inner sep=0] at (2.770,-1.290) {\href{https://arxiv.org/abs/2403.09227}{\includegraphics[width=2.660cm,height=2.128cm]{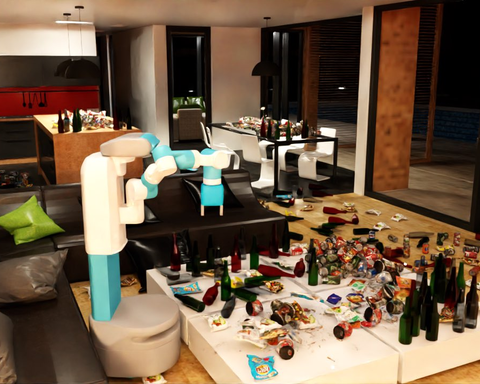}}};
\draw[black!35,line width=0.35pt] (2.770,-1.290) rectangle (5.430,-3.418);
\node[anchor=north west,text=ACMDarkBlue,font=\scriptsize,inner sep=0] at (2.790,-3.438) {\href{https://arxiv.org/abs/2403.09227}{BEHAVIOR-1K}};
\node[anchor=north west,text=black!55,font=\tiny,inner sep=0] at (2.790,-3.668) {\href{https://arxiv.org/abs/2403.09227}{arXiv:2403.09227}};
\node[anchor=north west,inner sep=0] at (5.480,-1.290) {\href{https://arxiv.org/abs/2112.03227}{\includegraphics[width=2.660cm,height=2.128cm]{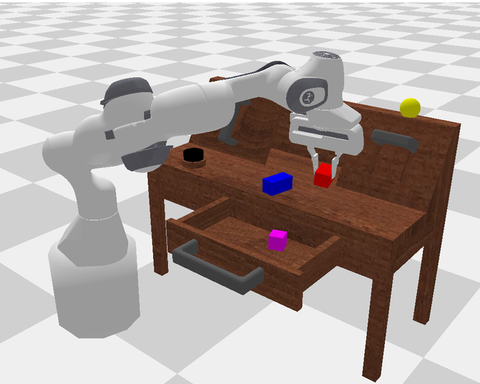}}};
\draw[black!35,line width=0.35pt] (5.480,-1.290) rectangle (8.140,-3.418);
\node[anchor=north west,text=ACMDarkBlue,font=\scriptsize,inner sep=0] at (5.500,-3.438) {\href{https://arxiv.org/abs/2112.03227}{CALVIN}};
\node[anchor=north west,text=black!55,font=\tiny,inner sep=0] at (5.500,-3.668) {\href{https://arxiv.org/abs/2112.03227}{arXiv:2112.03227}};
\node[anchor=north west,inner sep=0] at (8.190,-1.290) {\href{https://arxiv.org/abs/2410.01345}{\includegraphics[width=2.660cm,height=2.128cm]{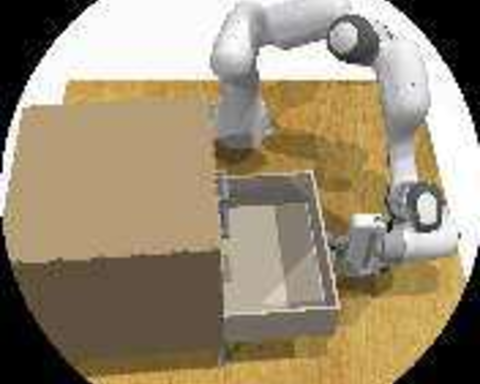}}};
\draw[black!35,line width=0.35pt] (8.190,-1.290) rectangle (10.850,-3.418);
\node[anchor=north west,text=ACMDarkBlue,font=\scriptsize,inner sep=0] at (8.210,-3.438) {\href{https://arxiv.org/abs/2410.01345}{GemBench}};
\node[anchor=north west,text=black!55,font=\tiny,inner sep=0] at (8.210,-3.668) {\href{https://arxiv.org/abs/2410.01345}{arXiv:2410.01345}};
\node[anchor=north west,inner sep=0] at (10.900,-1.290) {\href{https://arxiv.org/abs/2302.04659}{\includegraphics[width=2.660cm,height=2.128cm]{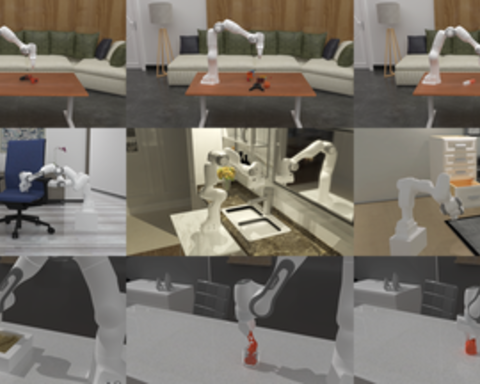}}};
\draw[black!35,line width=0.35pt] (10.900,-1.290) rectangle (13.560,-3.418);
\node[anchor=north west,text=ACMDarkBlue,font=\scriptsize,inner sep=0] at (10.920,-3.438) {\href{https://arxiv.org/abs/2302.04659}{ManiSkill2}};
\node[anchor=north west,text=black!55,font=\tiny,inner sep=0] at (10.920,-3.668) {\href{https://arxiv.org/abs/2302.04659}{arXiv:2302.04659}};
\node[anchor=north west,inner sep=0] at (0.060,-4.048) {\href{https://arxiv.org/abs/1910.10897}{\includegraphics[width=2.660cm,height=2.128cm]{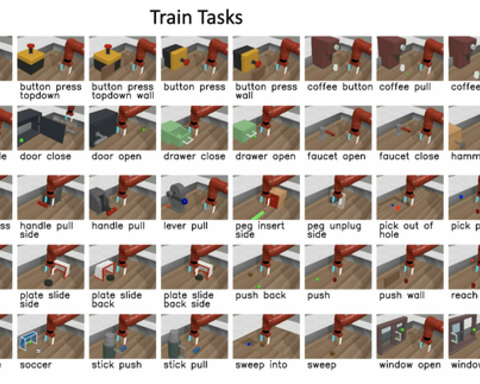}}};
\draw[black!35,line width=0.35pt] (0.060,-4.048) rectangle (2.720,-6.176);
\node[anchor=north west,text=ACMDarkBlue,font=\scriptsize,inner sep=0] at (0.080,-6.196) {\href{https://arxiv.org/abs/1910.10897}{Meta-World}};
\node[anchor=north west,text=black!55,font=\tiny,inner sep=0] at (0.080,-6.426) {\href{https://arxiv.org/abs/1910.10897}{arXiv:1910.10897}};
\node[anchor=north west,inner sep=0] at (2.770,-4.048) {\href{https://arxiv.org/abs/2406.02523}{\includegraphics[width=2.660cm,height=2.128cm]{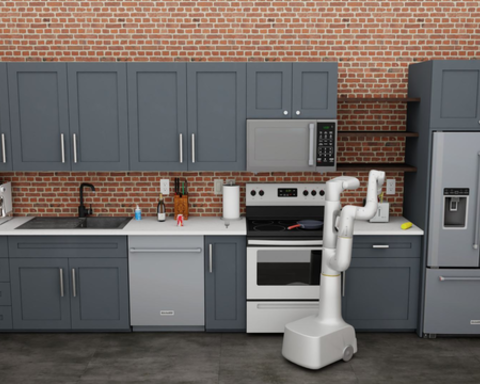}}};
\draw[black!35,line width=0.35pt] (2.770,-4.048) rectangle (5.430,-6.176);
\node[anchor=north west,text=ACMDarkBlue,font=\scriptsize,inner sep=0] at (2.790,-6.196) {\href{https://arxiv.org/abs/2406.02523}{RoboCasa}};
\node[anchor=north west,text=black!55,font=\tiny,inner sep=0] at (2.790,-6.426) {\href{https://arxiv.org/abs/2406.02523}{arXiv:2406.02523}};
\node[anchor=north west,inner sep=0] at (5.480,-4.048) {\href{https://arxiv.org/abs/2306.03310}{\includegraphics[width=2.660cm,height=2.128cm]{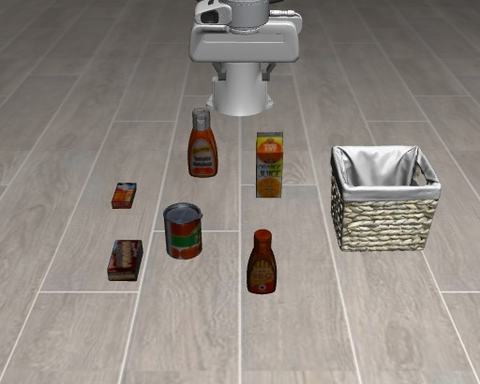}}};
\draw[black!35,line width=0.35pt] (5.480,-4.048) rectangle (8.140,-6.176);
\node[anchor=north west,text=ACMDarkBlue,font=\scriptsize,inner sep=0] at (5.500,-6.196) {\href{https://arxiv.org/abs/2306.03310}{LIBERO}};
\node[anchor=north west,text=black!55,font=\tiny,inner sep=0] at (5.500,-6.426) {\href{https://arxiv.org/abs/2306.03310}{arXiv:2306.03310}};
\node[anchor=north west,inner sep=0] at (8.190,-4.048) {\href{https://arxiv.org/abs/2412.18194}{\includegraphics[width=2.660cm,height=2.128cm]{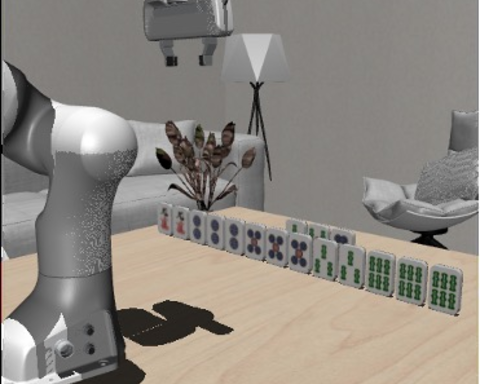}}};
\draw[black!35,line width=0.35pt] (8.190,-4.048) rectangle (10.850,-6.176);
\node[anchor=north west,text=ACMDarkBlue,font=\scriptsize,inner sep=0] at (8.210,-6.196) {\href{https://arxiv.org/abs/2412.18194}{VLABench}};
\node[anchor=north west,text=black!55,font=\tiny,inner sep=0] at (8.210,-6.426) {\href{https://arxiv.org/abs/2412.18194}{arXiv:2412.18194}};
\node[anchor=north west,inner sep=0] at (10.900,-4.048) {\href{https://arxiv.org/abs/2310.13724}{\includegraphics[width=2.660cm,height=2.128cm]{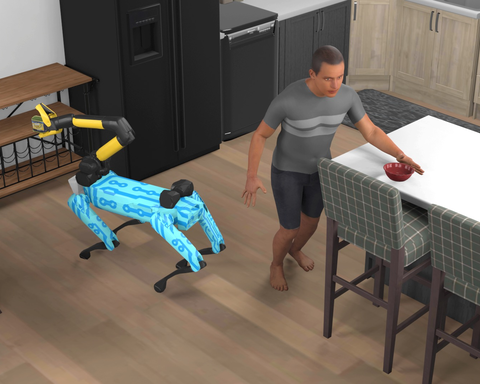}}};
\draw[black!35,line width=0.35pt] (10.900,-4.048) rectangle (13.560,-6.176);
\node[anchor=north west,text=ACMDarkBlue,font=\scriptsize,inner sep=0] at (10.920,-6.196) {\href{https://arxiv.org/abs/2310.13724}{Habitat 3.0}};
\node[anchor=north west,text=black!55,font=\tiny,inner sep=0] at (10.920,-6.426) {\href{https://arxiv.org/abs/2310.13724}{arXiv:2310.13724}};
\fill[sgWM!78!black] (0,-6.906) rectangle (13.80,-7.406);
\node[anchor=west,text=white,font=\footnotesize\bfseries,inner sep=0] at (0.12,-7.156) {OPEN-LOOP world-model \& video-generation evaluation};
\node[anchor=north west,inner sep=0] at (0.060,-7.476) {\href{https://arxiv.org/abs/2410.15461}{\includegraphics[width=2.660cm,height=2.128cm]{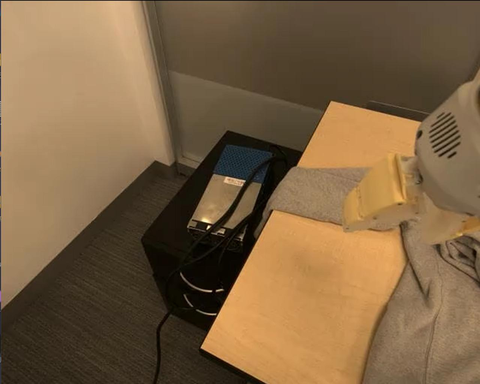}}};
\draw[black!35,line width=0.35pt] (0.060,-7.476) rectangle (2.720,-9.604);
\node[anchor=north west,text=ACMDarkBlue,font=\scriptsize,inner sep=0] at (0.080,-9.624) {\href{https://arxiv.org/abs/2410.15461}{EVA-Bench}};
\node[anchor=north west,text=black!55,font=\tiny,inner sep=0] at (0.080,-9.854) {\href{https://arxiv.org/abs/2410.15461}{arXiv:2410.15461}};
\node[anchor=north west,inner sep=0] at (2.770,-7.476) {\href{https://arxiv.org/abs/2505.09694}{\includegraphics[width=2.660cm,height=2.128cm]{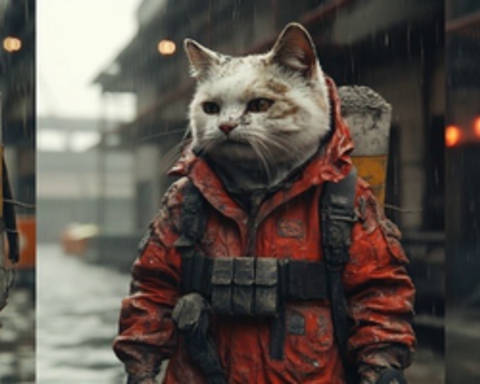}}};
\draw[black!35,line width=0.35pt] (2.770,-7.476) rectangle (5.430,-9.604);
\node[anchor=north west,text=ACMDarkBlue,font=\scriptsize,inner sep=0] at (2.790,-9.624) {\href{https://arxiv.org/abs/2505.09694}{EWMBench}};
\node[anchor=north west,text=black!55,font=\tiny,inner sep=0] at (2.790,-9.854) {\href{https://arxiv.org/abs/2505.09694}{arXiv:2505.09694}};
\node[anchor=north west,inner sep=0] at (5.480,-7.476) {\href{https://arxiv.org/abs/2501.09038}{\includegraphics[width=2.660cm,height=2.128cm]{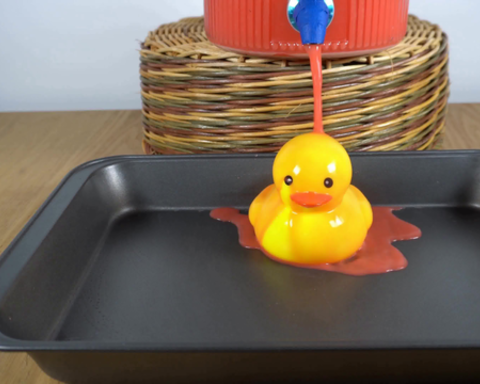}}};
\draw[black!35,line width=0.35pt] (5.480,-7.476) rectangle (8.140,-9.604);
\node[anchor=north west,text=ACMDarkBlue,font=\scriptsize,inner sep=0] at (5.500,-9.624) {\href{https://arxiv.org/abs/2501.09038}{Physics-IQ}};
\node[anchor=north west,text=black!55,font=\tiny,inner sep=0] at (5.500,-9.854) {\href{https://arxiv.org/abs/2501.09038}{arXiv:2501.09038}};
\node[anchor=north west,inner sep=0] at (8.190,-7.476) {\href{https://arxiv.org/abs/2106.08261}{\includegraphics[width=2.660cm,height=2.128cm]{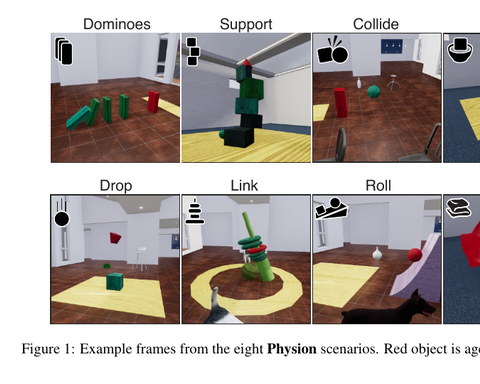}}};
\draw[black!35,line width=0.35pt] (8.190,-7.476) rectangle (10.850,-9.604);
\node[anchor=north west,text=ACMDarkBlue,font=\scriptsize,inner sep=0] at (8.210,-9.624) {\href{https://arxiv.org/abs/2106.08261}{Physion}};
\node[anchor=north west,text=black!55,font=\tiny,inner sep=0] at (8.210,-9.854) {\href{https://arxiv.org/abs/2106.08261}{arXiv:2106.08261}};
\node[anchor=north west,inner sep=0] at (10.900,-7.476) {\href{https://arxiv.org/abs/2306.15668}{\includegraphics[width=2.660cm,height=2.128cm]{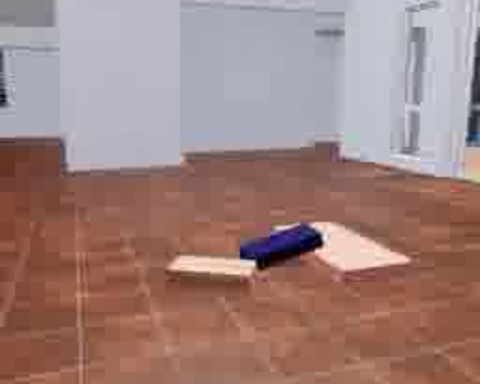}}};
\draw[black!35,line width=0.35pt] (10.900,-7.476) rectangle (13.560,-9.604);
\node[anchor=north west,text=ACMDarkBlue,font=\scriptsize,inner sep=0] at (10.920,-9.624) {\href{https://arxiv.org/abs/2306.15668}{Physion++}};
\node[anchor=north west,text=black!55,font=\tiny,inner sep=0] at (10.920,-9.854) {\href{https://arxiv.org/abs/2306.15668}{arXiv:2306.15668}};
\node[anchor=north west,inner sep=0] at (0.060,-10.234) {\href{https://arxiv.org/abs/2406.03520}{\includegraphics[width=2.660cm,height=2.128cm]{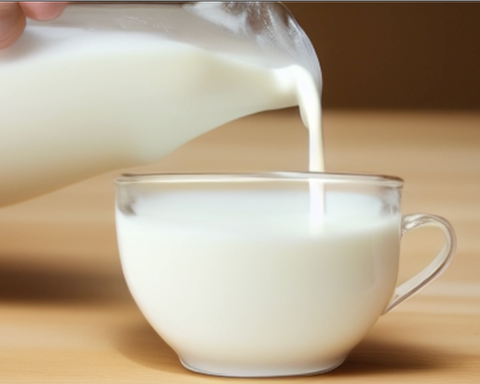}}};
\draw[black!35,line width=0.35pt] (0.060,-10.234) rectangle (2.720,-12.362);
\node[anchor=north west,text=ACMDarkBlue,font=\scriptsize,inner sep=0] at (0.080,-12.382) {\href{https://arxiv.org/abs/2406.03520}{VideoPhy}};
\node[anchor=north west,text=black!55,font=\tiny,inner sep=0] at (0.080,-12.612) {\href{https://arxiv.org/abs/2406.03520}{arXiv:2406.03520}};
\node[anchor=north west,inner sep=0] at (2.770,-10.234) {\href{https://arxiv.org/abs/2502.20694}{\includegraphics[width=2.660cm,height=2.128cm]{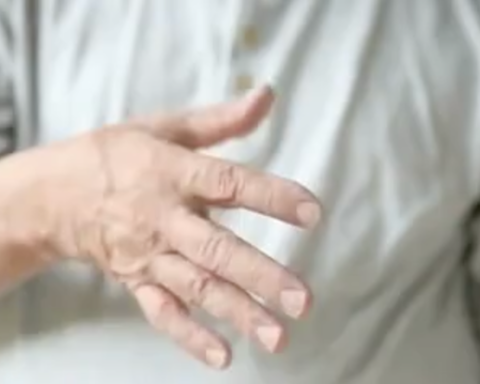}}};
\draw[black!35,line width=0.35pt] (2.770,-10.234) rectangle (5.430,-12.362);
\node[anchor=north west,text=ACMDarkBlue,font=\scriptsize,inner sep=0] at (2.790,-12.382) {\href{https://arxiv.org/abs/2502.20694}{WorldModelBench}};
\node[anchor=north west,text=black!55,font=\tiny,inner sep=0] at (2.790,-12.612) {\href{https://arxiv.org/abs/2502.20694}{arXiv:2502.20694}};
\node[anchor=north west,inner sep=0] at (5.480,-10.234) {\href{https://arxiv.org/abs/2504.00983}{\includegraphics[width=2.660cm,height=2.128cm]{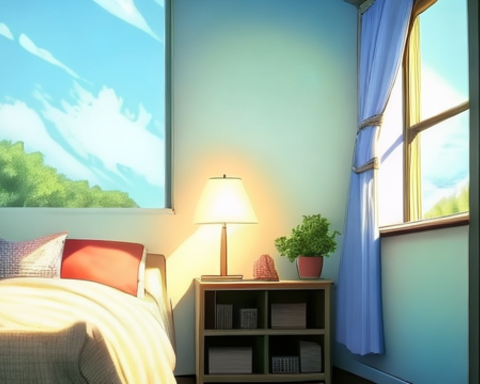}}};
\draw[black!35,line width=0.35pt] (5.480,-10.234) rectangle (8.140,-12.362);
\node[anchor=north west,text=ACMDarkBlue,font=\scriptsize,inner sep=0] at (5.500,-12.382) {\href{https://arxiv.org/abs/2504.00983}{WorldScore}};
\node[anchor=north west,text=black!55,font=\tiny,inner sep=0] at (5.500,-12.612) {\href{https://arxiv.org/abs/2504.00983}{arXiv:2504.00983}};
\node[anchor=north west,inner sep=0] at (8.190,-10.234) {\href{https://arxiv.org/abs/2503.21755}{\includegraphics[width=2.660cm,height=2.128cm]{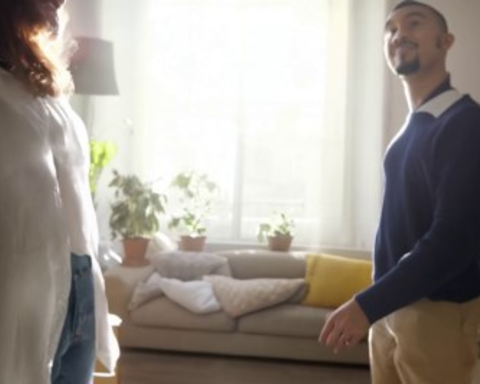}}};
\draw[black!35,line width=0.35pt] (8.190,-10.234) rectangle (10.850,-12.362);
\node[anchor=north west,text=ACMDarkBlue,font=\scriptsize,inner sep=0] at (8.210,-12.382) {\href{https://arxiv.org/abs/2503.21755}{VBench-2.0}};
\node[anchor=north west,text=black!55,font=\tiny,inner sep=0] at (8.210,-12.612) {\href{https://arxiv.org/abs/2503.21755}{arXiv:2503.21755}};
\node[anchor=north west,inner sep=0] at (10.900,-10.234) {\href{https://arxiv.org/abs/2506.04363}{\includegraphics[width=2.660cm,height=2.128cm]{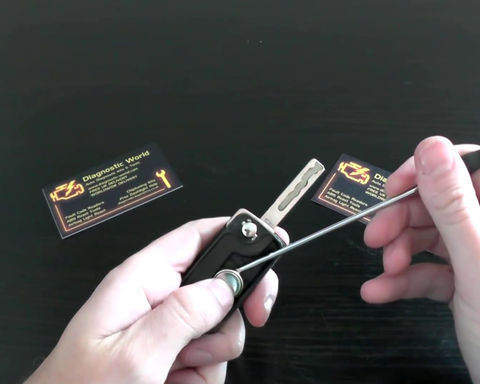}}};
\draw[black!35,line width=0.35pt] (10.900,-10.234) rectangle (13.560,-12.362);
\node[anchor=north west,text=ACMDarkBlue,font=\scriptsize,inner sep=0] at (10.920,-12.382) {\href{https://arxiv.org/abs/2506.04363}{WorldPrediction}};
\node[anchor=north west,text=black!55,font=\tiny,inner sep=0] at (10.920,-12.612) {\href{https://arxiv.org/abs/2506.04363}{arXiv:2506.04363}};
\fill[sgBridge!78!black] (0,-13.092) rectangle (13.80,-13.592);
\node[anchor=west,text=white,font=\footnotesize\bfseries,inner sep=0] at (0.12,-13.342) {PREDICTION-TO-ACTION bridges};
\node[anchor=north west,inner sep=0] at (0.060,-13.662) {\href{https://arxiv.org/abs/2410.18072}{\includegraphics[width=2.660cm,height=2.128cm]{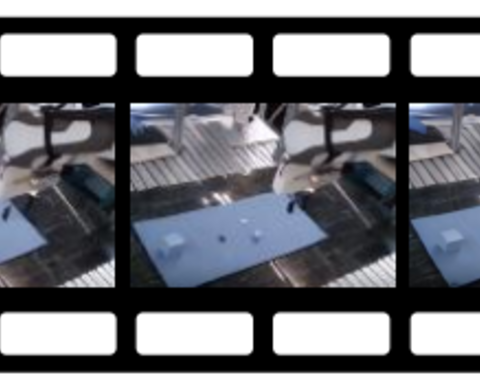}}};
\draw[black!35,line width=0.35pt] (0.060,-13.662) rectangle (2.720,-15.790);
\node[anchor=north west,text=ACMDarkBlue,font=\scriptsize,inner sep=0] at (0.080,-15.810) {\href{https://arxiv.org/abs/2410.18072}{WorldSimBench}};
\node[anchor=north west,text=black!55,font=\tiny,inner sep=0] at (0.080,-16.040) {\href{https://arxiv.org/abs/2410.18072}{arXiv:2410.18072}};
\node[anchor=north west,inner sep=0] at (2.770,-13.662) {\href{https://arxiv.org/abs/2510.18135}{\includegraphics[width=2.660cm,height=2.128cm]{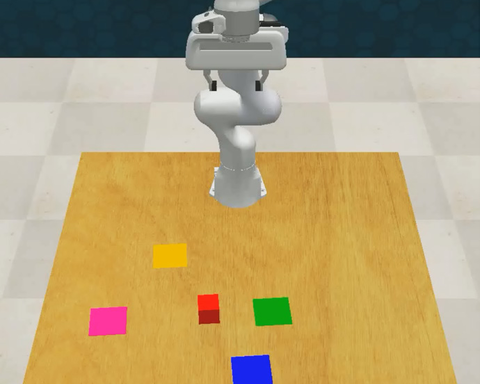}}};
\draw[black!35,line width=0.35pt] (2.770,-13.662) rectangle (5.430,-15.790);
\node[anchor=north west,text=ACMDarkBlue,font=\scriptsize,inner sep=0] at (2.790,-15.810) {\href{https://arxiv.org/abs/2510.18135}{World-in-World}};
\node[anchor=north west,text=black!55,font=\tiny,inner sep=0] at (2.790,-16.040) {\href{https://arxiv.org/abs/2510.18135}{arXiv:2510.18135}};
\node[anchor=north west,inner sep=0] at (5.480,-13.662) {\href{https://arxiv.org/abs/2604.19092}{\includegraphics[width=2.660cm,height=2.128cm]{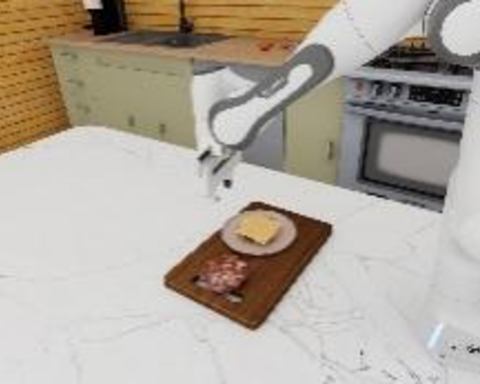}}};
\draw[black!35,line width=0.35pt] (5.480,-13.662) rectangle (8.140,-15.790);
\node[anchor=north west,text=ACMDarkBlue,font=\scriptsize,inner sep=0] at (5.500,-15.810) {\href{https://arxiv.org/abs/2604.19092}{RoboWM-Bench}};
\node[anchor=north west,text=black!55,font=\tiny,inner sep=0] at (5.500,-16.040) {\href{https://arxiv.org/abs/2604.19092}{arXiv:2604.19092}};
\node[anchor=north west,inner sep=0] at (8.190,-13.662) {\href{https://arxiv.org/abs/2602.08971}{\includegraphics[width=2.660cm,height=2.128cm]{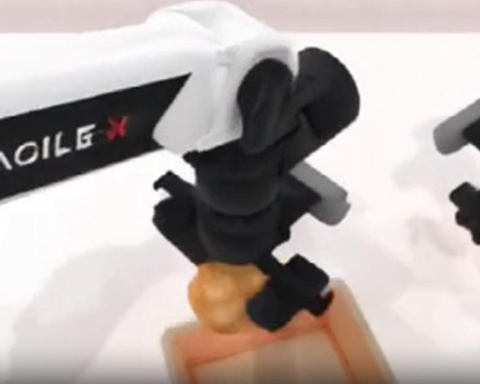}}};
\draw[black!35,line width=0.35pt] (8.190,-13.662) rectangle (10.850,-15.790);
\node[anchor=north west,text=ACMDarkBlue,font=\scriptsize,inner sep=0] at (8.210,-15.810) {\href{https://arxiv.org/abs/2602.08971}{WorldArena}};
\node[anchor=north west,text=black!55,font=\tiny,inner sep=0] at (8.210,-16.040) {\href{https://arxiv.org/abs/2602.08971}{arXiv:2602.08971}};
\end{tikzpicture}}
\caption{\textbf{Benchmarks across the surveyed corpus.} Representative subject figures from the catalogue, grouped by evaluation lane rather than embodiment: \emph{closed-loop} task-success suites, \emph{open-loop} world-model and video-generation evaluation, and the \emph{prediction-to-action bridges}. Each panel is a live hyperlink to the source paper on arXiv and is annotated with its arXiv id (per-panel figure provenance in Table~\ref{tab:panel_provenance}); the figure is vector with photographs embedded at full resolution, so it can be zoomed without pixelation. Only figures that depict the benchmark's task or scene are shown; results and leaderboard figures appear in Figure~\ref{fig:results_gallery}. Benchmarks with no locatable subject figure are omitted, not fabricated. Images \textcopyright\ their respective authors, reproduced for scholarly review.}
\Description{A gallery of benchmark subject figures in three lane bands. The top band, closed-loop task-success suites, shows ten policy and embodied benchmark scenes (robot arms and simulated households). The middle band, open-loop world-model and video-generation evaluation, shows ten generated-video or physical-scene samples. The bottom band, prediction-to-action bridges, shows four world-model benchmark scenes. Each panel is labelled with the benchmark name and its arXiv id.}
\label{fig:subject_gallery}
\end{figure*}

%% file: tables/tab_compare_capability.tex
{\footnotesize
\setlength{\tabcolsep}{5pt}
\renewcommand{\arraystretch}{1.08}
\begin{longtable}{@{}p{0.26\textwidth} c c >{\raggedright\arraybackslash}p{0.48\textwidth} c@{}}
\caption{\textbf{Capability deep-dive}: the benchmarks that probe each of the $\beta$ capability
cuts, and whether each builds a native VLA-vs-world-model contrast. Native contrasts cluster in the
hidden-state, theory-of-mind and counterfactual \emph{reasoning} cuts (LIBERO-Mem, ComPhy, AGENT,
BIB, ACRE, CoPhy, CausalWorld); the long-horizon and social \emph{task-success} cuts are largely
model-agnostic (ALFWorld's abstract-to-grounded transfer is the lone partial). \textbf{Eval}: \dopen/\dclosed. \textbf{VLA vs WM}: \yy\ native contrast,
\pp\ partial, \dm\ model-agnostic; each cut's header tallies the fraction of its benchmarks that
build any (native or partial) contrast.}\label{tab:compare_capability}\\
\toprule
\textbf{Benchmark} & \textbf{Year} & \textbf{Eval} & \textbf{What it probes} & \textbf{VLA vs WM}\\
\midrule
\endfirsthead
\multicolumn{5}{@{}l}{\footnotesize\emph{Table~\ref{tab:compare_capability} (continued)}}\\
\toprule
\textbf{Benchmark} & \textbf{Year} & \textbf{Eval} & \textbf{What it probes} & \textbf{VLA vs WM}\\
\midrule
\endhead
\bottomrule
\endlastfoot
\multicolumn{4}{@{}l}{\rule{0pt}{2.2ex}\textbf{\textsf{Hidden-state / memory / occlusion}}} & {\footnotesize\itshape 3/4}\\[1pt]
\textbf{LIBERO-Mem}~\cite{liberomem} & 2025 & \dclosed & occluded object state carried across time & \yy\\
\textbf{TEACh}~\cite{teach} & 2021 & \dclosed & partner belief and history via dialog & \dm\\
\textbf{ComPhy}~\cite{comphy} & 2022 & \dopen & hidden mass and charge inferred from collisions & \yy\\
\textbf{Bongard-HOI}~\cite{bongardhoi} & 2022 & \dopen & few-shot hidden-concept induction & \pp\\
\midrule
\multicolumn{4}{@{}l}{\rule{0pt}{2.2ex}\textbf{\textsf{Theory-of-mind}}} & {\footnotesize\itshape 3/4}\\[1pt]
\textbf{AGENT}~\cite{agent} & 2021 & \dopen & expectations over an agent's goals and actions & \yy\\
\textbf{BIB}~\cite{bib} & 2021 & \dopen & infant-style goal inference from behaviour & \yy\\
\textbf{PHASE}~\cite{phase} & 2021 & \dclosed & social relations and goals inferred from motion & \pp\\
\textbf{SocialAI}~\cite{socialai} & 2021 & \dclosed & language-based social interaction with agents & \dm\\
\midrule
\multicolumn{4}{@{}l}{\rule{0pt}{2.2ex}\textbf{\textsf{Counterfactual / physical inference}}} & {\footnotesize\itshape 3/6}\\[1pt]
\textbf{WorldPrediction}~\cite{worldprediction} & 2025 & \dopen & discriminate correct vs counterfactual action & \dm\\
\textbf{Physion}~\cite{physion} & 2021 & \dopen & predict a physical outcome from vision & \dm\\
\textbf{Physion++}~\cite{physionpp} & 2023 & \dopen & infer latent physical properties online & \dm\\
\textbf{ACRE}~\cite{acre} & 2021 & \dopen & abstract causal reasoning over blicket tests & \yy\\
\textbf{CoPhy}~\cite{cophy} & 2019 & \dopen & counterfactual outcome under a changed cause & \yy\\
\textbf{CausalWorld}~\cite{causalworld} & 2020 & \dclosed & causal do-interventions on task variables & \yy\\
\midrule
\multicolumn{4}{@{}l}{\rule{0pt}{2.2ex}\textbf{\textsf{Long-horizon / compositional}}} & {\footnotesize\itshape 1/6}\\[1pt]
\textbf{CALVIN}~\cite{calvin} & 2021 & \dclosed & chain several language subtasks in a row & \dm\\
\textbf{BEHAVIOR-1K}~\cite{behavior1k} & 2024 & \dclosed & 1000 long-horizon household activities & \dm\\
\textbf{ALFRED}~\cite{alfred} & 2019 & \dclosed & instruction to action under partial view & \dm\\
\textbf{ALFWorld}~\cite{alfworld} & 2020 & \dclosed & abstract-text to grounded-task transfer & \pp\\
\textbf{EmbodiedBench}~\cite{embodiedbench} & 2025 & \dclosed & broad long-horizon embodied planning & \dm\\
\textbf{PARTNR}~\cite{partnr} & 2024 & \dclosed & human-robot collaborative household tasks & \dm\\
\midrule
\multicolumn{4}{@{}l}{\rule{0pt}{2.2ex}\textbf{\textsf{Social / multi-agent}}} & {\footnotesize\itshape 0/5}\\[1pt]
\textbf{Habitat 3.0}~\cite{habitat30} & 2023 & \dclosed & human-robot co-habitation and hand-off & \dm\\
\textbf{SocNavBench}~\cite{socnavbench} & 2021 & \dclosed & socially compliant navigation & \dm\\
\textbf{Melting Pot}~\cite{meltingpot} & 2021 & \dclosed & mixed-motive multi-agent generalization & \dm\\
\textbf{SMAC}~\cite{smac} & 2019 & \dclosed & cooperative multi-agent micro-control & \dm\\
\textbf{CrowdNav}~\cite{crowdnav} & 2018 & \dclosed & socially-aware navigation among pedestrians & \dm\\
\end{longtable}
}

%% file: tables/tab_compare_wmeval.tex
{\footnotesize
\setlength{\tabcolsep}{5pt}
\renewcommand{\arraystretch}{1.0}
\newcommand{\mgen}{\textcolor{teal!72!black}{\textsf{Gen}}}
\newcommand{\mqa}{\textcolor{blue!62!black}{\textsf{QA}}}
\newcommand{\mpred}{\textcolor{violet!72!black}{\textsf{Pred}}}
\newcommand{\sauto}{\textcolor{black!55}{\textsf{Auto}}}
\newcommand{\sacc}{\textcolor{ForestGreen!88!black}{\textsf{Acc}}}
\newcommand{\svoe}{\textcolor{orange!85!black}{\textsf{VOE}}}
\newcommand{\shuman}{\textcolor{magenta!72!black}{\textsf{Human}}}
\begin{longtable}{@{}p{0.25\textwidth} c c >{\raggedright\arraybackslash}p{0.40\textwidth} c c@{}}
\caption{\textbf{World-model-evaluation deep-dive}: the 34 benchmarks of the world-model-evaluation
lane, by sub-type -- the P3 counterpart of the skill-building branch deep-dive. \textbf{Mode}:
\mgen\ evaluates generated video, \mqa\ video question-answering, \mpred\ physical prediction.
\textbf{Signal}: \sauto\ computed/learned metric, \sacc\ accuracy vs ground truth, \svoe\
violation-of-expectation, \shuman\ human study. \textbf{Physical}: targets physical-law
understanding, \yy\ core, \pp\ adjacent, \nn\ general. Generation-quality suites dominate and are
physics-agnostic; only the physical-reasoning and physical-commonsense sub-types test physical law,
and every lane benchmark scores open-loop, never in a closed loop.}\label{tab:compare_wmeval}\\
\toprule
\textbf{Benchmark} & \textbf{Year} & \textbf{Mode} & \textbf{What it scores} & \textbf{Signal} & \textbf{Physical}\\
\midrule
\endfirsthead
\multicolumn{6}{@{}l}{\footnotesize\emph{Table~\ref{tab:compare_wmeval} (continued)}}\\
\toprule
\textbf{Benchmark} & \textbf{Year} & \textbf{Mode} & \textbf{What it scores} & \textbf{Signal} & \textbf{Physical}\\
\midrule
\endhead
\bottomrule
\endlastfoot
\multicolumn{6}{@{}l}{\rule{0pt}{2.0ex}\textbf{\textsf{Generation quality (multi-dimensional)}}}\\[1pt]
\textbf{EvalCrafter}~\cite{evalcrafter} & 2023 & \mgen & 17-metric text-to-video quality suite & \sauto & \nn\\
\textbf{FETV}~\cite{fetv} & 2023 & \mgen & fine-grained text-to-video quality & \sauto & \nn\\
\textbf{VBench}~\cite{vbench} & 2023 & \mgen & 16 disentangled generation dimensions & \sauto & \nn\\
\textbf{StoryBench}~\cite{storybench} & 2023 & \mgen & continuous story-visualisation quality & \sauto & \nn\\
\textbf{DEVIL}~\cite{devil} & 2024 & \mgen & content-dynamics quality of T2V & \sauto & \nn\\
\textbf{T2V-CompBench}~\cite{t2vcompbench} & 2024 & \mgen & compositional text-to-video quality & \sauto & \nn\\
\textbf{T2VScore}~\cite{t2vscore} & 2024 & \mgen & learned alignment + quality metric & \sauto & \nn\\
\textbf{TC-Bench}~\cite{tcbench} & 2024 & \mgen & temporal compositionality in generation & \sauto & \nn\\
\textbf{VideoScore}~\cite{videoscore} & 2024 & \mgen & learned automatic quality metric & \sauto & \nn\\
\textbf{EWMBench}~\cite{ewmbench} & 2025 & \mgen & scene / motion / semantic quality & \sauto & \nn\\
\textbf{VBench-2.0}~\cite{vbench20} & 2025 & \mgen & generation faithfulness & \sauto & \nn\\
\midrule
\multicolumn{6}{@{}l}{\rule{0pt}{2.0ex}\textbf{\textsf{Intuitive \& physical reasoning}}}\\[1pt]
\textbf{IntPhys}~\cite{intphys} & 2018 & \mpred & intuitive physics from video & \svoe & \yy\\
\textbf{Physion}~\cite{physion} & 2021 & \mpred & physical outcome prediction from vision & \sacc & \yy\\
\textbf{GRASP}~\cite{grasp} & 2023 & \mqa & intuitive physics in video MLLMs & \sacc & \yy\\
\textbf{Physion++}~\cite{physionpp} & 2023 & \mpred & prediction with online property inference & \sacc & \yy\\
\textbf{ContPhy}~\cite{contphy} & 2024 & \mqa & soft-body / fluid physical reasoning & \sacc & \yy\\
\textbf{Physics-IQ}~\cite{physicsiq} & 2025 & \mgen & physical realism of generated video & \sauto & \yy\\
\textbf{PhysBench}~\cite{physbench} & 2025 & \mqa & physical-world understanding for VLMs & \sacc & \yy\\
\textbf{PHYBench}~\cite{phybench} & 2025 & \mqa & physical perception and reasoning & \sacc & \yy\\
\textbf{IntPhys 2}~\cite{intphys2} & 2025 & \mpred & intuitive physics in complex scenes & \svoe & \yy\\
\midrule
\multicolumn{6}{@{}l}{\rule{0pt}{2.0ex}\textbf{\textsf{Physical commonsense in generated video}}}\\[1pt]
\textbf{PhyGenBench}~\cite{phygenbench} & 2024 & \mgen & physical-commonsense correctness in gen & \sauto & \yy\\
\textbf{VideoPhy}~\cite{videophy} & 2024 & \mgen & physical commonsense in generated video & \shuman & \yy\\
\textbf{VideoPhy-2}~\cite{videophy2} & 2025 & \mgen & action-centric physical commonsense & \shuman & \yy\\
\textbf{PhyWorldBench}~\cite{phyworldbench} & 2025 & \mgen & physical realism in text-to-video & \shuman & \yy\\
\textbf{IPV-Bench}~\cite{ipvbench} & 2025 & \mgen & impossible-video generation + understanding & \sacc & \yy\\
\midrule
\multicolumn{6}{@{}l}{\rule{0pt}{2.0ex}\textbf{\textsf{Counterfactual / causal / temporal}}}\\[1pt]
\textbf{CATER}~\cite{cater} & 2019 & \mqa & compositional-action temporal reasoning & \sacc & \nn\\
\textbf{CLEVRER}~\cite{clevrer} & 2019 & \mqa & causal / counterfactual video reasoning & \sacc & \pp\\
\textbf{WorldPrediction}~\cite{worldprediction} & 2025 & \mqa & counterfactual action recognition & \sacc & \nn\\
\midrule
\multicolumn{6}{@{}l}{\rule{0pt}{2.0ex}\textbf{\textsf{Video-language alignment / hallucination}}}\\[1pt]
\textbf{VideoCon}~\cite{videocon} & 2023 & \mqa & robust video-language alignment & \sauto & \nn\\
\textbf{VideoHallucer}~\cite{videohallucer} & 2024 & \mqa & hallucination in video-language models & \sacc & \nn\\
\midrule
\multicolumn{6}{@{}l}{\rule{0pt}{2.0ex}\textbf{\textsf{World-model prediction \& world generation}}}\\[1pt]
\textbf{Vista}~\cite{vista} & 2024 & \mgen & driving world-model generation & \sauto & \pp\\
\textbf{EVA-Bench}~\cite{evabench} & 2024 & \mpred & embodied video anticipation & \sauto & \pp\\
\textbf{WorldModelBench}~\cite{worldmodelbench} & 2025 & \mgen & video world-model prediction quality & \shuman & \pp\\
\textbf{WorldScore}~\cite{worldscore} & 2025 & \mgen & controllable world-generation quality & \sauto & \nn\\
\end{longtable}
}

%% file: figures/fig_results_gallery.tex
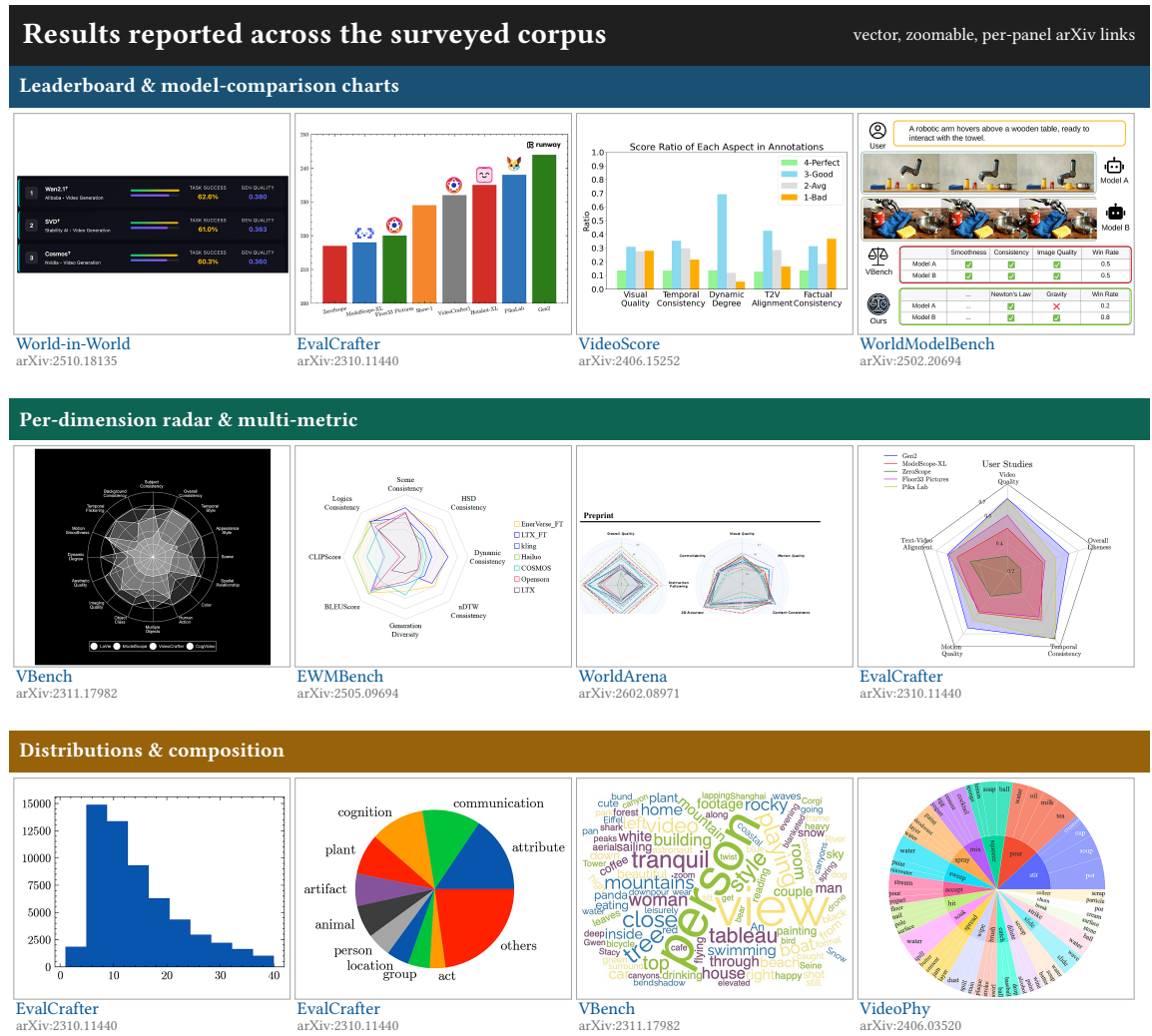
\begin{figure*}[p]
\centering
\definecolor{rgA}{HTML}{1F618D}\definecolor{rgB}{HTML}{117A65}\definecolor{rgC}{HTML}{B9770E}%
\resizebox{\textwidth}{!}{%
\begin{tikzpicture}[x=1cm,y=1cm]
\fill[black!88] (0,0) rectangle (13.62,-0.72);
\node[anchor=west,text=white,font=\large\bfseries,inner sep=0] at (0.16,-0.36) {Results reported across the surveyed corpus};
\node[anchor=east,text=white!70,font=\scriptsize,inner sep=0] at (13.46,-0.36) {vector, zoomable, per-panel arXiv links};
\fill[rgA!82!black] (0,-0.720) rectangle (13.62,-1.220);
\node[anchor=west,text=white,font=\footnotesize\bfseries,inner sep=0] at (0.12,-0.970) {Leaderboard \& model-comparison charts};
\node[anchor=north west,inner sep=0] at (0.060,-1.290) {\href{https://arxiv.org/abs/2510.18135}{\includegraphics[width=3.300cm,height=2.640cm]{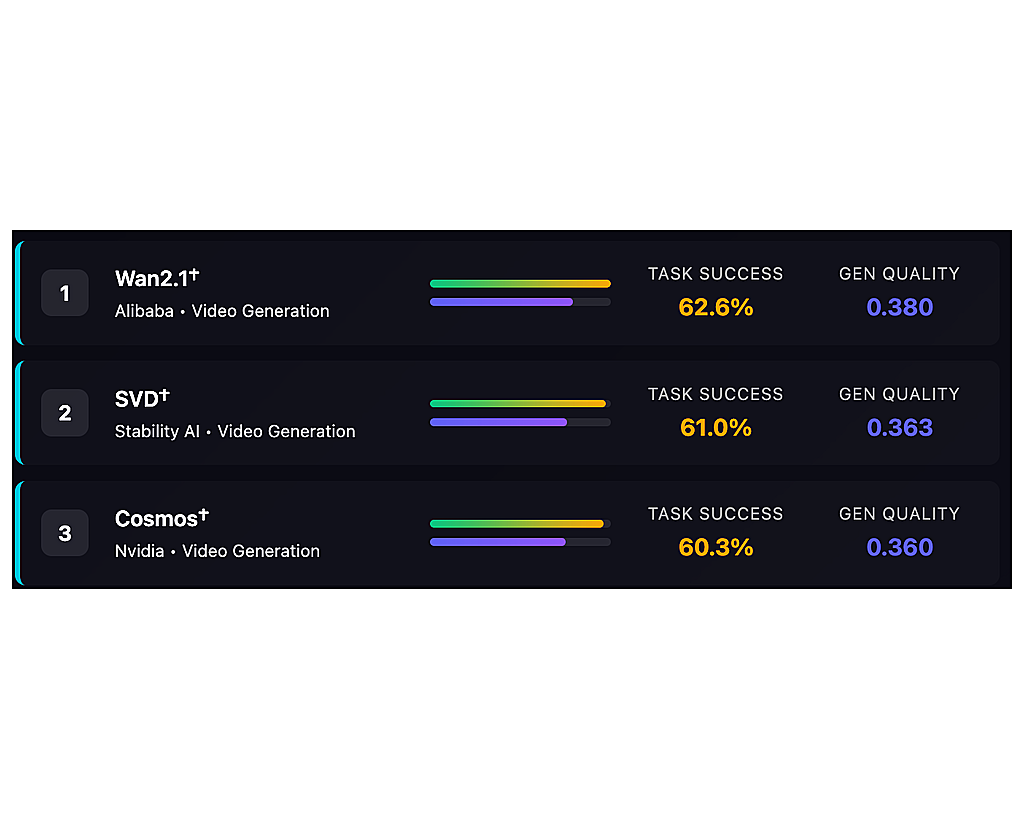}}};
\draw[black!35,line width=0.35pt] (0.060,-1.290) rectangle (3.360,-3.930);
\node[anchor=north west,text=ACMDarkBlue,font=\scriptsize,inner sep=0] at (0.080,-3.950) {\href{https://arxiv.org/abs/2510.18135}{World-in-World}};
\node[anchor=north west,text=black!55,font=\tiny,inner sep=0] at (0.080,-4.180) {\href{https://arxiv.org/abs/2510.18135}{arXiv:2510.18135}};
\node[anchor=north west,inner sep=0] at (3.420,-1.290) {\href{https://arxiv.org/abs/2310.11440}{\includegraphics[width=3.300cm,height=2.640cm]{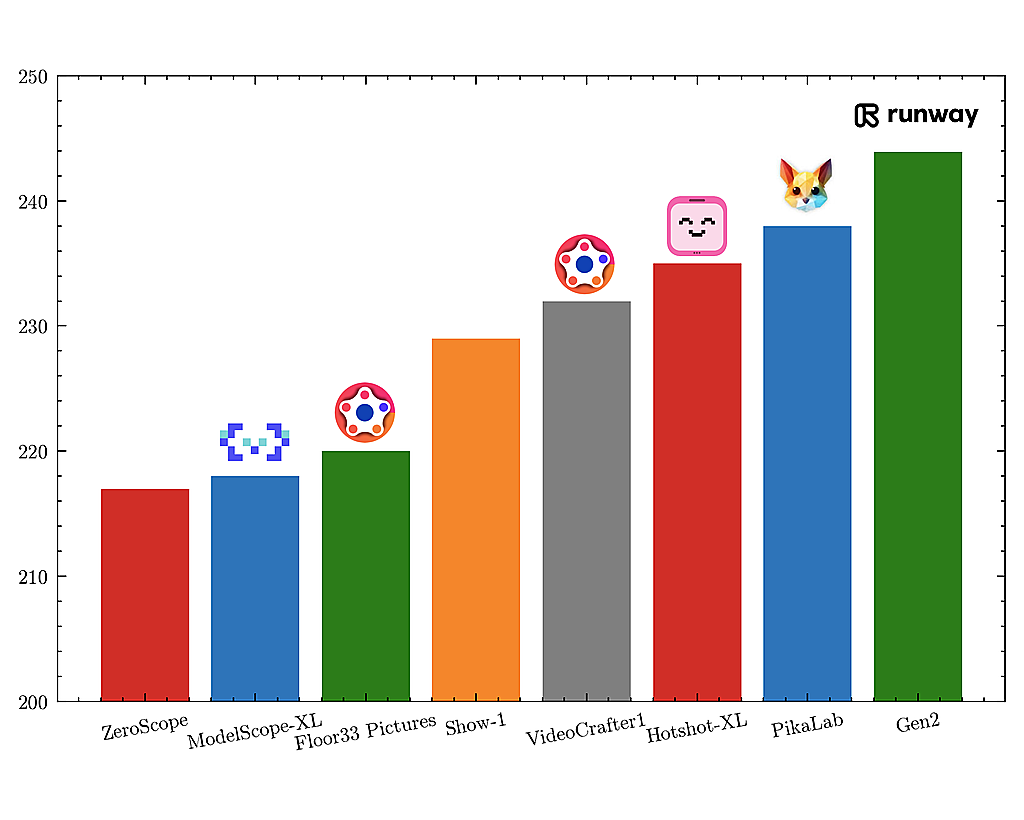}}};
\draw[black!35,line width=0.35pt] (3.420,-1.290) rectangle (6.720,-3.930);
\node[anchor=north west,text=ACMDarkBlue,font=\scriptsize,inner sep=0] at (3.440,-3.950) {\href{https://arxiv.org/abs/2310.11440}{EvalCrafter}};
\node[anchor=north west,text=black!55,font=\tiny,inner sep=0] at (3.440,-4.180) {\href{https://arxiv.org/abs/2310.11440}{arXiv:2310.11440}};
\node[anchor=north west,inner sep=0] at (6.780,-1.290) {\href{https://arxiv.org/abs/2406.15252}{\includegraphics[width=3.300cm,height=2.640cm]{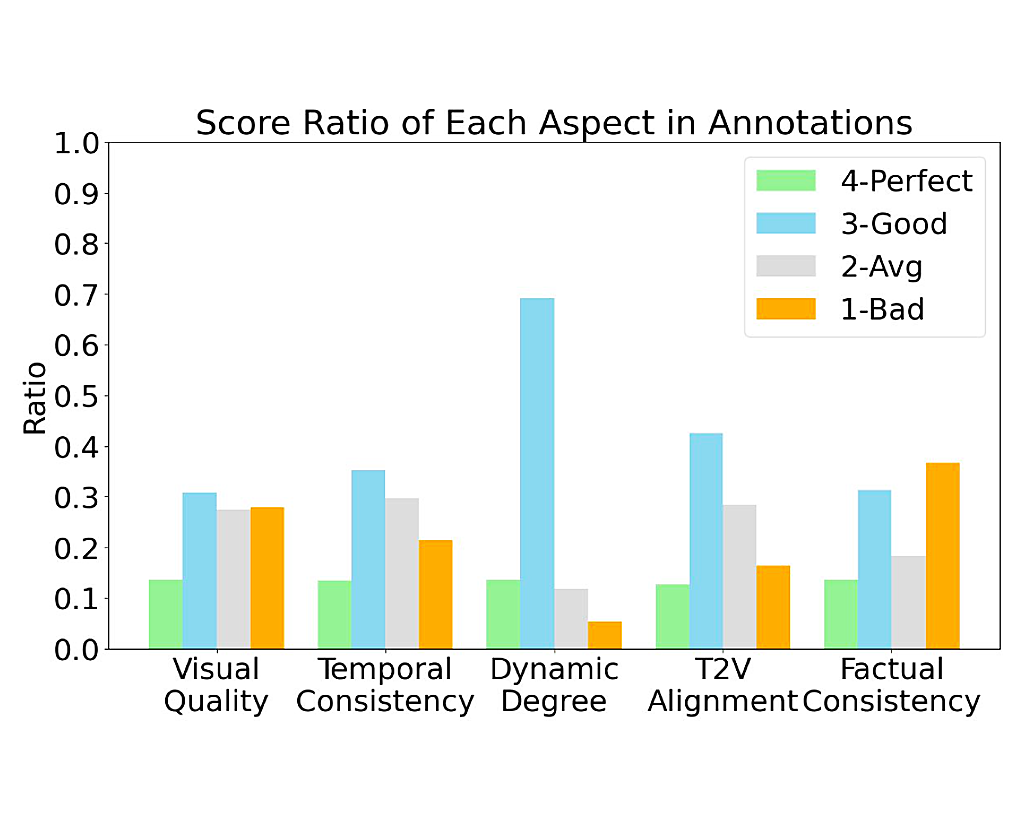}}};
\draw[black!35,line width=0.35pt] (6.780,-1.290) rectangle (10.080,-3.930);
\node[anchor=north west,text=ACMDarkBlue,font=\scriptsize,inner sep=0] at (6.800,-3.950) {\href{https://arxiv.org/abs/2406.15252}{VideoScore}};
\node[anchor=north west,text=black!55,font=\tiny,inner sep=0] at (6.800,-4.180) {\href{https://arxiv.org/abs/2406.15252}{arXiv:2406.15252}};
\node[anchor=north west,inner sep=0] at (10.140,-1.290) {\href{https://arxiv.org/abs/2502.20694}{\includegraphics[width=3.300cm,height=2.640cm]{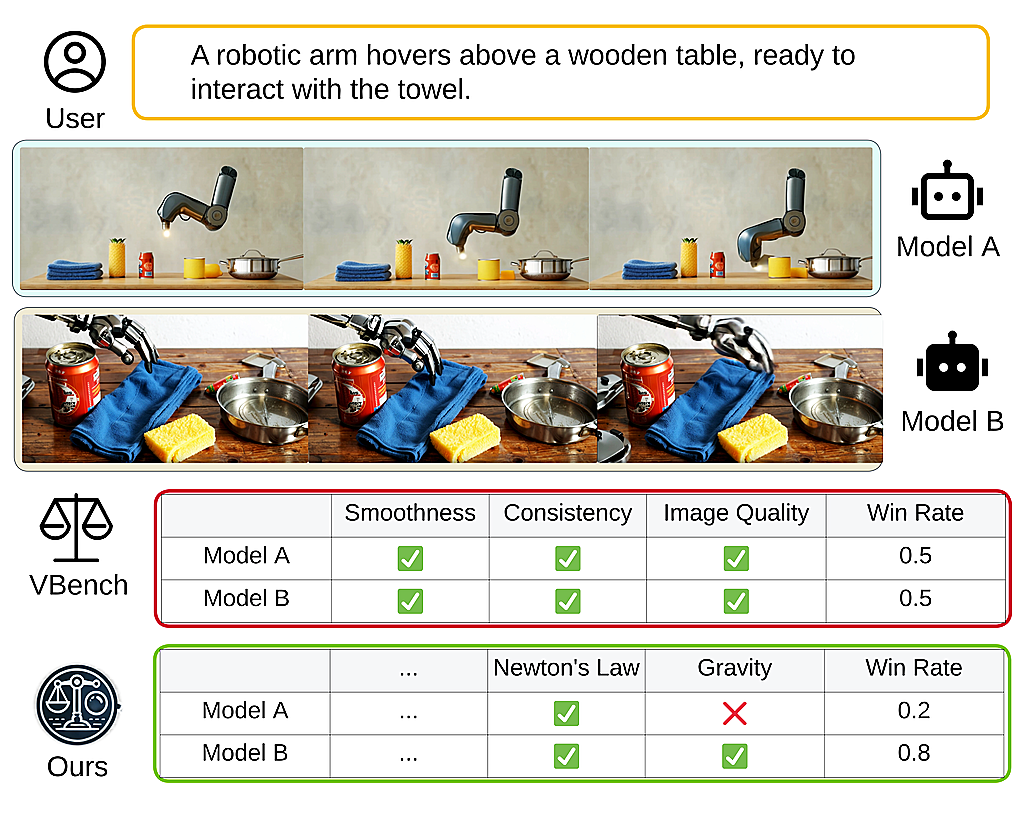}}};
\draw[black!35,line width=0.35pt] (10.140,-1.290) rectangle (13.440,-3.930);
\node[anchor=north west,text=ACMDarkBlue,font=\scriptsize,inner sep=0] at (10.160,-3.950) {\href{https://arxiv.org/abs/2502.20694}{WorldModelBench}};
\node[anchor=north west,text=black!55,font=\tiny,inner sep=0] at (10.160,-4.180) {\href{https://arxiv.org/abs/2502.20694}{arXiv:2502.20694}};
\fill[rgB!82!black] (0,-4.690) rectangle (13.62,-5.190);
\node[anchor=west,text=white,font=\footnotesize\bfseries,inner sep=0] at (0.12,-4.940) {Per-dimension radar \& multi-metric};
\node[anchor=north west,inner sep=0] at (0.060,-5.260) {\href{https://arxiv.org/abs/2311.17982}{\includegraphics[width=3.300cm,height=2.640cm]{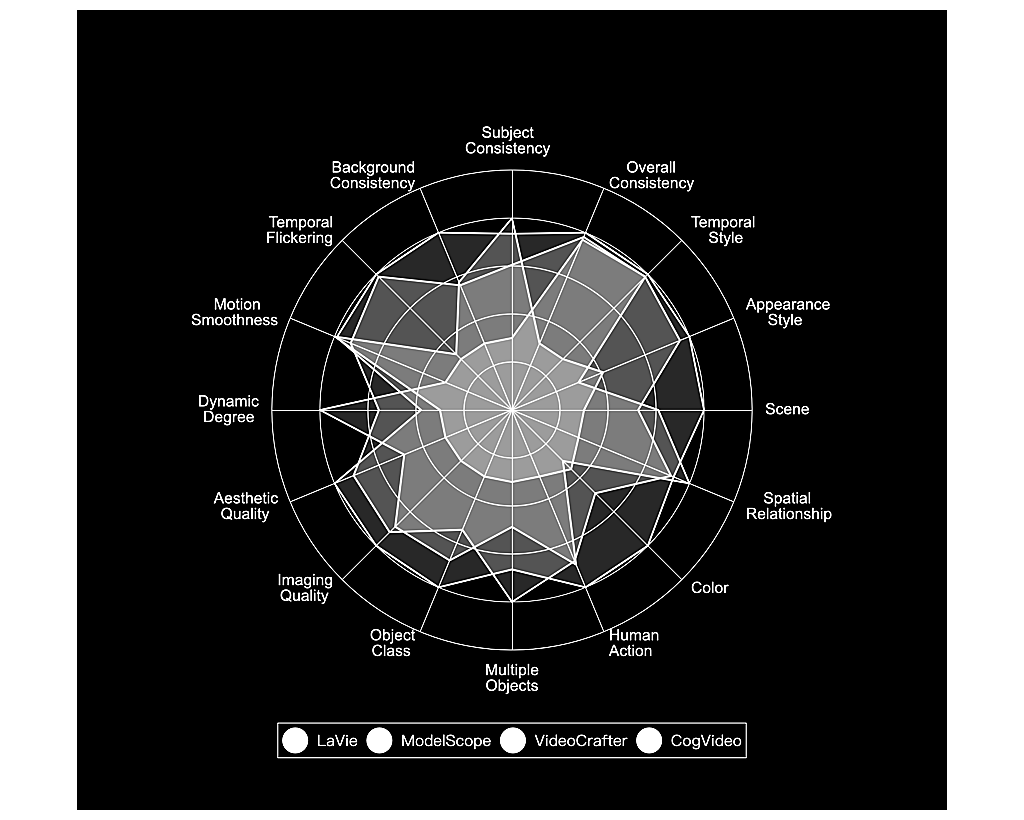}}};
\draw[black!35,line width=0.35pt] (0.060,-5.260) rectangle (3.360,-7.900);
\node[anchor=north west,text=ACMDarkBlue,font=\scriptsize,inner sep=0] at (0.080,-7.920) {\href{https://arxiv.org/abs/2311.17982}{VBench}};
\node[anchor=north west,text=black!55,font=\tiny,inner sep=0] at (0.080,-8.150) {\href{https://arxiv.org/abs/2311.17982}{arXiv:2311.17982}};
\node[anchor=north west,inner sep=0] at (3.420,-5.260) {\href{https://arxiv.org/abs/2505.09694}{\includegraphics[width=3.300cm,height=2.640cm]{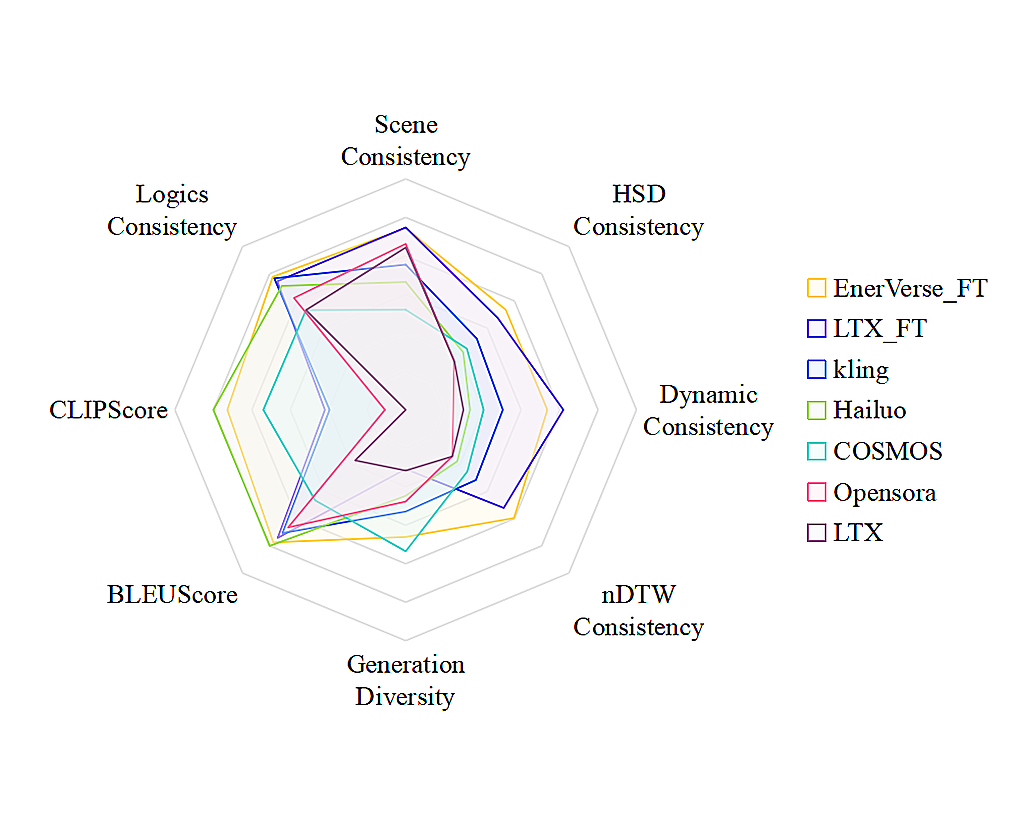}}};
\draw[black!35,line width=0.35pt] (3.420,-5.260) rectangle (6.720,-7.900);
\node[anchor=north west,text=ACMDarkBlue,font=\scriptsize,inner sep=0] at (3.440,-7.920) {\href{https://arxiv.org/abs/2505.09694}{EWMBench}};
\node[anchor=north west,text=black!55,font=\tiny,inner sep=0] at (3.440,-8.150) {\href{https://arxiv.org/abs/2505.09694}{arXiv:2505.09694}};
\node[anchor=north west,inner sep=0] at (6.780,-5.260) {\href{https://arxiv.org/abs/2602.08971}{\includegraphics[width=3.300cm,height=2.640cm]{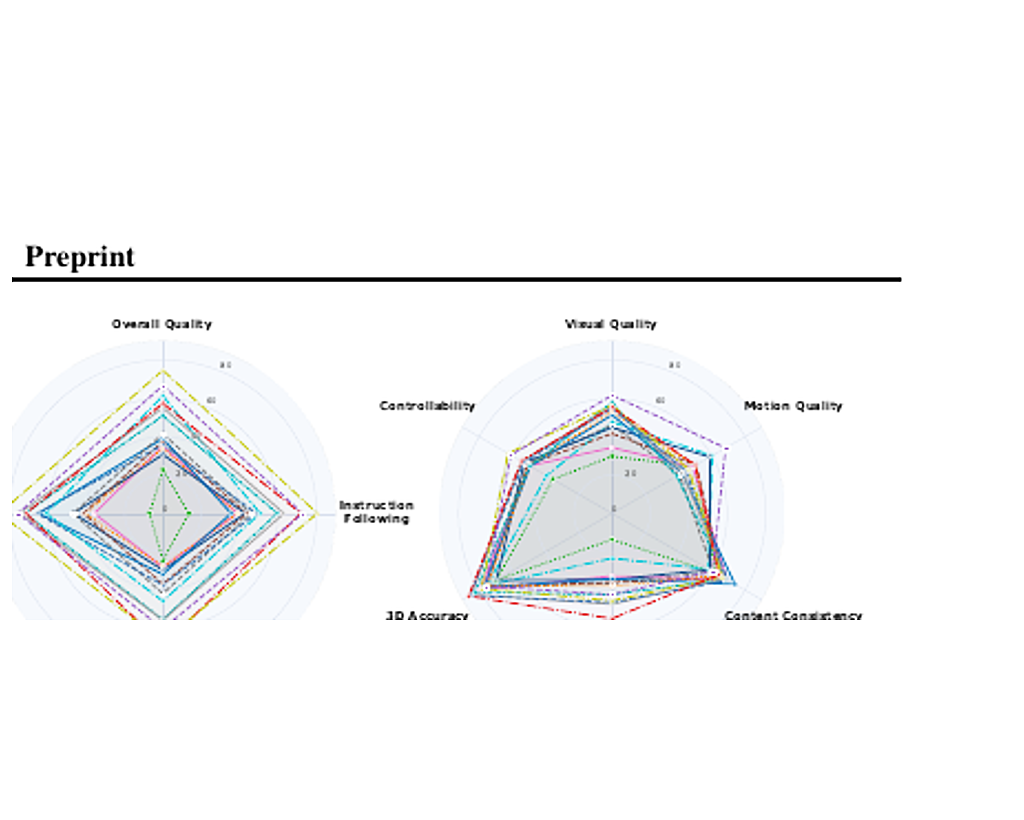}}};
\draw[black!35,line width=0.35pt] (6.780,-5.260) rectangle (10.080,-7.900);
\node[anchor=north west,text=ACMDarkBlue,font=\scriptsize,inner sep=0] at (6.800,-7.920) {\href{https://arxiv.org/abs/2602.08971}{WorldArena}};
\node[anchor=north west,text=black!55,font=\tiny,inner sep=0] at (6.800,-8.150) {\href{https://arxiv.org/abs/2602.08971}{arXiv:2602.08971}};
\node[anchor=north west,inner sep=0] at (10.140,-5.260) {\href{https://arxiv.org/abs/2310.11440}{\includegraphics[width=3.300cm,height=2.640cm]{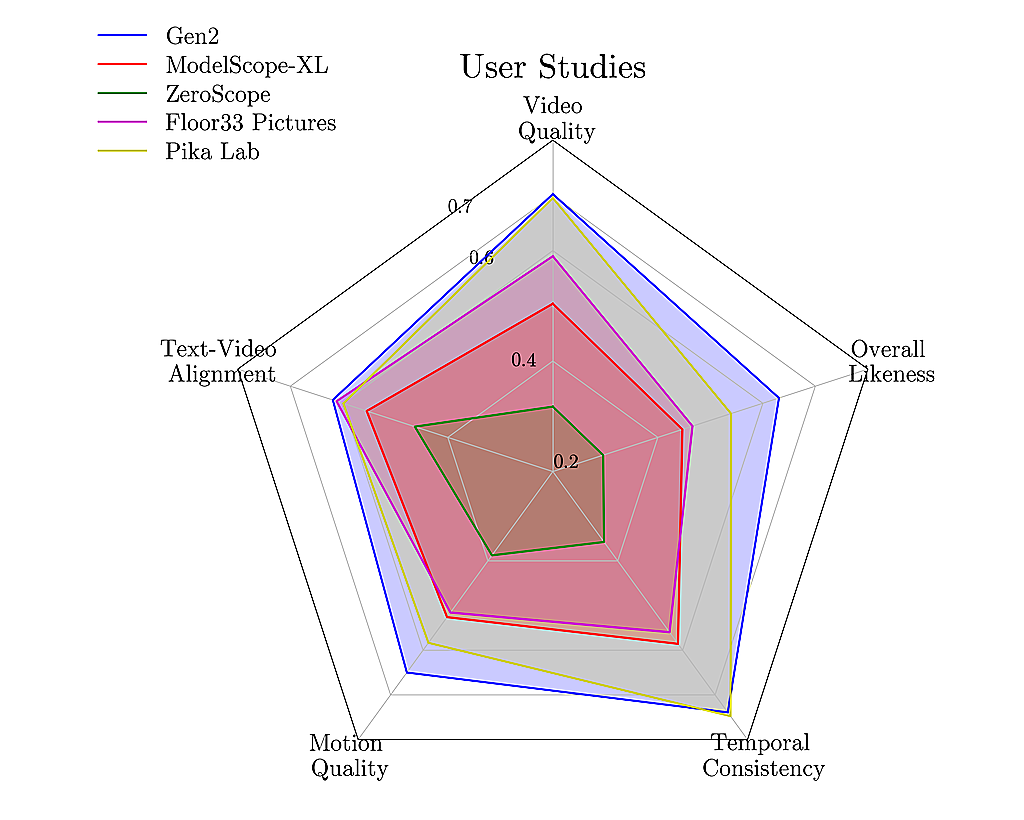}}};
\draw[black!35,line width=0.35pt] (10.140,-5.260) rectangle (13.440,-7.900);
\node[anchor=north west,text=ACMDarkBlue,font=\scriptsize,inner sep=0] at (10.160,-7.920) {\href{https://arxiv.org/abs/2310.11440}{EvalCrafter}};
\node[anchor=north west,text=black!55,font=\tiny,inner sep=0] at (10.160,-8.150) {\href{https://arxiv.org/abs/2310.11440}{arXiv:2310.11440}};
\fill[rgC!82!black] (0,-8.660) rectangle (13.62,-9.160);
\node[anchor=west,text=white,font=\footnotesize\bfseries,inner sep=0] at (0.12,-8.910) {Distributions \& composition};
\node[anchor=north west,inner sep=0] at (0.060,-9.230) {\href{https://arxiv.org/abs/2310.11440}{\includegraphics[width=3.300cm,height=2.640cm]{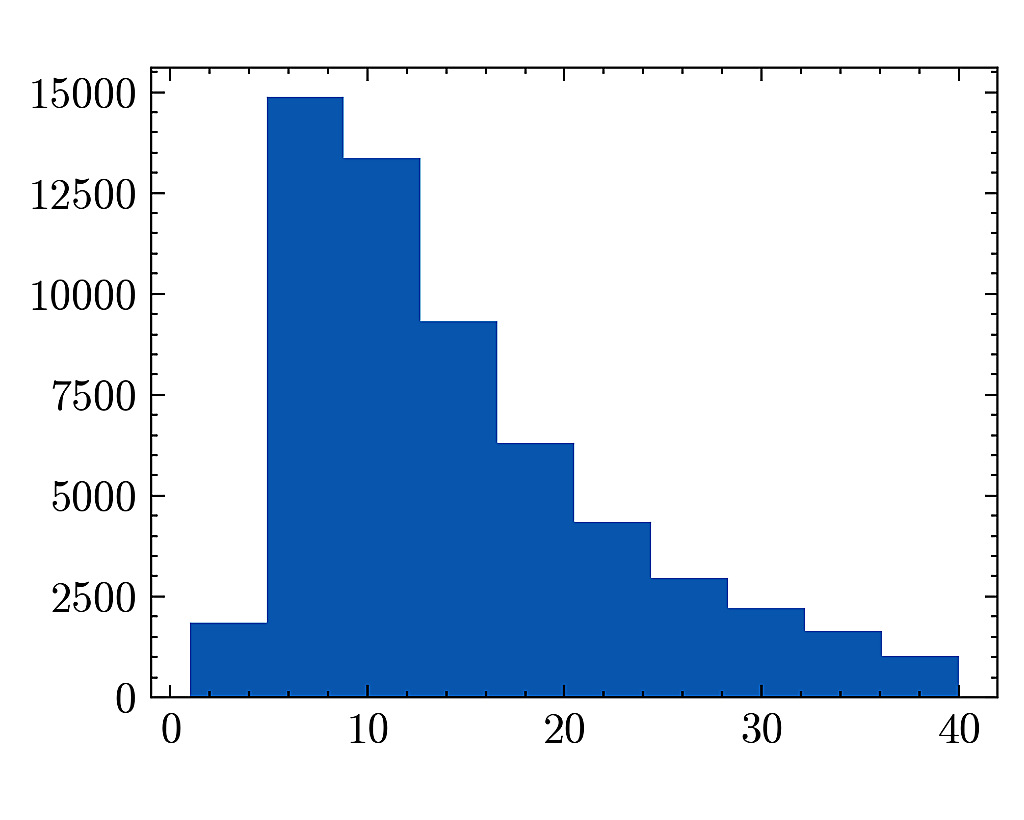}}};
\draw[black!35,line width=0.35pt] (0.060,-9.230) rectangle (3.360,-11.870);
\node[anchor=north west,text=ACMDarkBlue,font=\scriptsize,inner sep=0] at (0.080,-11.890) {\href{https://arxiv.org/abs/2310.11440}{EvalCrafter}};
\node[anchor=north west,text=black!55,font=\tiny,inner sep=0] at (0.080,-12.120) {\href{https://arxiv.org/abs/2310.11440}{arXiv:2310.11440}};
\node[anchor=north west,inner sep=0] at (3.420,-9.230) {\href{https://arxiv.org/abs/2310.11440}{\includegraphics[width=3.300cm,height=2.640cm]{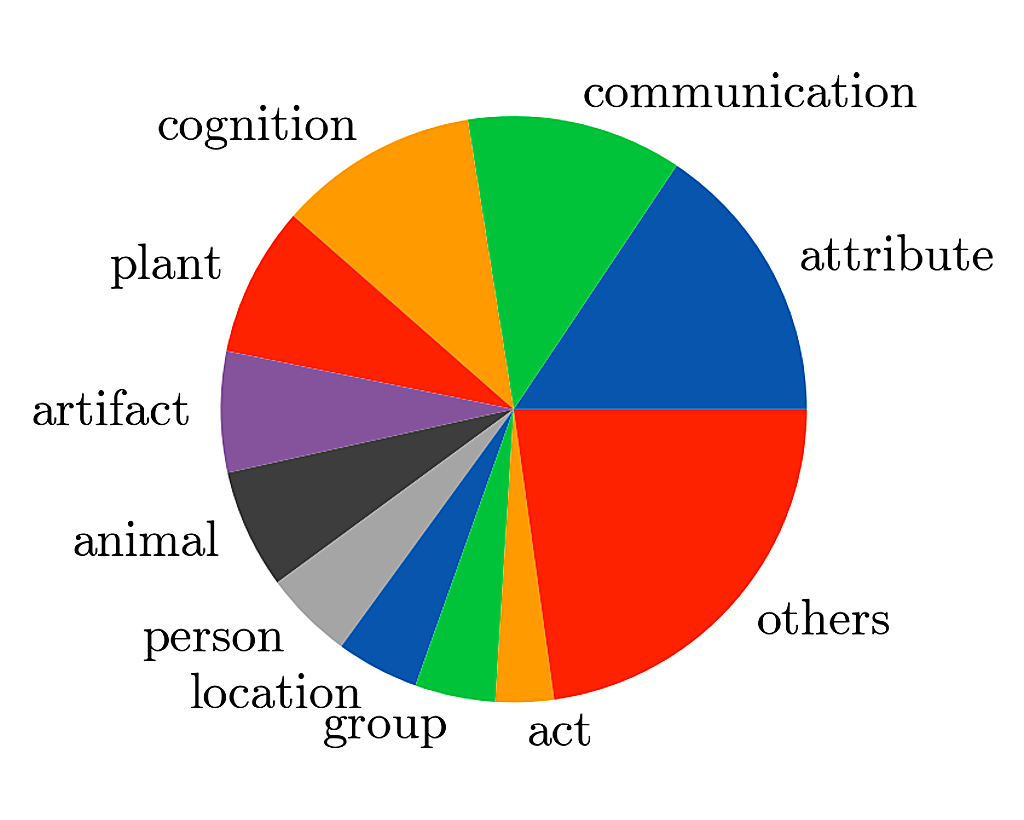}}};
\draw[black!35,line width=0.35pt] (3.420,-9.230) rectangle (6.720,-11.870);
\node[anchor=north west,text=ACMDarkBlue,font=\scriptsize,inner sep=0] at (3.440,-11.890) {\href{https://arxiv.org/abs/2310.11440}{EvalCrafter}};
\node[anchor=north west,text=black!55,font=\tiny,inner sep=0] at (3.440,-12.120) {\href{https://arxiv.org/abs/2310.11440}{arXiv:2310.11440}};
\node[anchor=north west,inner sep=0] at (6.780,-9.230) {\href{https://arxiv.org/abs/2311.17982}{\includegraphics[width=3.300cm,height=2.640cm]{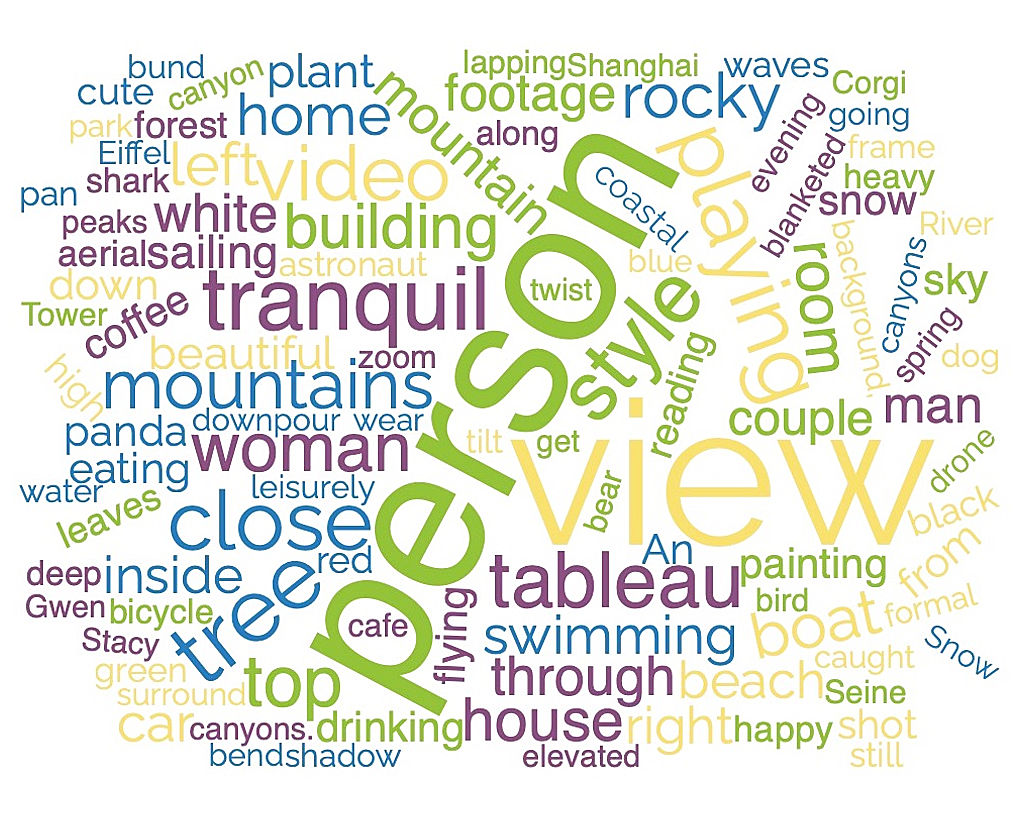}}};
\draw[black!35,line width=0.35pt] (6.780,-9.230) rectangle (10.080,-11.870);
\node[anchor=north west,text=ACMDarkBlue,font=\scriptsize,inner sep=0] at (6.800,-11.890) {\href{https://arxiv.org/abs/2311.17982}{VBench}};
\node[anchor=north west,text=black!55,font=\tiny,inner sep=0] at (6.800,-12.120) {\href{https://arxiv.org/abs/2311.17982}{arXiv:2311.17982}};
\node[anchor=north west,inner sep=0] at (10.140,-9.230) {\href{https://arxiv.org/abs/2406.03520}{\includegraphics[width=3.300cm,height=2.640cm]{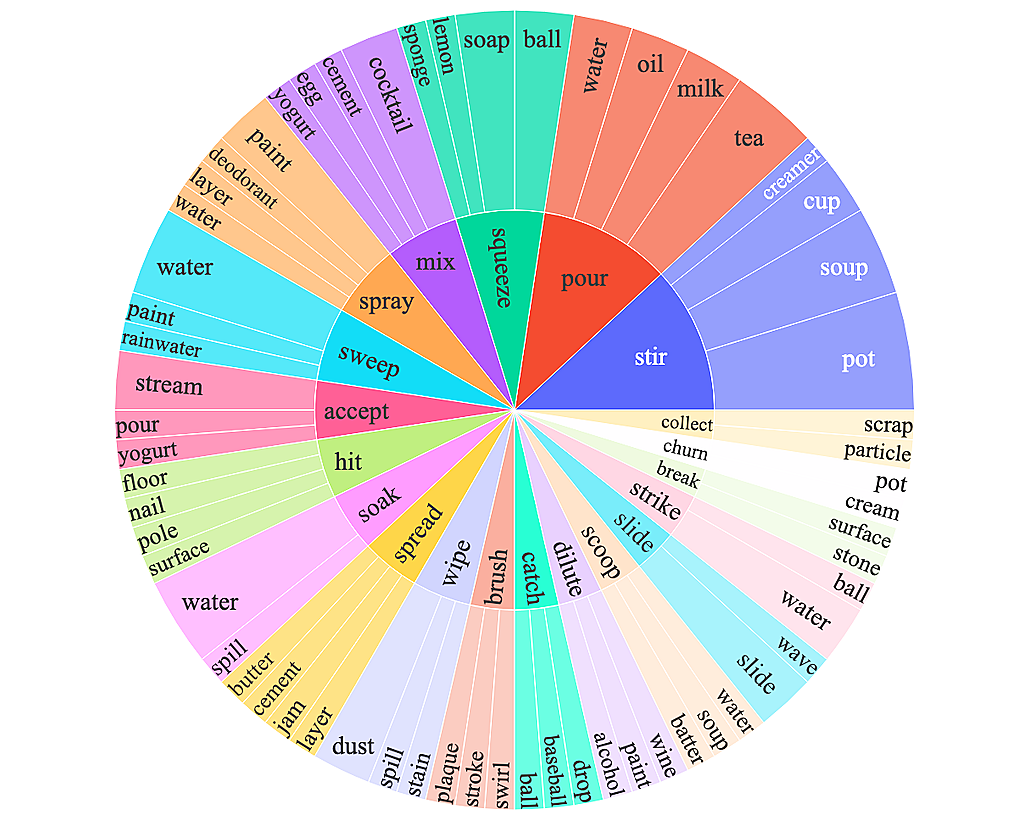}}};
\draw[black!35,line width=0.35pt] (10.140,-9.230) rectangle (13.440,-11.870);
\node[anchor=north west,text=ACMDarkBlue,font=\scriptsize,inner sep=0] at (10.160,-11.890) {\href{https://arxiv.org/abs/2406.03520}{VideoPhy}};
\node[anchor=north west,text=black!55,font=\tiny,inner sep=0] at (10.160,-12.120) {\href{https://arxiv.org/abs/2406.03520}{arXiv:2406.03520}};
\end{tikzpicture}}
\caption{\textbf{Results reported across the surveyed corpus.} Representative results figures from the catalogue, grouped by type of evidence: \emph{leaderboard and model-comparison charts}, \emph{per-dimension radar and multi-metric} plots, and \emph{distributions and composition}. Each panel is a live hyperlink to the source paper on arXiv and is annotated with its arXiv id (per-panel provenance in Table~\ref{tab:results_provenance}); the figure is vector with plots embedded at full resolution. Only results/leaderboard/metric figures are shown; task and scene figures appear in Figure~\ref{fig:subject_gallery}. As a benchmark survey the reported evidence is dominated by rankings, radars and distributions rather than training curves, and is sparser than the subject figures, so this gallery is smaller; benchmarks that publish no results figure are omitted, not fabricated. A benchmark may appear more than once when it reports several evidence types. Images \textcopyright\ their respective authors, reproduced for scholarly review.}
\Description{A gallery of benchmark results figures in three evidence-type bands. The top band, leaderboard and model-comparison charts, shows four ranking bar charts and win-rate tables. The middle band, per-dimension radar and multi-metric, shows three radar plots over evaluation dimensions. The bottom band, distributions and composition, shows a prompt-length histogram, a meta-type pie chart, and a prompt word cloud. Each panel is labelled with the benchmark name and its arXiv id.}
\label{fig:results_gallery}
\end{figure*}

%% file: sections/4_gap.tex
\section{Does prediction help acting? The contrast gap}
\label{sec:gap}

The survey's question can be stated precisely. Let a \emph{direct policy} map observations to actions,
and let a \emph{world-model policy} first predict future observations and then plan an action against
that prediction. The advantage of prediction is the closed-loop task-success \emph{gain} of the
world-model policy over a matched direct policy, and it is only meaningful when reported per
capability, since prediction should help more where hidden state, occlusion, or long horizons make
foresight valuable. Figure~\ref{fig:eval_loop} draws the evaluation loop this definition requires: a
task forks into a direct-VLA branch and a world-model branch, both run in one closed loop with state
feedback, both scored by the same task-success measure, and the difference read off per capability.

\input{figures/fig_eval_loop}

Today's benchmarks break this loop in one of two ways, visible as the two poles of
Figure~\ref{fig:corpus_collage}. Model-agnostic policy and embodied suites (Sections~\ref{sec:lane_policy},
\ref{sec:lane_embodied}) close the loop but run a single branch, so there is no contrast to read.
World-model-evaluation suites (Section~\ref{sec:lane_wmeval}) score the prediction branch open-loop
and never execute it, so a high score certifies generation quality, not task success. The
prediction-to-action bridges (Section~\ref{sec:lane_bridge}) close the loop for a world-model policy
but do not yet run the matched VLA branch per capability. No lane completes the diagram.

\subsection{``World model'' is three things, none of them sufficient alone}

Part of why the contrast is hard to build is that the term ``world model'' is overloaded.
Table~\ref{tab:wm_senses} separates two axes the literature routinely conflates. Along a
\emph{representation} axis a world model may be a latent-dynamics predictor (LWM; DreamerV3, TD-MPC2),
an action-conditioned future-frame predictor (WAM; iVideoGPT, GR-2), or a generative video model
(WFM; Sora, Cosmos, Genie). Along a \emph{role} axis it may be a neural simulator that trains or
evaluates a policy, or the object that a benchmark scores. The crucial fact is that no single sense
is simultaneously action-conditioned, run in a closed loop, and readily comparable to a VLA policy
under one protocol: latent and action-conditioned models act but are rarely scored by a standard
generation metric, while generative models carry the metrics but are rarely closed into a loop. A
benchmark that contrasts a world-model policy with a VLA policy therefore has to assemble properties
that no existing sense of the term supplies at once.

\input{tables/tab_wm_senses}

\subsection{Why each lane leaves the contrast unbuilt}

Table~\ref{tab:branch_limits} compares the four lanes on six evaluation axes and makes the pattern
lane-by-lane. Policy and embodied suites score the right quantity, closed-loop success, on the right
capabilities, but hold the model family fixed and so never contrast it. World-model-evaluation suites
vary the model family richly but score the wrong quantity for our question, open-loop prediction
quality, and never close the loop. The bridges are the only lane that scores closed-loop success of a
world model, but they are four benchmarks and none slices a VLA head-to-head by capability. The
result is a coverage hole precisely where the survey's question lives: the intersection of
closed-loop scoring, per-capability slicing, and a world-model-versus-VLA family contrast.

\input{tables/tab_branch_limits}

\subsection{Measuring the advantage: assessed, not adopted}

Closing the loop is necessary but not sufficient; the gain also needs an instrument. Proposals exist,
including behavioural advantage-of-prediction scores such as ACTION-ATLAS's World Advantage Score,
but they are put forward in position papers~\cite{pos_yangyu} rather than established across a
benchmark suite. We therefore \emph{assess} such instruments as candidates rather than adopting any
one as ground truth, and Section~\ref{sec:future} specifies how an advantage-of-prediction quantity
should be reported: as a per-capability distribution or curve against a matched baseline, never as a
single headline number. The empirical scale of the gap is worth restating: of 160 benchmarks, 138
are model-agnostic, 11 build a partial contrast, and 11 build an explicit one; counterfactual
capability, where foresight should matter most, is almost entirely unmeasured; and only four
benchmarks reach executed action at all.

\FloatBarrier

%% file: figures/fig_eval_loop.tex
\begin{figure}[t]
\centering
\includegraphics[width=\textwidth]{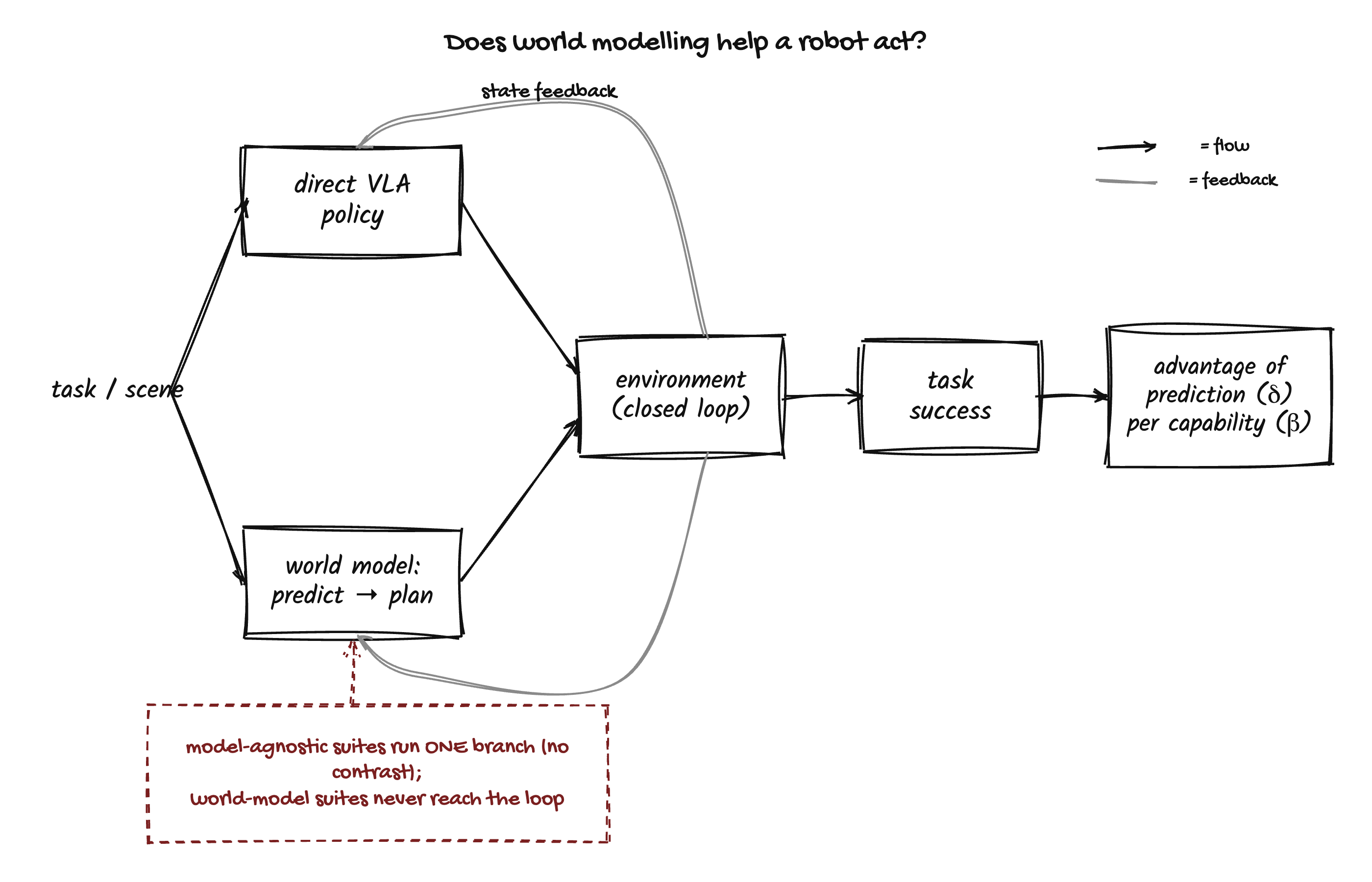}
\caption{\textbf{The evaluation loop that isolates the advantage of prediction.} A benchmark that
can answer ``does world modelling help a robot act'' must run a direct Vision-Language-Action policy
\emph{and} a world-model policy (predict then plan) under one closed loop, then report the
closed-loop gain ($\delta$) sliced by capability ($\beta$). Today's suites break this loop in one of
two ways: model-agnostic policy suites host a single policy and never build the contrast, while
world-model suites score the prediction open-loop and never execute it (\S\ref{sec:gap}).}
\Description{A left-to-right diagram. A task/scene forks into two branches, a direct VLA policy and a
world model that predicts then plans; both feed an environment closed loop with state feedback,
producing task success, which feeds an advantage-of-prediction measurement sliced per capability. An
annotation notes that model-agnostic suites run only one branch and world-model suites never close
the loop, so the contrast is missing.}
\label{fig:eval_loop}
\end{figure}

%% file: tables/tab_wm_senses.tex
\begin{table*}[t]
\centering
\footnotesize
\setlength{\tabcolsep}{3.5pt}
\renewcommand{\arraystretch}{1.25}
\resizebox{\textwidth}{!}{%
\begin{tabular}{@{}p{0.16\textwidth} p{0.23\textwidth} p{0.18\textwidth} c c c c@{}}
\toprule
\textbf{Sense of ``world model''} & \textbf{What it is} & \textbf{Examples} &
\textbf{Act.\ cond.} & \textbf{Loop} & \textbf{Metric} & \textbf{VLA-cmp.} \\
\midrule
\multicolumn{7}{@{}l}{\rule{0pt}{2.4ex}\emph{As a representation (what the model is)}}\\
latent WM (LWM) & latent-dynamics predictor $z_{t+1}=f(z_t,a_t)$ & DreamerV3, TD-MPC2 & \yy & \yy & \pp & \pp \\
action-cond.\ WM (WAM) & action-conditioned future-frame predictor & iVideoGPT, GR-2 & \yy & \pp & \pp & \pp \\
generative WM (WFM) & text / image-to-video generator & Sora, Cosmos, Genie & \pp & \nn & \yy & \nn \\
\multicolumn{7}{@{}l}{\rule{0pt}{2.6ex}\emph{As a role (what the model is used for)}}\\
neural simulator & learned environment for policy training / eval & UniSim & \yy & \yy & \pp & \pp \\
evaluation target & benchmarks that \emph{score} world models & WorldSimBench, WorldModelBench, EWMBench & \dm & \pp & \yy & \nn \\
\bottomrule
\end{tabular}}
\caption{The term ``world model'' spans two orthogonal axes the literature routinely conflates: a
\emph{representation} axis (what the model is: latent-state dynamics, action-conditioned video, or
generative video, a soft boundary since action-conditioned generators straddle WAM and WFM) and a
\emph{role} axis (what it is used for: a neural simulator, or the object a benchmark scores).
Columns record whether a sense is \textbf{act}ion-\textbf{cond}itioned, is used in a closed
\textbf{loop}, carries a standard evaluation \textbf{metric}, and is readily \textbf{VLA-comp}arable
under one protocol (\yy\ yes, \pp\ partial, \nn\ no, \dm\ n/a). No single sense is action-conditioned,
closed-loop, and VLA-comparable at once, which is why a benchmark that contrasts a world-model policy
with a VLA policy per capability (\S\ref{sec:gap}) is still missing. Evaluation benchmarks score the
world model as an \emph{object}; they are not world models acting as judges.}
\label{tab:wm_senses}
\end{table*}

%% file: tables/tab_branch_limits.tex
\begin{table*}[t]
\centering
\scriptsize
\setlength{\tabcolsep}{3pt}
\renewcommand{\arraystretch}{1.45}
\resizebox{\textwidth}{!}{%
\begin{tabular}{@{}>{\raggedright\arraybackslash}p{0.135\textwidth} >{\raggedright\arraybackslash}p{0.195\textwidth} >{\raggedright\arraybackslash}p{0.195\textwidth} >{\raggedright\arraybackslash}p{0.195\textwidth} >{\raggedright\arraybackslash}p{0.195\textwidth}@{}}
\toprule
 & \textbf{Policy suites}~\cite{libero} & \textbf{Embodied nav / social}~\cite{habitat30}
 & \textbf{World-model eval}~\cite{worldmodelbench} & \textbf{Bridges}~\cite{worldinworld} \\
\midrule
\textbf{Loop closure} &
\yy\ closed-loop; steps the environment~\cite{calvin} &
\yy\ closed-loop; long-horizon rollouts~\cite{behavior1k} &
\nn\ open-loop; scores generation, no action~\cite{physicsiq} &
\pp\ converts a prediction into an executed action~\cite{worldsimbench} \\
\textbf{Task horizon} &
\pp\ short reactive skills; long by chaining~\cite{libero} &
\yy\ long-horizon household and dialog~\cite{teach} &
\pp\ short clips; temporal but not task-level~\cite{ewmbench} &
\pp\ a single executed rollout~\cite{robowmbench} \\
\textbf{Capability breadth} &
\pp\ manipulation + transfer; hidden-state rare~\cite{thecolosseum} &
\yy\ navigation, social, some hidden-state~\cite{robothor} &
\pp\ physical / temporal; counterfactual barely~\cite{worldprediction} &
\nn\ narrow; executability only~\cite{worldarena} \\
\textbf{Model-family contrast} &
\nn\ model-agnostic host of any policy~\cite{roboarena} &
\pp\ memory-vs-Markovian in one suite~\cite{liberomem} &
\nn\ world-model-only scores~\cite{worldmodelbench} &
\pp\ WM as policy substrate; no VLA pole~\cite{worldinworld} \\
\textbf{Reproducibility} &
\pp\ sim; some real-vs-sim transfer~\cite{simpler} &
\pp\ sim-heavy; a sim-to-real gap~\cite{robothor} &
\yy\ fixed offline sets; deterministic scoring~\cite{vbench20} &
\nn\ new, still-unstandardised protocols~\cite{worldarena} \\
\textbf{Structural blind spot} &
hosts a world-model policy but never builds the VLA contrast &
capability cuts exist, but no per-capability family ablation &
visual quality is not task success; no behavioural utility &
reaches action, but omits the per-capability VLA head-to-head \\
\bottomrule
\end{tabular}}
\caption{\textbf{Evaluation-lane comparison matrix}: the four benchmark lanes on six evaluation
axes (loop closure, task horizon, capability breadth, model-family contrast, reproducibility, and
characteristic blind spot). Marks grade each lane on the axis (\yy\ favourable, \pp\ mixed, \nn\
weak), with an anchoring benchmark per cell. No lane closes the loop, slices capability, \emph{and}
builds a native VLA-vs-world-model contrast at once; the bridge lane comes closest yet still omits
the per-capability VLA head-to-head, which is the empty cell this survey names
(\S\ref{sec:gap}).}
\label{tab:branch_limits}
\end{table*}

%% file: sections/5_future.tex
\section{An agenda for advantage-aware evaluation}
\label{sec:future}

If the gap is that no benchmark is built to ask whether prediction helps acting, the remedy is a
small, measurable protocol that any benchmark could adopt. Table~\ref{tab:future_metrics} sets out
four quantities, each with a definition, a concrete testbed, and the axis it stresses, and
Figure~\ref{fig:future_protocol} sketches how they compose. The figure shows target shapes, not
measured results: it is a protocol, not a leaderboard.

The first quantity is an \emph{advantage-of-prediction curve}: the closed-loop success gain of a
world-model policy over a matched direct-VLA baseline, reported per capability as a curve rather than
a single endpoint, instantiated over LIBERO~\cite{libero} and CALVIN~\cite{calvin} run under one
harness. The second is \emph{counterfactual accuracy}: task success on episodes that require
reasoning over unobserved or hypothetical dynamics, built from WorldPrediction~\cite{worldprediction}
together with occlusion and memory splits in the style of LIBERO-Mem~\cite{liberomem}; this targets
the capability the corpus most neglects. The third is the \emph{prediction-to-action fidelity gap}:
the difference between a world model's open-loop generation quality and its closed-loop task success,
measured on bridge protocols such as WorldSimBench~\cite{worldsimbench} and
World-in-World~\cite{worldinworld}, which quantifies directly the ``visual quality is not task
success'' warning of Section~\ref{sec:gap}. The fourth is \emph{per-capability contrast coverage}:
the fraction of capabilities for which a benchmark actually runs a VLA-versus-world-model head-to-head,
computed by auditing the landscape (Table~\ref{tab:landscape}) against a declared contrast schema.
Each is reported as a distribution or a curve, never as one number, so that ``prediction helps'' can
be replaced by ``prediction helps \emph{here}, by \emph{this much}''.

\input{tables/tab_future_metrics}
\input{figures/fig_future_protocol}

\FloatBarrier

%% file: tables/tab_future_metrics.tex
\begin{table*}[t]
\centering
\footnotesize
\setlength{\tabcolsep}{4pt}
\renewcommand{\arraystretch}{1.35}
\resizebox{\textwidth}{!}{%
\begin{tabular}{@{}p{0.19\textwidth} p{0.30\textwidth} p{0.27\textwidth} p{0.15\textwidth}@{}}
\toprule
\textbf{Metric} & \textbf{Definition} & \textbf{Instantiation (testbed)} & \textbf{Axis stressed} \\
\midrule
Advantage-of-prediction curve &
closed-loop success \emph{gain} of a predictive policy over a matched direct-VLA baseline,
reported per capability as a curve, not a single endpoint &
a capability-sliced protocol over LIBERO~\cite{libero} and CALVIN~\cite{calvin} that runs a
world-model policy and a VLA policy under one harness &
$\gamma$ family, $\delta$ metric \\
Counterfactual accuracy &
task success on episodes that require inference over unobserved or hypothetical dynamics
(occlusion, object permanence, ``what if'') &
WorldPrediction~\cite{worldprediction} plus occlusion / memory splits in the style of
LIBERO-Mem~\cite{liberomem} &
$\beta$ capability \\
Prediction-to-action fidelity gap &
the difference between a world model's open-loop generation quality and its closed-loop task
success (``visual quality is not task success'') &
bridge protocols such as WorldSimBench~\cite{worldsimbench} and
World-in-World~\cite{worldinworld} &
$\alpha$ mode \\
Per-capability contrast coverage &
the fraction of capabilities for which a benchmark actually runs a VLA-vs-world-model
head-to-head under a single protocol &
audits of the landscape (Table~\ref{tab:landscape}) against a declared contrast schema &
$\gamma$ family, $\varepsilon$ catalog \\
\bottomrule
\end{tabular}}
\caption{\textbf{An actionable protocol for the future agenda}: four measurable quantities, their
definitions, and concrete testbeds. The first anchors the advantage-of-prediction question
($\delta$) on LIBERO / CALVIN-style suites; the second targets the near-unmeasured counterfactual
capability; the third quantifies the prediction-to-action gap the bridges expose; the fourth audits
per-capability contrast coverage across the landscape. Each is reported as a distribution or curve,
never a single headline number.}
\label{tab:future_metrics}
\end{table*}

%% file: figures/fig_future_protocol.tex
\begin{figure}[t]
\centering
\resizebox{\textwidth}{!}{%
\begin{tabular}{@{}c@{\hspace{9mm}}c@{}}
\begin{tikzpicture}[baseline]
  \draw[->,gray] (0,0)--(3.8,0) node[right,font=\scriptsize\bfseries,text=black]{capability};
  \draw[->,gray] (0,0)--(0,2.4);
  \node[rotate=90,font=\scriptsize\bfseries,anchor=south,text=gray!10!black] at (-0.2,1.25){closed-loop success};
  \fill[secE!12] (0,1.25) .. controls (1.5,1.85) and (2.4,1.7) .. (3.5,1.65)
                 -- (3.5,0.8) .. controls (2.4,0.95) and (1.5,1.0) .. (0,0.9) -- cycle;
  \draw[gray,line width=1pt,dashed] (0,0.9) .. controls (1.5,1.0) and (2.4,0.95) .. (3.5,0.8);
  \draw[secE!70!black,line width=1.3pt] (0,1.25) .. controls (1.5,1.85) and (2.4,1.7) .. (3.5,1.65);
  \node[secE!75!black,font=\scriptsize\bfseries,anchor=west] at (1.0,1.95){world-model policy};
  \node[gray!10!black,font=\scriptsize\bfseries,anchor=west] at (1.1,0.62){direct VLA baseline};
  \node[font=\footnotesize\bfseries] at (1.9,2.75){(a) Advantage of prediction ($\delta$)};
  \node[font=\scriptsize\bfseries,text=gray!10!black] at (1.9,-0.42){testbed: LIBERO / CALVIN, both families};
\end{tikzpicture}
&
\begin{tikzpicture}[baseline]
  \draw[->,gray] (0,0)--(3.8,0); \draw[->,gray] (0,0)--(0,2.4);
  \node[rotate=90,font=\scriptsize\bfseries,anchor=south,text=gray!10!black] at (-0.2,1.25){accuracy};
  \fill[cat-vla!55!black] (0.7,0) rectangle (1.5,0.4);
  \node[font=\scriptsize\bfseries] at (1.1,-0.22){today};
  \fill[cat-skill!38!black] (2.3,0) rectangle (3.1,1.95);
  \node[font=\scriptsize\bfseries] at (2.7,-0.22){target};
  \draw[gray,dashed,line width=0.6pt] (0,0.4)--(3.5,0.4) node[right,font=\scriptsize\bfseries,text=gray!10!black]{chance};
  \node[font=\footnotesize\bfseries] at (1.9,2.75){(b) Counterfactual accuracy};
  \node[font=\scriptsize\bfseries,text=gray!10!black] at (1.9,-0.42){testbed: WorldPrediction, LIBERO-Mem};
\end{tikzpicture}
\\[8mm]
\begin{tikzpicture}[baseline]
  \draw[->,gray] (0,0)--(3.8,0); \draw[->,gray] (0,0)--(0,2.4);
  \node[rotate=90,font=\scriptsize\bfseries,anchor=south,text=gray!10!black] at (-0.2,1.1){score};
  \fill[cat-skill!52!black] (0.6,0) rectangle (1.5,2.0);
  \fill[cat-vla!52!black] (2.2,0) rectangle (3.1,0.85);
  \draw[BrickRed,{Stealth}-{Stealth},line width=0.8pt] (1.5,2.0)--(2.2,2.0);
  \node[BrickRed,font=\scriptsize\bfseries,above] at (1.85,2.0){gap};
  \draw[BrickRed,dashed,line width=0.6pt] (1.5,2.0)--(3.1,2.0);
  \node[font=\scriptsize\bfseries,align=center] at (1.05,-0.32){open-loop\\quality};
  \node[font=\scriptsize\bfseries,align=center] at (2.65,-0.32){closed-loop\\success};
  \node[font=\footnotesize\bfseries] at (1.9,2.75){(c) Prediction-to-action gap};
  \node[font=\scriptsize\bfseries,text=gray!10!black] at (1.9,-0.72){testbed: WorldSimBench, World-in-World};
\end{tikzpicture}
&
\begin{tikzpicture}[baseline]
  \draw[->,gray] (0,0)--(3.8,0); \draw[->,gray] (0,0)--(0,2.4);
  \node[rotate=90,font=\scriptsize\bfseries,anchor=south,text=gray!10!black] at (-0.2,1.25){coverage};
  \foreach [count=\i] \l/\v/\c in {{hid.}/0.3/cat-code!62!black, {c-fact}/0.1/cat-skill!48!black,
                                   {long}/0.45/ForestGreen!68!black, {soc.}/0.2/BrickRed!72!black}{
    \fill[\c] ({\i*0.78-0.2},0) rectangle ({\i*0.78+0.2},{\v});
    \node[font=\scriptsize\bfseries] at ({\i*0.78},-0.22) {\l};
  }
  \draw[BrickRed,dashed,line width=0.6pt] (0,2.0)--(3.5,2.0) node[right,font=\scriptsize\bfseries,text=BrickRed]{target};
  \node[font=\footnotesize\bfseries] at (1.7,2.75){(d) Per-capability contrast};
  \node[font=\scriptsize\bfseries,text=gray!10!black] at (1.7,-0.52){testbed: landscape audit};
\end{tikzpicture}
\end{tabular}}
\caption{\textbf{An actionable evaluation protocol for the future agenda} (Table~\ref{tab:future_metrics}).
The panels are \emph{schematic}: they show the axes and the \emph{target} shape of each measurement,
not measured results. (a)~a world-model policy should beat a matched direct-VLA baseline in
closed-loop success, and the per-capability gap is the advantage of prediction ($\delta$), on
LIBERO / CALVIN-style suites run with both families; (b)~counterfactual accuracy is near chance today
and should rise, on WorldPrediction and occlusion splits; (c)~a world model's open-loop quality far
exceeds its closed-loop task success, and that gap must be reported, on bridge suites; (d)~the
fraction of capabilities with a native VLA-vs-world-model contrast is far below target across the
landscape.}
\Description{Four schematic panels showing target measurement shapes, not real data. (a) Two curves
over capability, a world-model policy above a dashed direct-VLA baseline, with the gap shaded as the
advantage of prediction. (b) A short today bar near a chance line versus a tall target bar for
counterfactual accuracy. (c) A tall open-loop quality bar and a short closed-loop success bar with a
red gap arrow. (d) Four short per-capability contrast-coverage bars well below a dashed target line.}
\label{fig:future_protocol}
\end{figure}
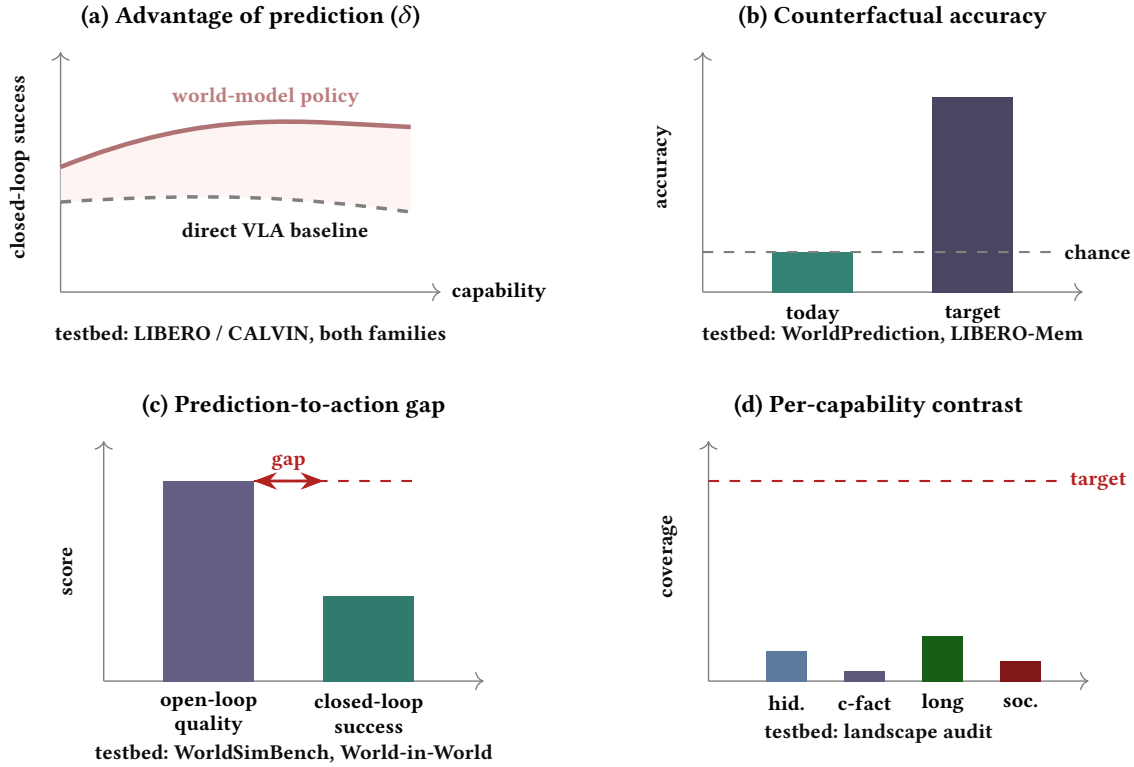

%% file: sections/limitations.tex
\section{Limitations}
\label{sec:limitations}

Three limits bound the survey's claims. First, its unit is the benchmark, not the model: the corpus
maps how the field \emph{measures} predictive embodied intelligence and deliberately says nothing
about which world models or VLA policies are strongest. A reader looking for a model ranking will not
find one here, by design. Second, the contrast axis rests on an operational definition
(Table~\ref{tab:defs}): a benchmark counts as building a VLA-versus-world-model contrast only when
both branches are run under one protocol. A more permissive reading would move some benchmarks from
model-agnostic to partial, though the qualitative picture, that explicit contrasts are rare and
counterfactual coverage is thin, is robust to the exact threshold. Third, coverage is biased towards
recent, arXiv-indexed, English-language work reachable by web search; the funnel in
Figure~\ref{fig:prisma} records only logged stages, and the two galleries include only benchmarks
whose figures could be located and provenance-checked, so absence from a gallery is not evidence of
absence of results. Finally, the advantage-of-prediction instruments discussed in
Section~\ref{sec:gap} are assessed as candidates rather than adopted; we do not rank benchmarks by any
metric that the field has not yet validated.

%% file: sections/conclusion.tex
\section{Conclusion}
\label{sec:conclusion}

Do world models make better robots? On the evidence of the benchmarks the field has built, the honest
answer is that we cannot yet tell. The question asks for a closed-loop, per-capability comparison
between a world-model policy and a direct VLA policy, and our catalogue of 160 web-verified benchmarks
shows that almost none is built to run it: 138 of 160 are model-agnostic, only 11 build an explicit
contrast, counterfactual capability is nearly unmeasured, and only four benchmarks carry prediction
into executed action. The obstacle is methodological. World-model suites score prediction without
executing it, task-success suites execute a single policy without contrasting families, and the term
``world model'' itself names three different objects, none of which is action-conditioned,
closed-loop, and VLA-comparable at once.

This survey's contribution is to make that gap precise and measurable. We give an operational
taxonomy of the landscape across four evaluation lanes, a coverage comparison showing that no prior
survey crosses capability with model family, an evaluation loop that isolates the advantage of
prediction, and a four-metric protocol, anchored on named testbeds, that any benchmark could adopt to
report where and by how much prediction helps. The next benchmark that matters will not be the one
with the most tasks or the most realistic renders; it will be the one that runs both branches in one
loop and reads off the difference. Building it is the empty cell this survey names.

%% file: sections/appendix.tex
\section{The full benchmark landscape}
\label{app:landscape}

The core taxonomy (Figure~\ref{fig:taxonomy_main}) shows 86 representative benchmarks, those that are
cleanly axis-placeable, distinct, and fully characterisable from their papers. For completeness and
reproducibility, Table~\ref{tab:landscape} lists the full catalogue of 160 web-verified benchmarks,
grouped by the same four evaluation lanes and sub-grouped by capability, and compared on year,
capability, evaluation mode, and whether the benchmark builds a native VLA-versus-world-model
contrast. Every entry is a real, individually citeable benchmark; the column tallies are the source
of the counts reported throughout the survey and of the distributions in
Figure~\ref{fig:corpus_dist}.

\input{tables/tab_landscape}

\section{Provenance of the galleries and collage}
\label{app:provenance}

Every panel in the subject and results galleries (Figures~\ref{fig:subject_gallery},
\ref{fig:results_gallery}) and every tile in the hero collage (Figure~\ref{fig:corpus_collage}) is a
real figure reproduced from the benchmark's own paper, hyperlinked to its arXiv source. The tables
below record, for each panel and tile, its source benchmark and the verified figure number within
that paper; a panel whose figure could not be located was dropped rather than guessed.
Table~\ref{tab:panel_provenance} covers the subject gallery, Table~\ref{tab:results_provenance} the
results gallery, and Table~\ref{tab:collage_provenance} the collage.

\input{tables/tab_panel_provenance}
\input{tables/tab_results_provenance}
\input{tables/tab_collage_provenance}

%% file: tables/tab_landscape.tex
{\scriptsize
\setlength{\tabcolsep}{4pt}
\renewcommand{\arraystretch}{1.12}
\begin{longtable}{@{}p{0.35\textwidth} c >{\raggedright\arraybackslash}p{0.43\textwidth} c c@{}}
\caption{\textbf{The robot-evaluation benchmark landscape: 160 web-verified benchmarks, compared on four axes.} Grouped into four evaluation lanes and sub-grouped by capability. Columns: \textbf{Year}; \textbf{Capability} (what the benchmark stresses); \textbf{Eval} (evaluation mode: open-loop prediction/generation quality, closed-loop task success, or a prediction-to-action bridge); and \textbf{VLA vs WM} (does the benchmark build a native direct-policy vs world-model contrast: \yy\ explicit, \pp\ partial, \dm\ model-agnostic). Every entry is a real, individually citeable benchmark.}\label{tab:landscape}\\
\toprule
\textbf{Benchmark} & \textbf{Year} & \textbf{Capability} & \textbf{Eval} & \textbf{VLA vs WM}\\
\midrule
\endfirsthead
\multicolumn{5}{@{}l}{\scriptsize\emph{Table~\ref{tab:landscape} (continued)}}\\
\toprule
\textbf{Benchmark} & \textbf{Year} & \textbf{Capability} & \textbf{Eval} & \textbf{VLA vs WM}\\
\midrule
\endhead
\bottomrule
\endlastfoot
\multicolumn{5}{@{}l}{\rule{0pt}{2.8ex}\textbf{\textsf{Policy / manipulation suites}}~\textcolor{gray}{(37)}}\\[1pt]
\multicolumn{5}{@{}l}{\hspace{0.6em}\emph{Manipulation}}\\
\hspace{0.6em}\textbf{ARNOLD}~\cite{arnold} & 2023 & language-grounded continuous-state 3D manipulation & \textsf{closed} & \dm\\
\hspace{0.6em}\textbf{BiGym}~\cite{bigym} & 2024 & mobile bimanual manipulation & \textsf{closed} & \dm\\
\hspace{0.6em}\textbf{FMB}~\cite{fmb} & 2024 & real-world functional multi-step assembly manipulation & \textsf{closed} & \dm\\
\hspace{0.6em}\textbf{Franka Kitchen}~\cite{frankakitchen} & 2019 & multi-task kitchen manipulation & \textsf{closed} & \dm\\
\hspace{0.6em}\textbf{FurnitureBench}~\cite{furniturebench} & 2023 & real-world furniture assembly & \textsf{closed} & \dm\\
\hspace{0.6em}\textbf{GRUtopia}~\cite{grutopia} & 2024 & city-scale embodied navigation and manipulation & \textsf{closed} & \dm\\
\hspace{0.6em}\textbf{KitchenShift} & 2021 & kitchen distribution-shift generalization & \textsf{closed} & \dm\\
\hspace{0.6em}\textbf{ManiSkill2}~\cite{maniskill2} & 2023 & generalizable skills & \textsf{closed} & \dm\\
\hspace{0.6em}\textbf{PerAct}~\cite{peract} & 2022 & language-conditioned 6-DoF single-arm manipulation & \textsf{closed} & \dm\\
\hspace{0.6em}\textbf{PerAct2}~\cite{peract2} & 2024 & bimanual language-conditioned manipulation & \textsf{closed} & \dm\\
\hspace{0.6em}\textbf{Ravens}~\cite{ravens} & 2020 & vision-based pick-and-place rearrangement & \textsf{closed} & \dm\\
\hspace{0.6em}\textbf{RGB-Stacking}~\cite{rgbstacking} & 2021 & vision-based stacking of diverse shapes & \textsf{closed} & \dm\\
\hspace{0.6em}\textbf{RLBench}~\cite{rlbench} & 2020 & manipulation breadth & \textsf{closed} & \dm\\
\hspace{0.6em}\textbf{RoboAgent}~\cite{roboagent} & 2023 & multi-task manipulation with RoboSet & \textsf{closed} & \dm\\
\hspace{0.6em}\textbf{robomimic}~\cite{robomimic} & 2021 & single-arm manipulation from human demonstrations & \textsf{closed} & \dm\\
\hspace{0.6em}\textbf{robosuite}~\cite{robosuite} & 2020 & modular simulated single-arm manipulation tasks & \textsf{closed} & \dm\\
\hspace{0.6em}\textbf{RoboTwin}~\cite{robotwin} & 2024 & dual-arm bimanual manipulation & \textsf{closed} & \dm\\
\hspace{0.6em}\textbf{VLMbench}~\cite{vlmbench} & 2022 & vision-and-language compositional manipulation & \textsf{closed} & \dm\\
\multicolumn{5}{@{}l}{\hspace{0.6em}\emph{Generalization \& robustness}}\\
\hspace{0.6em}\textbf{GemBench}~\cite{gembench} & 2024 & generalization levels & \textsf{closed} & \dm\\
\hspace{0.6em}\textbf{LIBERO}~\cite{libero} & 2023 & short-horizon + transfer & \textsf{closed} & \dm\\
\hspace{0.6em}\textbf{RoboArena}~\cite{roboarena} & 2025 & real-world generalization & \textsf{closed} & \dm\\
\hspace{0.6em}\textbf{SIMPLER / SimplerEnv}~\cite{simpler} & 2024 & generalization (real vs sim) & \textsf{closed} & \dm\\
\hspace{0.6em}\textbf{THE COLOSSEUM}~\cite{thecolosseum} & 2024 & generalization / robustness & \textsf{closed} & \dm\\
\hspace{0.6em}\textbf{VIMA-Bench}~\cite{vimabench} & 2023 & multimodal-prompt generalization & \textsf{closed} & \dm\\
\multicolumn{5}{@{}l}{\hspace{0.6em}\emph{Other}}\\
\hspace{0.6em}\textbf{CortexBench}~\cite{cortexbench} & 2023 & pretrained visual representations for embodied control & \textsf{closed} & \dm\\
\hspace{0.6em}\textbf{GenManip}~\cite{genmanip} & 2025 & LLM-scene tasks & \textsf{closed} & \pp\\
\hspace{0.6em}\textbf{HandoverSim}~\cite{handoversim} & 2022 & human-to-robot object handover & \textsf{closed} & \dm\\
\hspace{0.6em}\textbf{Meta-World}~\cite{metaworld} & 2020 & multi-task / meta-RL & \textsf{closed} & \dm\\
\hspace{0.6em}\textbf{RoboHive}~\cite{robohive} & 2023 & unified robot-learning environments & \textsf{closed} & \dm\\
\multicolumn{5}{@{}l}{\hspace{0.6em}\emph{Dexterous manipulation}}\\
\hspace{0.6em}\textbf{Bi-DexHands}~\cite{bidexhands} & 2022 & bimanual dexterous manipulation & \textsf{closed} & \dm\\
\hspace{0.6em}\textbf{DeformableGym} & 2023 & 3D deformable-object grasping & \textsf{closed} & \dm\\
\hspace{0.6em}\textbf{DexArt}~\cite{dexart} & 2023 & dexterous articulated-object manipulation & \textsf{closed} & \dm\\
\hspace{0.6em}\textbf{UniDexGrasp}~\cite{unidexgrasp} & 2023 & universal dexterous grasping & \textsf{closed} & \dm\\
\multicolumn{5}{@{}l}{\hspace{0.6em}\emph{Long-horizon \& planning}}\\
\hspace{0.6em}\textbf{CALVIN}~\cite{calvin} & 2021 & long-horizon language & \textsf{closed} & \dm\\
\hspace{0.6em}\textbf{RoboCasa}~\cite{robocasa} & 2024 & long-horizon (data scaling) & \textsf{closed} & \dm\\
\hspace{0.6em}\textbf{VLABench}~\cite{vlabench} & 2025 & long-horizon reasoning & \textsf{closed} & \dm\\
\multicolumn{5}{@{}l}{\hspace{0.6em}\emph{Deformable-object manipulation}}\\
\hspace{0.6em}\textbf{SoftGym}~\cite{softgym} & 2020 & deformable-object manipulation & \textsf{closed} & \dm\\
\multicolumn{5}{@{}l}{\rule{0pt}{2.8ex}\textbf{\textsf{Embodied navigation, planning \& social}}~\textcolor{gray}{(85)}}\\[1pt]
\multicolumn{5}{@{}l}{\hspace{0.6em}\emph{Long-horizon \& planning}}\\
\hspace{0.6em}\textbf{Alexa Arena}~\cite{alexaarena} & 2023 & task-planning & \textsf{closed} & \dm\\
\hspace{0.6em}\textbf{ALFRED}~\cite{alfred} & 2019 & long-horizon, partial-obs & \textsf{closed} & \dm\\
\hspace{0.6em}\textbf{BEHAVIOR-1K}~\cite{behavior1k} & 2024 & long-horizon household & \textsf{closed} & \dm\\
\hspace{0.6em}\textbf{CoELA (C-WAH/TDW-MAT)}~\cite{coela} & 2023 & long-horizon & \textsf{closed} & \dm\\
\hspace{0.6em}\textbf{DialFRED}~\cite{dialfred} & 2022 & task-planning & \textsf{closed} & \dm\\
\hspace{0.6em}\textbf{EgoPlan-Bench}~\cite{egoplanbench} & 2023 & task-planning & \textsf{closed} & \dm\\
\hspace{0.6em}\textbf{EmbodiedBench}~\cite{embodiedbench} & 2025 & long-horizon & \textsf{closed} & \dm\\
\hspace{0.6em}\textbf{EmbodiedCity}~\cite{embodiedcity} & 2024 & long-horizon & \textsf{closed} & \dm\\
\hspace{0.6em}\textbf{EmbodiedEval}~\cite{embodiedeval} & 2025 & long-horizon & \textsf{closed} & \dm\\
\hspace{0.6em}\textbf{HomeRobot OVMM}~\cite{homerobotovmm} & 2023 & long-horizon & \textsf{closed} & \dm\\
\hspace{0.6em}\textbf{LEGENT}~\cite{legent} & 2024 & long-horizon & \textsf{closed} & \dm\\
\hspace{0.6em}\textbf{LH-VLN}~\cite{lhvln} & 2024 & long-horizon & \textsf{closed} & \dm\\
\hspace{0.6em}\textbf{LoHoRavens}~\cite{lohoravens} & 2023 & long-horizon & \textsf{closed} & \dm\\
\hspace{0.6em}\textbf{LoTa-Bench}~\cite{lotabench} & 2024 & task-planning & \textsf{closed} & \dm\\
\hspace{0.6em}\textbf{MFE-ETP}~\cite{mfeetp} & 2024 & task-planning & \textsf{closed} & \dm\\
\hspace{0.6em}\textbf{PARTNR}~\cite{partnr} & 2024 & task-planning & \textsf{closed} & \dm\\
\hspace{0.6em}\textbf{PlanBench}~\cite{planbench} & 2022 & task-planning & \textsf{closed} & \dm\\
\hspace{0.6em}\textbf{SafeAgentBench}~\cite{safeagentbench} & 2024 & task-planning & \textsf{closed} & \dm\\
\hspace{0.6em}\textbf{SMART-LLM}~\cite{smartllm} & 2023 & task-planning & \textsf{closed} & \dm\\
\hspace{0.6em}\textbf{TDW-Transport}~\cite{tdwtransport} & 2021 & task-planning & \textsf{closed} & \dm\\
\hspace{0.6em}\textbf{TidyBot}~\cite{tidybot} & 2023 & task-planning & \textsf{closed} & \dm\\
\hspace{0.6em}\textbf{VirtualHome}~\cite{virtualhome} & 2018 & task-planning & \textsf{closed} & \dm\\
\hspace{0.6em}\textbf{Watch-And-Help}~\cite{watchandhelp} & 2020 & long-horizon & \textsf{closed} & \dm\\
\multicolumn{5}{@{}l}{\hspace{0.6em}\emph{Social \& multi-agent}}\\
\hspace{0.6em}\textbf{CrowdNav}~\cite{crowdnav} & 2018 & social & \textsf{closed} & \dm\\
\hspace{0.6em}\textbf{Habitat 3.0}~\cite{habitat30} & 2023 & social / multi-agent & \textsf{closed} & \dm\\
\hspace{0.6em}\textbf{HuNavSim}~\cite{hunavsim} & 2023 & social & \textsf{closed} & \dm\\
\hspace{0.6em}\textbf{JRDB}~\cite{jrdb} & 2019 & social & \textsf{closed} & \dm\\
\hspace{0.6em}\textbf{JRDB-Act}~\cite{jrdbact} & 2021 & social & \textsf{closed} & \dm\\
\hspace{0.6em}\textbf{Melting Pot}~\cite{meltingpot} & 2021 & social & \textsf{closed} & \dm\\
\hspace{0.6em}\textbf{Overcooked-AI}~\cite{overcookedai} & 2019 & social & \textsf{closed} & \dm\\
\hspace{0.6em}\textbf{RoboSense}~\cite{robosense} & 2024 & social & \textsf{closed} & \dm\\
\hspace{0.6em}\textbf{SEAN}~\cite{sean} & 2020 & social & \textsf{closed} & \dm\\
\hspace{0.6em}\textbf{SMAC}~\cite{smac} & 2019 & social & \textsf{closed} & \dm\\
\hspace{0.6em}\textbf{SocialGym 2.0}~\cite{socialgym20} & 2023 & social & \textsf{closed} & \dm\\
\hspace{0.6em}\textbf{SocNavBench}~\cite{socnavbench} & 2021 & social navigation & \textsf{closed} & \dm\\
\multicolumn{5}{@{}l}{\hspace{0.6em}\emph{Navigation}}\\
\hspace{0.6em}\textbf{Gibson Env}~\cite{gibsonenv} & 2018 & navigation & \textsf{closed} & \dm\\
\hspace{0.6em}\textbf{GOAT-Bench}~\cite{goatbench} & 2024 & multi-modal lifelong navigation & \textsf{closed} & \dm\\
\hspace{0.6em}\textbf{Habitat}~\cite{habitat} & 2019 & navigation & \textsf{closed} & \dm\\
\hspace{0.6em}\textbf{Habitat 2.0}~\cite{habitat20} & 2021 & rearrangement & \textsf{closed} & \dm\\
\hspace{0.6em}\textbf{HM3D}~\cite{hm3d} & 2021 & navigation & \textsf{closed} & \dm\\
\hspace{0.6em}\textbf{iGibson}~\cite{igibson} & 2020 & interactive navigation & \textsf{closed} & \dm\\
\hspace{0.6em}\textbf{Matterport3D}~\cite{matterport3d} & 2017 & navigation & \textsf{closed} & \dm\\
\hspace{0.6em}\textbf{ObjectNav}~\cite{objectnav} & 2020 & object-goal navigation & \textsf{closed} & \dm\\
\hspace{0.6em}\textbf{ProcTHOR}~\cite{procthor} & 2022 & navigation & \textsf{closed} & \dm\\
\hspace{0.6em}\textbf{RoboTHOR}~\cite{robothor} & 2020 & navigation / sim-to-real & \textsf{closed} & \dm\\
\hspace{0.6em}\textbf{SoundSpaces}~\cite{soundspaces} & 2020 & audio-visual navigation & \textsf{closed} & \dm\\
\multicolumn{5}{@{}l}{\hspace{0.6em}\emph{Autonomous driving}}\\
\hspace{0.6em}\textbf{Bench2Drive}~\cite{bench2drive} & 2024 & driving & \textsf{closed} & \dm\\
\hspace{0.6em}\textbf{CARLA}~\cite{carla} & 2017 & driving & \textsf{closed} & \dm\\
\hspace{0.6em}\textbf{DriveArena}~\cite{drivearena} & 2024 & driving & \textsf{closed} & \yy\\
\hspace{0.6em}\textbf{MetaDrive}~\cite{metadrive} & 2021 & driving & \textsf{closed} & \dm\\
\hspace{0.6em}\textbf{NAVSIM}~\cite{navsim} & 2024 & driving & \textsf{closed} & \dm\\
\hspace{0.6em}\textbf{nuScenes}~\cite{nuscenes} & 2019 & driving & \textsf{closed} & \dm\\
\hspace{0.6em}\textbf{Waymax}~\cite{waymax} & 2023 & driving & \textsf{closed} & \dm\\
\hspace{0.6em}\textbf{Waymo Open}~\cite{waymoopen} & 2019 & driving & \textsf{closed} & \dm\\
\multicolumn{5}{@{}l}{\hspace{0.6em}\emph{Counterfactual reasoning}}\\
\hspace{0.6em}\textbf{ACRE}~\cite{acre} & 2021 & counterfactual & \textsf{open} & \yy\\
\hspace{0.6em}\textbf{CausalCity}~\cite{causalcity} & 2021 & counterfactual & \textsf{closed} & \pp\\
\hspace{0.6em}\textbf{CausalVQA}~\cite{causalvqa} & 2025 & counterfactual & \textsf{open} & \yy\\
\hspace{0.6em}\textbf{CausalWorld}~\cite{causalworld} & 2020 & counterfactual & \textsf{closed} & \yy\\
\hspace{0.6em}\textbf{CoPhy}~\cite{cophy} & 2019 & counterfactual & \textsf{open} & \yy\\
\hspace{0.6em}\textbf{CRAFT}~\cite{craft} & 2020 & counterfactual & \textsf{open} & \yy\\
\hspace{0.6em}\textbf{Filtered-CoPhy}~\cite{filteredcophy} & 2022 & counterfactual & \textsf{open} & \yy\\
\multicolumn{5}{@{}l}{\hspace{0.6em}\emph{Locomotion}}\\
\hspace{0.6em}\textbf{Barkour}~\cite{barkour} & 2023 & legged locomotion & \textsf{closed} & \dm\\
\hspace{0.6em}\textbf{Brax}~\cite{brax} & 2021 & locomotion & \textsf{closed} & \dm\\
\hspace{0.6em}\textbf{DERL}~\cite{derl} & 2021 & locomotion & \textsf{closed} & \dm\\
\hspace{0.6em}\textbf{HumanoidBench}~\cite{humanoidbench} & 2024 & humanoid locomotion & \textsf{closed} & \dm\\
\hspace{0.6em}\textbf{Isaac Gym}~\cite{isaacgym} & 2021 & locomotion & \textsf{closed} & \dm\\
\hspace{0.6em}\textbf{Legged Gym}~\cite{leggedgym} & 2021 & legged locomotion & \textsf{closed} & \dm\\
\multicolumn{5}{@{}l}{\hspace{0.6em}\emph{Theory of mind}}\\
\hspace{0.6em}\textbf{AGENT}~\cite{agent} & 2021 & theory-of-mind & \textsf{open} & \yy\\
\hspace{0.6em}\textbf{BIB}~\cite{bib} & 2021 & theory-of-mind & \textsf{open} & \yy\\
\hspace{0.6em}\textbf{PHASE}~\cite{phase} & 2021 & theory-of-mind & \textsf{closed} & \pp\\
\hspace{0.6em}\textbf{SocialAI}~\cite{socialai} & 2021 & theory-of-mind & \textsf{closed} & \dm\\
\hspace{0.6em}\textbf{ToMi} & 2019 & theory-of-mind & \textsf{closed} & \pp\\
\multicolumn{5}{@{}l}{\hspace{0.6em}\emph{Vision-language navigation}}\\
\hspace{0.6em}\textbf{R2R}~\cite{r2r} & 2018 & VLN & \textsf{closed} & \dm\\
\hspace{0.6em}\textbf{REVERIE}~\cite{reverie} & 2020 & VLN & \textsf{closed} & \dm\\
\hspace{0.6em}\textbf{RxR}~\cite{rxr} & 2020 & VLN & \textsf{closed} & \dm\\
\hspace{0.6em}\textbf{Touchdown}~\cite{touchdown} & 2019 & outdoor VLN & \textsf{closed} & \dm\\
\hspace{0.6em}\textbf{VLN-CE}~\cite{vlnce} & 2020 & VLN & \textsf{closed} & \dm\\
\multicolumn{5}{@{}l}{\hspace{0.6em}\emph{Hidden-state / object permanence}}\\
\hspace{0.6em}\textbf{Bongard-HOI}~\cite{bongardhoi} & 2022 & hidden-state & \textsf{open} & \pp\\
\hspace{0.6em}\textbf{ComPhy}~\cite{comphy} & 2022 & hidden-state & \textsf{open} & \yy\\
\hspace{0.6em}\textbf{LIBERO-Mem}~\cite{liberomem} & 2025 & hidden-state / occlusion & \textsf{closed} & \yy\\
\hspace{0.6em}\textbf{TEACh}~\cite{teach} & 2021 & hidden-state (dialog) & \textsf{closed} & \dm\\
\multicolumn{5}{@{}l}{\hspace{0.6em}\emph{Embodied question answering}}\\
\hspace{0.6em}\textbf{OpenEQA} & 2024 & EQA & \textsf{closed} & \dm\\
\hspace{0.6em}\textbf{RoboVQA}~\cite{robovqa} & 2023 & EQA & \textsf{closed} & \dm\\
\multicolumn{5}{@{}l}{\hspace{0.6em}\emph{Generalization \& robustness}}\\
\hspace{0.6em}\textbf{ALFWorld}~\cite{alfworld} & 2020 & abstract-vs-grounded transfer & \textsf{closed} & \pp\\
\multicolumn{5}{@{}l}{\hspace{0.6em}\emph{Other}}\\
\hspace{0.6em}\textbf{O-PIAAGETS} & 2022 & object permanence & \textsf{closed} & \pp\\
\multicolumn{5}{@{}l}{\rule{0pt}{2.8ex}\textbf{\textsf{World-model \& video-generation evaluation}}~\textcolor{gray}{(34)}}\\[1pt]
\multicolumn{5}{@{}l}{\hspace{0.6em}\emph{Generation quality}}\\
\hspace{0.6em}\textbf{CATER}~\cite{cater} & 2019 & compositional-action temporal reasoning & \textsf{open} & \dm\\
\hspace{0.6em}\textbf{DEVIL}~\cite{devil} & 2024 & content-dynamics quality of T2V models & \textsf{open} & \dm\\
\hspace{0.6em}\textbf{EVA-Bench}~\cite{evabench} & 2024 & embodied video anticipation & \textsf{open} & \dm\\
\hspace{0.6em}\textbf{EvalCrafter}~\cite{evalcrafter} & 2023 & generation quality (17-metric suite) & \textsf{open} & \dm\\
\hspace{0.6em}\textbf{EWMBench}~\cite{ewmbench} & 2025 & scene/motion/semantic quality & \textsf{open} & \dm\\
\hspace{0.6em}\textbf{FETV}~\cite{fetv} & 2023 & fine-grained text-to-video quality & \textsf{open} & \dm\\
\hspace{0.6em}\textbf{IPV-Bench}~\cite{ipvbench} & 2025 & impossible-video generation + understanding & \textsf{open} & \dm\\
\hspace{0.6em}\textbf{StoryBench}~\cite{storybench} & 2023 & continuous story visualization & \textsf{open} & \dm\\
\hspace{0.6em}\textbf{T2V-CompBench}~\cite{t2vcompbench} & 2024 & compositional text-to-video quality & \textsf{open} & \dm\\
\hspace{0.6em}\textbf{T2VScore}~\cite{t2vscore} & 2024 & text-to-video metric (alignment + quality) & \textsf{open} & \dm\\
\hspace{0.6em}\textbf{TC-Bench}~\cite{tcbench} & 2024 & temporal compositionality in video generation & \textsf{open} & \dm\\
\hspace{0.6em}\textbf{VBench}~\cite{vbench} & 2023 & generation quality (16 disentangled dimensions) & \textsf{open} & \dm\\
\hspace{0.6em}\textbf{VBench-2.0}~\cite{vbench20} & 2025 & generation faithfulness & \textsf{open} & \dm\\
\hspace{0.6em}\textbf{VideoCon}~\cite{videocon} & 2023 & robust video-language alignment & \textsf{open} & \dm\\
\hspace{0.6em}\textbf{VideoHallucer}~\cite{videohallucer} & 2024 & hallucination in video-language models & \textsf{open} & \dm\\
\hspace{0.6em}\textbf{VideoScore}~\cite{videoscore} & 2024 & learned automatic quality metric & \textsf{open} & \dm\\
\hspace{0.6em}\textbf{WorldModelBench}~\cite{worldmodelbench} & 2025 & video-WM prediction quality & \textsf{open} & \dm\\
\hspace{0.6em}\textbf{WorldScore}~\cite{worldscore} & 2025 & world-gen quality & \textsf{open} & \dm\\
\multicolumn{5}{@{}l}{\hspace{0.6em}\emph{Physical reasoning}}\\
\hspace{0.6em}\textbf{ContPhy}~\cite{contphy} & 2024 & continuum (soft-body/fluid) physical reasoning & \textsf{open} & \dm\\
\hspace{0.6em}\textbf{GRASP}~\cite{grasp} & 2023 & grounding + intuitive physics in video MLLMs & \textsf{open} & \dm\\
\hspace{0.6em}\textbf{IntPhys}~\cite{intphys} & 2018 & intuitive physics (violation-of-expectation) & \textsf{open} & \dm\\
\hspace{0.6em}\textbf{IntPhys 2}~\cite{intphys2} & 2025 & intuitive physics in complex scenes & \textsf{open} & \dm\\
\hspace{0.6em}\textbf{PHYBench}~\cite{phybench} & 2025 & physical perception and reasoning & \textsf{open} & \dm\\
\hspace{0.6em}\textbf{PhyGenBench}~\cite{phygenbench} & 2024 & physical-commonsense correctness in video gen & \textsf{open} & \dm\\
\hspace{0.6em}\textbf{PhysBench}~\cite{physbench} & 2025 & physical-world understanding for VLMs & \textsf{open} & \dm\\
\hspace{0.6em}\textbf{Physics-IQ}~\cite{physicsiq} & 2025 & physical realism & \textsf{open} & \dm\\
\hspace{0.6em}\textbf{Physion}~\cite{physion} & 2021 & physical prediction from vision & \textsf{open} & \dm\\
\hspace{0.6em}\textbf{Physion++}~\cite{physionpp} & 2023 & physical prediction with online property inference & \textsf{open} & \dm\\
\hspace{0.6em}\textbf{PhyWorldBench}~\cite{phyworldbench} & 2025 & physical realism in text-to-video & \textsf{open} & \dm\\
\hspace{0.6em}\textbf{VideoPhy}~\cite{videophy} & 2024 & physical commonsense in generated video & \textsf{open} & \dm\\
\hspace{0.6em}\textbf{VideoPhy-2}~\cite{videophy2} & 2025 & action-centric physical commonsense & \textsf{open} & \dm\\
\multicolumn{5}{@{}l}{\hspace{0.6em}\emph{Counterfactual reasoning}}\\
\hspace{0.6em}\textbf{CLEVRER}~\cite{clevrer} & 2019 & physical/causal video reasoning (counterfactual) & \textsf{open} & \dm\\
\hspace{0.6em}\textbf{WorldPrediction}~\cite{worldprediction} & 2025 & counterfactual action recognition & \textsf{open} & \dm\\
\multicolumn{5}{@{}l}{\hspace{0.6em}\emph{Autonomous driving}}\\
\hspace{0.6em}\textbf{Vista}~\cite{vista} & 2024 & driving world model & \textsf{open} & \dm\\
\multicolumn{5}{@{}l}{\rule{0pt}{2.8ex}\textbf{\textsf{Bridges (prediction to action)}}~\textcolor{gray}{(4)}}\\[1pt]
\multicolumn{5}{@{}l}{\hspace{0.6em}\emph{Other}}\\
\hspace{0.6em}\textbf{RoboWM-Bench}~\cite{robowmbench} & 2026 & prediction executability & \textsf{bridge} & \pp\\
\hspace{0.6em}\textbf{World-in-World}~\cite{worldinworld} & 2025 & closed-loop task success of WMs & \textsf{bridge} & \pp\\
\hspace{0.6em}\textbf{WorldArena}~\cite{worldarena} & 2026 & closed-loop WM arena & \textsf{bridge} & \pp\\
\multicolumn{5}{@{}l}{\hspace{0.6em}\emph{Generation quality}}\\
\hspace{0.6em}\textbf{WorldSimBench}~\cite{worldsimbench} & 2024 & video-to-action consistency & \textsf{bridge} & \pp\\
\end{longtable}
}

%% file: tables/tab_panel_provenance.tex
\begin{table*}[t]
\centering
\footnotesize
\setlength{\tabcolsep}{6pt}
\renewcommand{\arraystretch}{1.1}
\resizebox{\textwidth}{!}{%
\begin{tabular}{@{}l l l@{}}
\toprule
\textbf{Benchmark} & \textbf{arXiv} & \textbf{Panel source (subject scene)} \\
\midrule
\multicolumn{3}{@{}l}{\textbf{\textsf{Closed-loop task-success suites}}}\\
ALFRED & \href{https://arxiv.org/abs/1912.01734}{1912.01734} & sim task scene (rep.\ figure, pp.1--3) \\
BEHAVIOR-1K & \href{https://arxiv.org/abs/2403.09227}{2403.09227} & household sim scene (rep.\ figure, pp.1--3) \\
CALVIN & \href{https://arxiv.org/abs/2112.03227}{2112.03227} & tabletop manipulation scene (rep.\ figure, pp.1--3) \\
GemBench & \href{https://arxiv.org/abs/2410.01345}{2410.01345} & manipulation scene (rep.\ figure, pp.1--3) \\
ManiSkill2 & \href{https://arxiv.org/abs/2302.04659}{2302.04659} & multi-task sim scenes (rep.\ figure, pp.1--3) \\
Meta-World & \href{https://arxiv.org/abs/1910.10897}{1910.10897} & task-suite thumbnail grid (rep.\ figure, pp.1--3) \\
RoboCasa & \href{https://arxiv.org/abs/2406.02523}{2406.02523} & kitchen sim scene (rep.\ figure, pp.1--3) \\
LIBERO & \href{https://arxiv.org/abs/2306.03310}{2306.03310} & manipulation task scene (embedded raster, pp.1--8) \\
VLABench & \href{https://arxiv.org/abs/2412.18194}{2412.18194} & mahjong-tile manipulation scene (embedded raster, pp.1--8) \\
Habitat 3.0 & \href{https://arxiv.org/abs/2310.13724}{2310.13724} & human--robot collaboration scene (embedded raster, pp.1--8) \\
\midrule
\multicolumn{3}{@{}l}{\textbf{\textsf{Open-loop world-model \& video-generation evaluation}}}\\
EVA-Bench & \href{https://arxiv.org/abs/2410.15461}{2410.15461} & embodied-anticipation scene (rep.\ figure, pp.1--3) \\
EWMBench & \href{https://arxiv.org/abs/2505.09694}{2505.09694} & generated-video sample (rep.\ figure, pp.1--3) \\
Physics-IQ & \href{https://arxiv.org/abs/2501.09038}{2501.09038} & physical-realism sample (rep.\ figure, pp.1--3) \\
Physion & \href{https://arxiv.org/abs/2106.08261}{2106.08261} & Fig.~1 example-scenario frames (p.3) \\
Physion++ & \href{https://arxiv.org/abs/2306.15668}{2306.15668} & physical-prediction scene (rep.\ figure, pp.1--3) \\
VideoPhy & \href{https://arxiv.org/abs/2406.03520}{2406.03520} & generated-video sample (rep.\ figure, pp.1--3) \\
WorldModelBench & \href{https://arxiv.org/abs/2502.20694}{2502.20694} & generated-video sample (rep.\ figure, pp.1--3) \\
WorldScore & \href{https://arxiv.org/abs/2504.00983}{2504.00983} & generated world sample (rep.\ figure, pp.1--3) \\
VBench-2.0 & \href{https://arxiv.org/abs/2503.21755}{2503.21755} & generated-video sample (embedded raster, pp.1--8) \\
WorldPrediction & \href{https://arxiv.org/abs/2506.04363}{2506.04363} & procedural-action video frame (embedded raster, pp.1--8) \\
\midrule
\multicolumn{3}{@{}l}{\textbf{\textsf{Prediction-to-action bridges}}}\\
WorldSimBench & \href{https://arxiv.org/abs/2410.18072}{2410.18072} & video-prediction filmstrip, interactive-eval figure (embedded raster, pp.1--8) \\
World-in-World & \href{https://arxiv.org/abs/2510.18135}{2510.18135} & closed-loop manipulation scene (embedded raster, pp.1--8) \\
RoboWM-Bench & \href{https://arxiv.org/abs/2604.19092}{2604.19092} & manipulation rollout scene (embedded raster, pp.1--8) \\
WorldArena & \href{https://arxiv.org/abs/2602.08971}{2602.08971} & dual-arm manipulation scene (embedded raster, pp.1--8) \\
\bottomrule
\end{tabular}}
\caption{\textbf{Per-panel provenance for Figure~\ref{fig:subject_gallery}.} Every panel is a real
subject/scene figure cropped from the benchmark's own arXiv paper and hyperlinked to its source.
Exact figure numbers are given where verified from the paper (e.g.\ Physion Fig.~1); otherwise the
teaser-region basis is noted (``rep.\ figure, pp.1--3'' for tiles reused from the collage pool;
``embedded raster, pp.1--8'' for the nine gap tiles hand-picked from per-paper candidate contact
sheets after a subject/results/neither classification pass). No figure number is invented. Images
\textcopyright\ their respective authors, reproduced for scholarly review.}
\label{tab:panel_provenance}
\end{table*}

%% file: tables/tab_results_provenance.tex
\begin{table*}[t]
\centering
\footnotesize
\setlength{\tabcolsep}{6pt}
\renewcommand{\arraystretch}{1.1}
\resizebox{\textwidth}{!}{%
\begin{tabular}{@{}l l l@{}}
\toprule
\textbf{Benchmark} & \textbf{arXiv} & \textbf{Panel source (results plot)} \\
\midrule
\multicolumn{3}{@{}l}{\textbf{\textsf{Leaderboard \& model-comparison charts}}}\\
World-in-World & \href{https://arxiv.org/abs/2510.18135}{2510.18135} & task-success / gen-quality leaderboard (embedded raster) \\
EvalCrafter & \href{https://arxiv.org/abs/2310.11440}{2310.11440} & overall model-ranking bar chart (embedded raster) \\
VideoScore & \href{https://arxiv.org/abs/2406.15252}{2406.15252} & annotation score-ratio grouped bars (embedded raster) \\
WorldModelBench & \href{https://arxiv.org/abs/2502.20694}{2502.20694} & VBench-vs-Ours win-rate comparison (embedded raster) \\
\midrule
\multicolumn{3}{@{}l}{\textbf{\textsf{Per-dimension radar \& multi-metric}}}\\
VBench & \href{https://arxiv.org/abs/2311.17982}{2311.17982} & 16-dimension evaluation radar (embedded raster) \\
EWMBench & \href{https://arxiv.org/abs/2505.09694}{2505.09694} & EWMScore 5-dimension radar (embedded raster) \\
WorldArena & \href{https://arxiv.org/abs/2602.08971}{2602.08971} & Fig.~1 (p.2): EWMScore ranking bars + evaluation radars \\
\midrule
\multicolumn{3}{@{}l}{\textbf{\textsf{Distributions \& composition}}}\\
EvalCrafter & \href{https://arxiv.org/abs/2310.11440}{2310.11440} & prompt-length histogram (embedded raster) \\
EvalCrafter & \href{https://arxiv.org/abs/2310.11440}{2310.11440} & meta-type composition pie chart (embedded raster) \\
VBench & \href{https://arxiv.org/abs/2311.17982}{2311.17982} & prompt word cloud (embedded raster) \\
\bottomrule
\end{tabular}}
\caption{\textbf{Per-panel provenance for Figure~\ref{fig:results_gallery}.} Every panel is a real
results/leaderboard/metric plot cropped from the benchmark's own arXiv paper and hyperlinked to its
source. Exact figure numbers are given where verified from the paper (e.g.\ WorldArena Fig.~1);
otherwise the extraction basis is noted (``embedded raster''). Plots are contain-fitted, not
cropped-to-fill, so no bars, axes or radar spokes are cut off. No figure number is invented. A
benchmark appears more than once when it reports several evidence types. Images \textcopyright\ their
respective authors, reproduced for scholarly review.}
\label{tab:results_provenance}
\end{table*}

%% file: tables/tab_collage_provenance.tex
\begin{table*}[t]
\centering
\footnotesize
\setlength{\tabcolsep}{6pt}
\renewcommand{\arraystretch}{1.12}
\resizebox{\textwidth}{!}{%
\begin{tabular}{@{}l l l@{\hspace{9mm}} l l l@{}}
\toprule
\textbf{Benchmark} & \textbf{arXiv} & \textbf{Source} & \textbf{Benchmark} & \textbf{arXiv} & \textbf{Source} \\
\midrule
\multicolumn{3}{@{}l}{\textbf{\textsf{Closed-loop (task-success suites)}}} &
\multicolumn{3}{l}{\textbf{\textsf{Open-loop (world-model \& video eval)}}}\\
LIBERO & 2306.03310 & Fig.~1 (p.2) & WorldModelBench & 2502.20694 & rep.\ figure (pp.1--3) \\
CALVIN & 2112.03227 & rep.\ figure (pp.1--3) & Physics-IQ & 2501.09038 & rep.\ figure (pp.1--3) \\
Meta-World & 1910.10897 & rep.\ figure (pp.1--3) & WorldScore & 2504.00983 & rep.\ figure (pp.1--3) \\
ManiSkill2 & 2302.04659 & rep.\ figure (pp.1--3) & EWMBench & 2505.09694 & rep.\ figure (pp.1--3) \\
RoboCasa & 2406.02523 & rep.\ figure (pp.1--3) & EVA-Bench & 2410.15461 & rep.\ figure (pp.1--3) \\
Habitat 3.0 & 2310.13724 & rep.\ figure (pp.1--3) & VideoPhy & 2406.03520 & rep.\ figure (pp.1--3) \\
ALFRED & 1912.01734 & rep.\ figure (pp.1--3) & Physion++ & 2306.15668 & rep.\ figure (pp.1--3) \\
GemBench & 2410.01345 & rep.\ figure (pp.1--3) & Physion & 2106.08261 & Fig.~1 (p.3) \\
VLABench & 2412.18194 & Fig.~1 Overview (p.1) & EvalCrafter & 2310.11440 & rep.\ figure (pp.1--3) \\
\bottomrule
\end{tabular}}
\caption{\textbf{Per-tile provenance for Figure~\ref{fig:corpus_collage}.} Every tile is a
representative figure cropped from the benchmark's own arXiv paper (page/figure noted; ``rep.\
figure'' marks the paper's teaser-region figure on pp.1--3), reproduced for scholarly review and
hyperlinked to its source. Images \textcopyright\ their respective authors.}
\label{tab:collage_provenance}
\end{table*}